\documentclass[]{opendatalab}
\usepackage{caption}
\usepackage{xcolor}
\usepackage{longtable}
\usepackage{listings}
\usepackage[numbers]{natbib}
\usepackage{amssymb}
\usepackage{tabularx}
\usepackage{float}
\usepackage{graphicx}
\usepackage{makecell}
\usepackage[table]{xcolor}
\usepackage{cleveref}
\usepackage{array}
\usepackage[misc]{ifsym}

\makeatletter
\newcommand\andauthor{%
  \g@addto@macro\authorlist{\par\vspace{1mm}}
}
\newcommand\nlauthor[2][]{%
  \addtolist[#1]{#2}{\authorlist}{\authorformat}{}
}
\makeatother

\RequirePackage{xspace}
\makeatletter
\DeclareRobustCommand\onedot{\futurelet\@let@token\@onedot}
\def\@onedot{\ifx\@let@token.\else.\null\fi\xspace}

\makeatother
\newcommand{\mineru}{MinerU2.5}

\definecolor{codegreen}{rgb}{0,0.6,0}
\definecolor{codegray}{rgb}{0.5,0.5,0.5}
\definecolor{codepurple}{rgb}{0.58,0,0.82}
\definecolor{backcolour}{rgb}{0.95,0.95,0.92}
\definecolor{promptcolor}{HTML}{D1D0F2}
\definecolor{promptcolorheader}{HTML}{bdbcec}
\newcommand{\promptbox}[2]{
\begin{tcolorbox}[
top=0.3em,bottom=0.3em,left=0.5em,right=0.5em,
toptitle=0.3em,bottomtitle=0.2em,boxsep=0pt,
colframe=promptcolorheader,colback=promptcolor!50,boxrule=0.5pt,
]
\footnotesize
\end{tcolorbox}
}
\lstdefinestyle{mystyle}{
    backgroundcolor=\color{backcolour},   
    commentstyle=\color{codegreen},
    keywordstyle=\color{magenta},
    numberstyle=\tiny\color{codegray},
    stringstyle=\color{codepurple},
    basicstyle=\ttfamily\footnotesize,
    breakatwhitespace=false,         
    breaklines=true,                 
    captionpos=b,                    
    keepspaces=true,                 
    numbers=left,                    
    numbersep=5pt,                  
    showspaces=false,                
    showstringspaces=false,
    showtabs=false,                  
    tabsize=2
}

\title{MinerU2.5: A Decoupled Vision-Language Model for Efficient High-Resolution Document Parsing}
\vspace{10pt}

\author[1,2*]{\ \quad\quad\quad\quad Junbo Niu}
\author[1,2*]{Zheng Liu}
\author[1*]{Zhuangcheng Gu}
\author[1*\ddagger]{Bin Wang}
\author[1*]{Linke Ouyang}
\andauthor
\nlauthor[1*]{\ \quad\quad Zhiyuan Zhao}
\author[1*]{Tao Chu}
\author[1*]{Tianyao He}
\author[1*]{Fan Wu}
\author[1,2*]{Qintong Zhang}
\author[1*]{Zhenjiang Jin}
\andauthor
\nlauthor[1]{\ \quad\quad\ \  Guang Liang}
\author[1]{Rui Zhang}
\author[1,2]{Wenzheng Zhang}
\author[1]{Yuan Qu}
\author[1]{Zhifei Ren}
\author[1]{Yuefeng Sun}
\andauthor
\nlauthor[1]{\ \quad Yuanhong Zheng}
\author[1]{Dongsheng Ma}
\author[1,3]{Zirui Tang}
\author[1,3]{Boyu Niu}
\author[1]{Ziyang Miao}
\author[1]{Hejun Dong}
\andauthor
\nlauthor[1,2]{\ \ Siyi Qian}
\author[1]{Junyuan Zhang}
\author[1,2]{Jingzhou Chen}
\author[1]{Fangdong Wang}
\author[1]{Xiaomeng Zhao}
\author[1]{Liqun Wei}
\andauthor
\nlauthor[1]{\ \ Wei Li}
\author[1]{Shasha Wang}
\author[1]{\;Ruiliang Xu}
\author[1]{\;Yuanyuan Cao}
\author[1]{\;Lu Chen}
\author[1]{\;Qianqian Wu}
\author[1]{\;Huaiyu Gu}
\andauthor
\nlauthor[1]{\ \ \ Lindong Lu}
\author[1]{Keming Wang}
\author[1]{\;Dechen Lin}
\author[1]{\;Guanlin Shen}
\author[1,3]{\;Xuanhe Zhou}
\author[3]{Linfeng Zhang}
\andauthor
\nlauthor[1]{\ \ Yuhang Zang}
\author[1]{Xiaoyi Dong}
\author[1]{Jiaqi Wang}
\author[1]{Bo Zhang}
\author[1]{Lei Bai}
\author[1]{Pei Chu}
\author[1]{Weijia Li}
\author[1]{Jiang Wu}
\andauthor
\nlauthor[1]{\quad\quad\quad\  Lijun Wu}
\author[1]{Zhenxiang Li}
\author[1]{Guangyu Wang}
\author[1]{Zhongying Tu}
\author[1]{Chao Xu}
\author[1]{Kai Chen}
\andauthor
\nlauthor[1]{\quad\quad\quad\quad\quad Yu Qiao}
\author[1]{Bowen Zhou}
\author[1\ \textrm{\Letter}]{Dahua Lin}
\author[1,2\ \textrm{\Letter}]{Wentao Zhang}
\author[1\ \textrm{\Letter}]{Conghui He}

\affiliation[1]{Shanghai Artificial Intelligence Laboratory}
\affiliation[2]{Peking University}
\affiliation[3]{Shanghai Jiao Tong University}

\abstract{

We introduce \mineru{}, a 1.2B-parameter document parsing vision-language model that achieves state-of-the-art recognition accuracy while maintaining exceptional computational efficiency. Our approach employs a coarse-to-fine, two-stage parsing strategy that decouples global layout analysis from local content recognition. In the first stage, the model performs efficient layout analysis on downsampled images to identify structural elements, circumventing the computational overhead of processing high-resolution inputs. In the second stage, guided by the global layout, it performs targeted content recognition on native-resolution crops extracted from the original image, preserving fine-grained details in dense text, complex formulas, and tables. To support this strategy, we developed a comprehensive data engine that generates diverse, large-scale training corpora for both pretraining and fine-tuning. Ultimately, \mineru{} demonstrates strong document parsing ability, achieving state-of-the-art performance on multiple benchmarks, surpassing both general-purpose and domain-specific models across various recognition tasks, while maintaining significantly lower computational overhead.

}

\metadata[* Equal contribution\quad $\textrm{\Letter}$ Corresponding author \quad $\ddagger$ Project leader]{}
\correspondence{Conghui He, \email{heconghui@pjlab.org.cn}}
\metadata[Code:]{\url{https://github.com/opendatalab/MinerU}}
\metadata[Model:]{\url{https://huggingface.co/opendatalab/MinerU2.5-2509-1.2B}}
\date{\today}

\begin{document}

\maketitle

\newpage
\tableofcontents
\newpage
\section{Introduction}
\label{section:intro}

\begin{figure}[!t]
    \centering
    \includegraphics[width=1.0\linewidth]{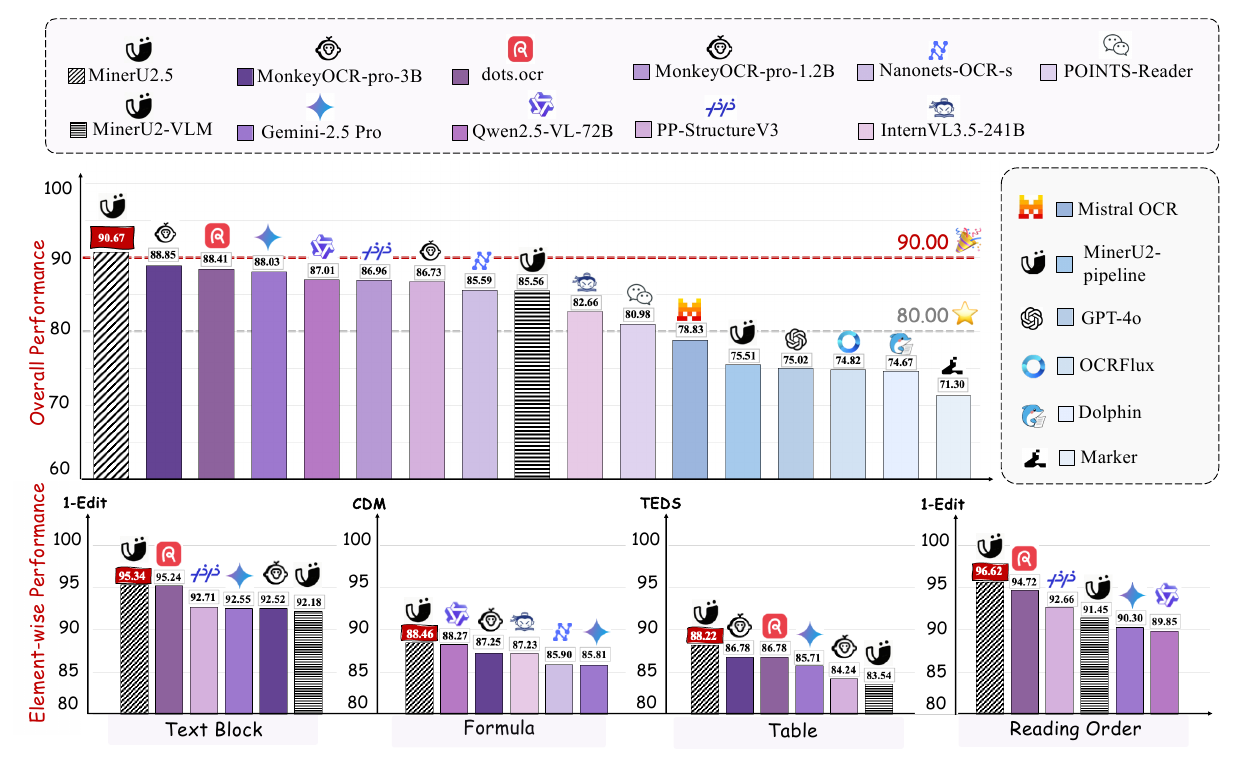}
    \caption{\textbf{Performance Highlights of \mineru{} on OmniDocBench.} \mineru{} consistently outperforms both general-purpose VLMs (e.g., Gemini-2.5 Pro, Qwen2.5-VL-72B, GPT-4o) and domain-specific models (e.g., MonkeyOCR, dots.ocr, PP-StructureV3), establishing new performance records in text recognition, formula recognition, table recognition, and reading order prediction. Detailed results are presented in~\Cref{tab:OmniDocbench-results}.}
\end{figure}

Document parsing~\cite{zhang2024document} serves as a fundamental task in multimodal understanding, underpinning a variety of downstream applications such as information extraction~\cite{liao2023doctr,wan2024omniparser}, Retrieval-Augmented Generation (RAG)~\cite{lin2024revolutionizing,zhang2024ocr,zhao2024retrieval} and intelligent document analysis~\cite{bai2022wukong,blecher2023nougat,tang2023unifying}. In contrast to natural images, document images are characterized by significantly higher resolutions, denser content, and more complex structural layouts~\cite{liu2024hrvda,wang2023vrdu,wei2024general}. These inherent properties introduce a unique set of challenges. Firstly, the high resolution and fine-grained layout structures necessitate models capable of processing images at their native resolution. Secondly, the text-dense and often lengthy nature of documents imposes stringent requirements on the parameter efficiency and robustness of the models. Thirdly, the success of OCR is contingent not only on precise text recognition but also heavily on reliable layout analysis and efficient inference.

Contemporary approaches to document parsing can be broadly categorized into two paradigms: pipeline-based approaches~\cite{cui2025paddleocr,livathinos2025docling,vik2024marker,wang2024mineru} and end-to-end approaches based on VLMs~\cite{achiam2023gpt,bai2025qwen2,comanici2025gemini,dots.ocr,wei2024general}. The former employs a modular design, decomposing the task into discrete stages such as layout detection, reading order prediction, and recognition of text lines, formulas, and tables. Each stage is handled by a specialized model. While this approach offers interpretability, it suffers from a cumbersome workflow and the potential for error propagation across modules. The latter paradigm exhibits superior semantic modeling capabilities, yet it is still widely constrained by the hallucination problem in long-document processing and suffers from severe efficiency bottlenecks when dealing with high-resolution inputs. A critical factor limiting the performance and efficiency of VLM-based parsing is token redundancy, arising from large blank or low-information regions within the document image.

In response to the aforementioned challenges, we introduce a new document parsing framework, \textbf{MinerU2.5}. The key innovation is a decoupled architecture that separates \textit{global layout analysis} from \textit{local content recognition} via an efficient coarse-to-fine, two-stage inference mechanism. In the first stage, the model conducts fast and holistic layout analysis on downsampled document images, capturing the global structural organization with minimal computational cost. In the second stage, guided by the detected layout, it crops key regions from the original high-resolution input and performs fine-grained recognition within local windows, thereby preserving native resolution and ensuring high accuracy.
This decoupled strategy not only reduces computational cost by an order of magnitude, primarily by avoiding the enormous number of visual tokens with $\mathcal{O}(N^2)$ complexity inherent in end-to-end native-resolution approaches~\cite{bai2025qwen2,chen2025ocean,dots.ocr}, but also brings multiple advantages: it significantly enhances the interpretability of parsing, effectively mitigates the common hallucination problem in VLMs, and allows the two stages to be independently optimized and iterated, resulting in more robust and efficient parsing capabilities. Ultimately, with its lightweight design of only 1.2B parameters, MinerU2.5 exhibits strong adaptability and efficiency in scenarios with long documents and high-density content while ensuring high parsing accuracy. 
Furthermore, to overcome the challenges of insufficient data diversity, sample imbalance, and inconsistent annotation quality in document parsing, we have developed a closed-loop data engine for complex documents. This engine systematically collects, processes, and generates large-scale, high-quality document corpora. This ensures that our model exhibits precise parsing capabilities and robustness across a wide spectrum of layouts, document types, and complex elements.

MinerU2.5 not only achieves state-of-the-art (SOTA) performance across a wide range of public benchmarks but also represents a qualitative leap in practical application and user experience over the previous MinerU2 version, as demonstrated by the examples in \hyperref[sec:app:qualitative]{Appendix~\ref{sec:app:qualitative}}
. Its key improvements include:
\begin{itemize}
    \item \textbf{Comprehensive and Granular Layout Analysis:} It not only preserves non-body elements like headers, footers, and page numbers to ensure full content integrity, but also employs a refined and standardized labeling schema. This enables a clearer, more structured representation of elements such as lists, references, and code blocks.
    \item \textbf{Breakthroughs in Formula Parsing:} Delivers high-quality parsing of complex, lengthy mathematical formulae and accurately recognizes mixed-language (Chinese-English) equations.
    \item \textbf{Enhanced Robustness in Table Parsing:} Effortlessly handles challenging cases, including rotated tables, borderless tables, and tables with partial borders.
\end{itemize}

\section{Related Work}
\label{section:rel}

\subsection{Traditional Pipelines}
Early OCR systems \citep{cui2025paddleocr,livathinos2025docling,vik2024marker,wang2024mineru} decompose document parsing into modular pipelines, sequentially executing layout detection \citep{wang2024yolov10,zhao2024doclayout}, text recognition \citep{cui2025paddleocr}, and reading order \citep{wang2021layoutreader}. For instance, Marker \citep{vik2024marker} implements a sequential pipeline integrating Surya OCR \cite{paruchuri2025surya} with layout analysis and reading order prediction modules to process diverse document types. MinerU \citep{wang2024mineru} leverages PDF-Extract-Kit \citep{pdfExtractKit2025} to orchestrate multiple specialized models for layout detection, formula recognition and table extraction. This modular architecture enables specialized optimization of individual components and facilitates targeted refinement of specific subtasks through well-defined module boundaries. However, pipeline-based methods are prone to error propagation across stages and exhibit limited robustness when confronted with complex layouts such as multi-column text or cross-page structures. Moreover, modular systems often entail multiple interdependencies in practice, rendering usage, maintenance, and updates cumbersome and less efficient.

\subsection{General-Purpose Vision Language Models}
General-purpose vision language models (VLMs) \citep{achiam2023gpt,bai2025qwen2,comanici2025gemini,zhu2025internvl3} have emerged as an alternative paradigm for document understanding. Gemini2.5 Pro \citep{comanici2025gemini} demonstrates strong OCR capabilities among general VLMs, surpassing traditional pipeline models like MinerU \citep{wang2024mineru} in text parsing and approaching specialized systems like UniMERNet \citep{wang2024unimernet} in formula recognition, showcasing the potential of VLMs in OCR applications. Among open-source models, Qwen2.5-VL-72B \citep{bai2025qwen2} achieves the best results, using native-resolution vision encoders \citep{dehghani2023patch} to adapt to different image sizes, demonstrating the effectiveness of arbitrary-resolution processing in OCR tasks. However, these general models exhibit inherent limitations for document-centric tasks. Proprietary models like Gemini2.5 Pro \citep{comanici2025gemini} are expensive and slow in processing, while open-source models require massive parameter scales for optimal performance, limiting practical deployment. Additionally, both types remain susceptible to hallucinations in densely populated text regions, affecting reliability in complex document layouts.
\subsection{Domain-Specific Vision Language Models}
\paragraph{End-to-End Approaches.}
Recent domain-specific models \citep{blecher2023nougat,chen2025ocean,kim2022ocr,liu2024textmonkey,poznanski2025olmocr,dots.ocr, wei2024general} adopt end-to-end architectures that unify document parsing within a single model, eliminating the need for cascaded processing stages. GOT \citep{wei2024general}, as an early representative of end-to-end approaches, pioneered the OCR 2.0 paradigm by establishing both model architecture and data methodology that unified recognition across diverse modalities—text, formulas, tables, and charts—within a single framework. Subsequent models like Ocean-OCR \cite{chen2025ocean}, olmOCR \citep{poznanski2025olmocr}, and dots.ocr \citep{dots.ocr} leverage native resolution vision encoders to process documents and construct massive document corpora, further advancing the performance of end-to-end architectures. However, end-to-end designs face scalability challenges: joint optimization of layout and content often reduces accuracy on complex documents, while native-resolution processing introduces prohibitive $\mathcal{O}(N^2)$ complexity. Despite strengths in semantic modeling, these models suffer from hallucinations on long documents and severe inefficiency with high-resolution inputs, where token redundancy from blank or low-information regions becomes a major bottleneck.

\paragraph{Multi-Stage Approaches.}
Recently, multi-stage methods \citep{feng2025dolphin,li2025monkeyocr} leveraging VLMs decouple layout analysis from content recognition, combining the efficiency of pipeline approaches with the accuracy of unified models. Dolphin \citep{feng2025dolphin} employs a Swin-Transformer VLM that first performs page-level layout, then conducts efficient parallel parsing of identified regions. However, Swin-Transformer's fixed resolution severely limits crop parsing—sub-regions with extreme aspect ratios suffer from distortion when resized to predetermined dimensions, degrading recognition quality while increasing computational overhead. MonkeyOCR \citep{li2025monkeyocr} adopts a similar multi-stage strategy but employs a native resolution vision encoder in its second stage, improving both performance and efficiency. However, MonkeyOCR requires multiple specialized models across different stages, increasing system complexity and deployment overhead. A single unified model with native resolution parsing presents a promising direction to address these limitations, which is precisely the goal that MinerU2.5 pursues.

\section{\mineru{}}
\label{section:\mineru{}}

\begin{figure*}[t]
    \centering
    \includegraphics[width=1.0\linewidth]{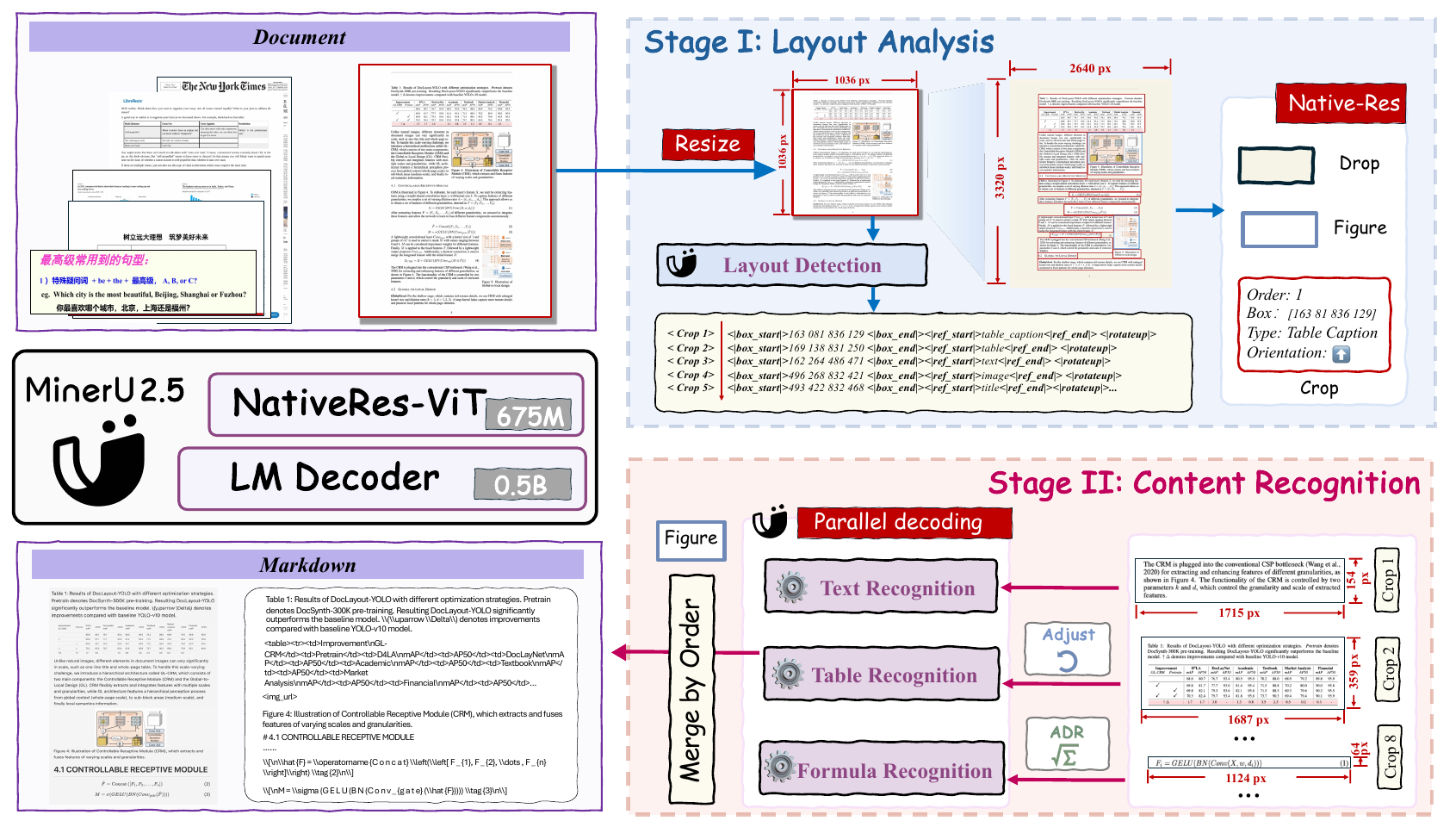}
    \caption{\textbf{The framework of \mineru{}.} In stage I, \mineru{} performs rapid, global layout analysis on a downsampled page. In stage II, \mineru{} leverages the layout results to crop key regions from the original high-resolution document, performing fine-grained content recognition (e.g., text, table, and formula recognition) within these native-resolution local regions. The detailed prompts used in the inference are illustrated in \hyperref[app:prompt_detail]{Appendix~\ref{app:prompt_detail}}.} 
    \label{fig:model}
\end{figure*}

\subsection{Model Architecture}
\label{section:\mineru{}arch}
\Cref{fig:model} illustrates the overall architecture of \mineru{}, which is inspired by the classical Qwen2-VL framework~\cite{wang2024qwen2}. The overall model architecture consists of three major components:

\paragraph{Language Model.}
For the decoder, we employ a 0.5B-parameter Qwen2-Instruct model~\cite{team2024qwen2}, as document parsing tasks typically exhibit relatively low dependency on large-scale language models. To better accommodate diverse resolutions and aspect ratios in cropped image parsing, we replace the original 1D-RoPE~\cite{su2024roformer} with M-RoPE~\cite{wang2024qwen2}, thus enhancing the model’s generalization ability across varying resolutions.

\paragraph{Vision Encoder.}
Inspired by Qwen2-VL, \mineru{} incorporates a native-resolution encoding mechanism. Although the Qwen2.5-VL series~\cite{bai2025qwen2} adopts window attention to improve efficiency, this design causes performance degradation in document parsing tasks. Therefore, we employ a 675M-parameter NaViT~\cite{dehghani2023patch} initialized from Qwen2-VL. This vision encoder supports dynamic image resolutions and employs 2D-RoPE for positional encoding, enabling it to flexibly handle inputs of various resolutions and aspect ratios. 

\paragraph{Patch Merger.}
To balance efficiency and performance, the architecture uses pixel-unshuffle~\cite{shi2016real} on adjacent 2 × 2 vision tokens, preprocessing the aggregated vision tokens before passing them into the large language model. This design effectively achieves a trade-off between computational efficiency and task performance.

\subsection{Two-Stage Parsing Strategy}

In high-resolution document parsing with VLMs, a large proportion of low-information blank regions introduces severe token redundancy, which substantially reduces overall efficiency. Existing end-to-end visual encoding strategies for VLMs face inherent limitations:

\begin{itemize}
\item \textbf{Crop-based approaches}~\cite{wei2024general,zhu2025internvl3} can partially reduce computational overhead but inevitably sacrifice semantic consistency and layout information.
\item \textbf{Native-resolution approaches}~\cite{bai2025qwen2,guo2025seed1,dots.ocr,niu2025native} preserve fine-grained details in high-resolution inputs, yet produce an enormous number of visual tokens with $\mathcal{O}(N^2)$ complexity, rendering them computationally impractical.
\end{itemize}

To address this dilemma, we propose a \textbf{two-stage parsing strategy}. This design decouples layout analysis from local content recognition, thereby improving interpretability, enhancing optimization potential for downstream tasks such as OCR, and effectively reducing the risk of hallucinations. Below, we provide more details of each stage.

\paragraph{Stage I: Layout Analysis.} 
In the first stage, the input image is uniformly resized to a thumbnail of $1036 \times 1036$ pixels, enabling global layout analysis while controlling computational cost. The parameter choice is determined through systematic analysis: the thumbnail size must balance global visibility and efficiency---too small leads to detail loss, while too large triggers the quadratic complexity of NaViT. In contrast to native-aspect-ratio thumbnails, adopting a fixed thumbnail size results in more stable bounding-box localization and facilitates more efficient training.

\paragraph{Stage II: Content Recognition.} 
In the second stage, the model leverages the detected layout to crop the native high-resolution image into local regions, which are then parsed at fine granularity. Cropped regions are fed at native resolution with an upper bound of $2048 \times 28  \times 28$ pixels, avoiding detail loss from overly small crops while preventing redundant computation from excessively large ones. This design ensures a robust trade-off between accuracy and efficiency across diverse document parsing scenarios.

\subsection{Training Recipe}

As described in~\Cref{section:\mineru{}arch}, \mineru{} consists of three core components: vision encoder, patch merger, and language model. Prior to the pre-training phase of \mineru{}, the vision encoder is initialized from Qwen2-VL-2B-Instruct, while the language model is initialized from Qwen2-Instruct-0.5B. The overall training procedure of \mineru{} is divided into three stages, as summarized in~\Cref{tab:training_strategy}.

\subsubsection{Stage 0-Modality Alignment}

To ensure that \mineru{} acquires the fundamental vision--language alignment ability as well as the OCR recognition capability, we first conduct two-stage modality alignment training on Visual Question Answering (VQA) datasets.

\paragraph{Language-Image Alignment.} 
Only the two-layer MLP within the patch merger is trained, while both the vision encoder and the language model are frozen. We use image-caption pairs\footnote{This dataset is sourced from \href{https://huggingface.co/datasets/liuhaotian/LLaVA-Pretrain}{\textit{LLaVA-Pretrain}}.} for training to effectively project visual features into the LLM embedding space, thus achieving alignment of the modal representation.

\paragraph{Visual Instruction Tuning.} 
All model parameters are unfrozen. The focus is on knowledge accumulation and ability expansion, particularly strengthening visual alignment and OCR capability. The training data\footnote{This dataset is sourced from \href{https://huggingface.co/datasets/liuhaotian/LLaVA-Instruct-150K/tree/main}{
\textit{LLaVA-Instruct}}.} mainly covers image captioning, interleaved text-image pairs, visual alignment, and OCR data. The goal is to enable \mineru{} to follow instructions across diverse visual tasks and generate reasonable responses.

\begin{table}[htp]
    \centering
    \setlength{\tabcolsep}{12pt}
    \renewcommand{\arraystretch}{1.2}
    \resizebox{\textwidth}{!}
    { 
    \begin{tabular}{@{}ll|c|c|c|c@{}}
    \toprule
    & & \multicolumn{2}{c|}{\textbf{Stage-0}} & \textbf{Stage-1} & \textbf{Stage-2} \\ \cmidrule(l){3-4}
    & & \textbf{a} & \textbf{b} & & \\
    \midrule 
    \multirow{2}{*}{\rotatebox[origin=c]{90}{\footnotesize \textit{Vision}}}
    & \textbf{Max Resolution}   
        & $2048 \times 28 \times 28$
        & $4096 \times 28 \times 28$
        & $2048 \times 28 \times 28$
        & $2048 \times 28 \times 28$ \\
    & \#Tokens per Image 
        & $4 \sim 2048$ 
        & $4 \sim 4096$  
        & $4 \sim 2048$ 
        & $4 \sim 2048$  \\
    \midrule 
    \multirow{2}{*}{\rotatebox[origin=c]{90}{\footnotesize \textit{Data}}}
    & \textbf{Dataset} 
        & Image Caption
        & VQA
        & Layout\&OCR
        & Layout\&OCR \\
    & \#Samples 
        & 558K 
        & 665K 
        & 6.9M 
        & 630K \\
    \midrule
    \multirow{3}{*}{\rotatebox[origin=c]{90}{\footnotesize \textit{Model}}}
    & \textbf{Trainable} 
        & MLP Adaptor 
        & All 
        & All 
        & All \\
    & \textbf{Sequence Length}
        & 4096 
        & 4096 
        & 8192 
        & 16384 \\
    & \textbf{Data Augmentation}
        & No 
        & No 
        & Yes 
        & Yes \\
    \midrule 
    \multirow{3}{*}{\rotatebox[origin=c]{90}{\footnotesize \textit{Training}}}
    & \textbf{Batch Size} 
        & 128 
        & 64 
        & 256 
        & 256 \\
    & \textbf{LR: $\psi_{\text{ViT}}$} 
        & 1 $\times 10^{-3}$ 
        & 1 $\times 10^{-5}$ 
        & 4 $\times 10^{-6}$ 
        & 4 $\times 10^{-6}$ \\    
    & \textbf{LR: $\{\theta_{\text{MLP}}, \phi_{\text{LM}}\}$} 
        & 1 $\times 10^{-3}$ 
        & 1 $\times 10^{-5}$ 
        & 4 $\times 10^{-5}$ 
        & 4 $\times 10^{-5}$ \\
    & \textbf{Epoch} 
        & 1 & 1 & 2 & 3 \\
    \bottomrule
    \end{tabular}
    } 
    \vspace{1mm}
    \caption{Training setup and hyperparameters in three training stages.} 
    \vspace{-5mm}
    \label{tab:training_strategy}
\end{table}

Empirical results demonstrate that \mineru{}, after VQA-based modality alignment training, exhibits significant improvements in tasks such as layout analysis and content recognition. Conversely, skipping this stage leads to higher losses and a clear drop in overall performance.

\subsubsection{Stage 1-Document Parsing Pre-training}

The objective of the document parsing pre-training stage is to enable \mineru{} to acquire two fundamental capabilities: \textbf{layout analysis} and \textbf{content recognition}. At this stage, all parameters of the model remain fully trainable.

\paragraph{Training Data.}
We leveraged a large-scale mixture of model-labeled data and public datasets to ensure both sufficient scale and document diversity. 
For layout analysis, in consideration of training efficiency, full document images were resized to a fixed resolution with corresponding relative coordinates, and the prompt ``\texttt{Layout Detection:}'' was used. 
For content recognition, we employed single-element image samples of text blocks, formula blocks, and table blocks as inputs, with prompts ``\texttt{Text Recognition:}'', ``\texttt{Formula Recognition:}'', and ``\texttt{Table Recognition:}'' respectively. More details are shown in the \hyperref[app:prompt_detail]{Appendix~\ref{app:prompt_detail}}.

\paragraph{Training Configuration.}
The model, initialized from Stage~0, was trained for 2 epochs. 
Each epoch consisted of a total of 6.9M samples, including 2.3M for layout analysis, 2.4M for text blocks, 1.1M for formula blocks, and 1.1M for table blocks. 

Through this document parsing pre-training, the model has acquired strong layout analysis and content recognition capabilities, demonstrating excellent performance across most simple and medium-level scenarios. 
The resulting model not only serves as a \textbf{strong baseline} for downstream fine-tuning, but also functions as an \textbf{efficient hard-sample miner} within our data engineering pipeline, facilitating the identification of challenging cases for human annotation and further improving document parsing performance.

\subsubsection{Stage 2-Document Parsing Fine-tuning}

The objective of the document parsing fine-tuning stage is to further enhance parsing performance in challenging scenarios, while maintaining the detection and parsing capabilities already acquired by \mineru{}.

\paragraph{Training Data.}  
To achieve this goal, it is crucial to construct a compact yet high-quality dataset:  
\begin{itemize}
    \item To preserve the model's fundamental capabilities, we sampled high-quality and diverse examples from the pre-training dataset via data engineering and incorporated them into Stage~2 training, ensuring broad coverage across different document element types.
    \item From a large-scale, multi-source PDF corpus, we employed data engineering to identify cases where the model still underperformed. We summarized these difficult scenarios and conducted targeted data collection with manual annotation to obtain high-quality samples representing challenging cases. 
\end{itemize}


\paragraph{Training Configuration.}
We fine-tuned the pre-trained model for 3 epochs. Each epoch contained a total of 630K samples, consisting of 43K for layout analysis, 300K for text blocks, 147K for formula blocks, and 140K for table blocks. 

With this targeted data iteration strategy, Stage~2 fine-tuning enables the model to not only retain its established document parsing abilities but also achieve significant improvements in previously challenging scenarios.

\subsubsection{Data Augmentation Strategies}
To enhance the model's robustness in handling diverse documents in an open-world setting, we designed a variety of targeted data augmentation strategies during both Stage~1 and Stage~2. 
These augmentations simulate common types of document interference, and can be categorized as shown in~\Cref{tab:doc-aug-1}.

\begin{table}[h]
\centering
\begin{tabular}{l|l}
\toprule
\textbf{Augmentation Type} & \textbf{Operations} \\
\midrule
Spatial Transformations & Scaling, Grid Distortion, Rotation \\
Background Transformations & Texture, Weather effect, Image background, \\
 & Watermark, Scanlines, Shadow \\
Color Transformations & Brightness Contrast, Illumination, RGB Shift \\
Degradation Transformations & PSF Blur, Vibration Blur, Gaussian Blur, \\
 & Erosion / Dilation \\
\bottomrule
\end{tabular}
\caption{Data augmentation strategies for document parsing.}
\label{tab:doc-aug-1}
\end{table}

Note that spatial transformations are not applied to layout analysis samples. 
For different element types, we carefully design augmentation parameters and probabilities in order to strike a balance between model performance and robustness.

\subsection{Model Deployment}
We implement an efficient offline inference pipeline for \mineru{} based on vLLM~\citep{kwon2023efficient}. While vLLM provides high-throughput serving for large language models, we introduce two additional optimizations tailored for our two-stage document parsing pipeline to further minimize end-to-end latency. First, we employ an asynchronous backend to handle batching submission of page-level requests, enabling better overlap between CPU and GPU workloads. Second, we decouple Stage~I and Stage~II into independent inference tasks, allowing downstream processing to begin as soon as individual results become available, rather than waiting for entire batches.

A key challenge during deployment was suppressing degenerate token repetition without penalizing legitimate repetitive structures (e.g., tables, equations, or structured content). To address this, we dynamically adjust sampling parameters like \texttt{frequency\_penalty} and \texttt{presence\_penalty} in Stage~II based on the layout type detected in Stage~I. For instance, higher penalties are applied to text paragraphs, while lower values are used for tabular content. 

Furthermore, we carefully tuned key vLLM scheduling parameters, including \texttt{max\_num\_batched\_tokens}, \texttt{max\_num\_seqs}, and \texttt{cuda\_graph\_sizes}, to improve batch utilization and kernel launch efficiency.

We evaluate all compared models on OmniDocBench~\citep{ouyang2025omnidocbench}, a dataset of 1,355 document pages with an average of over 1,100 tokens per page. All models are tested using their official inference scripts under a consistent batched parallel processing protocol, with vLLM startup overhead excluded for fair comparison. After preliminary optimization, \mineru{} achieves an end-to-end throughput of \textbf{2.12 pages/s}. The end-to-end generation speed, measured only on valid output tokens from Stage~II, reaches \textbf{2337.25 tokens/s}\footnote{The end-to-end generation speed is calculated based on the number of valid tokens produced by Stage II divided by the total processing time for both stages.}. As shown in~\Cref{tab:infer-throughput_c2e}, \mineru{} outperforms MonkeyOCR-Pro-3B by $4\times$ and dots.ocr by $7\times$ in page throughput, demonstrating strong inherent efficiency for large-scale document parsing. Notably, even without any deployment optimizations, \mineru{} achieves a baseline throughput of 0.95~pages/s and 1045.14~tokens/s, already surpassing other compared models under default configurations.

\begin{table}[h]
\centering
\setlength{\tabcolsep}{12pt}
\renewcommand{\arraystretch}{1.2}
\resizebox{1.0\textwidth}{!}{
\begin{tabular}{c@{\hspace{1.2pt}}c|c|c|cc}
\toprule
\textbf{Model} & \textbf{Parameters} & \textbf{Backend} & \textbf{Hardware} & \textbf{Tokens/sec} & \textbf{Pages/sec} \\
\midrule
MinerU2-VLM \citep{wang2024mineru} & 0.9B & SGLang~\citep{zheng2024sglang} & \multirow{5}{*}{A100 80G} & 3091.23 & 2.84 \\
\cline{3-3}
dots.ocr \citep{dots.ocr} & 3.0B & \multirow{4}{*}{vLLM~\citep{kwon2023efficient}} & & 311.06 & 0.28 \\
MonkeyOCR-pro-3B \citep{li2025monkeyocr} & 3.7B & & & 520.16 & 0.47 \\
MonkeyOCR-pro-1.2B \citep{li2025monkeyocr} & 1.9B & & & 589.76 & 0.53 \\
Nanonets-OCR-s \citep{nanonets2025} & 3.7B & & & 605.92 & 0.55 \\
\midrule
\multirow{3}{*}{\mineru{}} & \multirow{3}{*}{1.2B} & \multirow{3}{*}{vLLM} & RTX 4090 48G & 1875.82 & 1.70 \\
 & & & A100 80G & \textbf{2337.25} & \textbf{2.12} \\
 & & & H200 141G & 4938.31 & 4.47 \\
\bottomrule
\end{tabular}
}
\caption{Inference performance comparison of specialized VLMs and \mineru{} across different backends and GPUs.}
\label{tab:infer-throughput_c2e}
\end{table}

\begin{figure*}[h]
    \centering
    \includegraphics[width=1.0\linewidth]{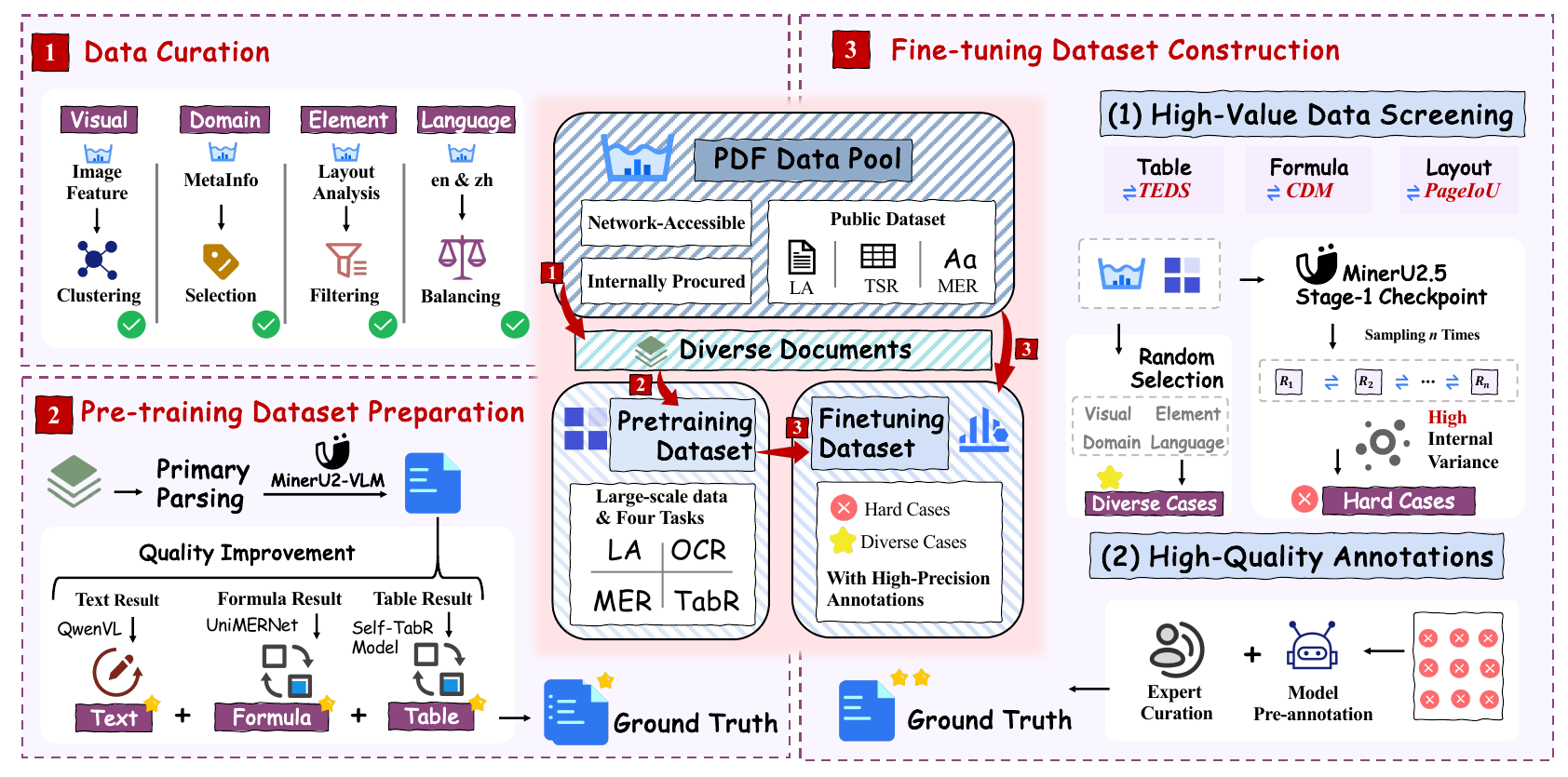}
    \caption{\textbf{Overview of the Data Engine.} Our data pipeline consists of three core stages. (1) \textbf{Data Curation:} We filter a massive, raw document pool to construct a diverse and balanced dataset based on layout, document type, element balance, and language. (2) \textbf{Pre-training Data Preparation:} We generate automated annotations for the curated data and then refine them using specialized, powerful models for text, tables, and formulas to ensure high quality. (3) \textbf{Fine-tuning Dataset Construction:} We employ our Iterative Mining via Inference Consistency (IMIC) strategy to automatically discover hard cases, which then undergo meticulous expert curation to create a high-quality SFT dataset.}
    \label{fig:data_pipeline}
\end{figure*}

\section{Data Engine}
\label{section:data}


The state-of-the-art performance of \mineru{} is underpinned by a systematic Data Engine designed to generate large-scale, high-quality training data with uniform annotation standards. This engine first establishes a vast and diverse foundation through rigorous data curation and refined automated annotation for pre-training. Building upon this foundation, we introduce our novel Iterative Mining via Inference Consistency (IMIC) strategy, which efficiently identifies complex ``hard cases" for targeted human annotation. This multi-stage approach creates a virtuous cycle of improvement, progressively enhancing the model's capabilities. The entire process is illustrated in~\Cref{fig:data_pipeline}.

\subsection{Overall Workflow}

\subsubsection{Data Curation}

Our process begins with a large-scale internal document pool comprising publicly available web data and commercially procured documents. While diverse, this raw pool suffers from a significant long-tail distribution. To mitigate this imbalance and enhance training robustness, we implement a rigorous curation process to build a balanced Chinese-English dataset with high diversity across multiple dimensions:

\begin{itemize} 
    \item \textbf{Layout Diversity:} We employ page-level image clustering to select exemplars from a wide spectrum of visual layouts and styles. 
    \item \textbf{Document Type Diversity:} Using document metadata (e.g., discipline, tags), we perform stratified sampling to ensure a balanced representation of types such as academic papers, textbooks, reports, and presentations. 
    \item \textbf{Element Balance:} A preliminary detection model helps ensure a balanced class distribution of key elements like titles, paragraphs, tables, formulas, and figures in the curated set. 
    \item \textbf{Language Balance:} We filter the data to maintain a comparable volume of Chinese and English documents. 
\end{itemize}

\subsubsection{Pre-training Dataset Preparation}

Initial annotations for the curated dataset are generated using our MinerU2-pipeline, establishing a baseline for subsequent refinement. To move beyond this baseline quality, we perform a multi-step refinement process using specialized, expert models for different content types:

\begin{itemize} 

    \item \textbf{Textual Content:} We leverage the powerful Qwen2.5-VL-72B-Instruct to verify and correct initial text recognition results on cropped text regions. 
    \item \textbf{Formula Content:} Recognized formulas are substituted with higher-fidelity outputs from an in-house UniMERNet model, which we retrained on our extensive formula dataset to boost its accuracy. 
    \item \textbf{Table Content:} All table structures are re-generated using an in-house, high-performance table parsing model. 

\end{itemize}

This refinement workflow yields a high-quality pre-training dataset of image-annotation pairs, covering our four core tasks: layout analysis, text recognition, formula recognition, and table recognition.

\subsubsection{Fine-tuning Dataset Construction}

While pre-training ensures broad capabilities, the noise inherent in automated annotations creates a ceiling for model performance. To break through this ceiling, our fine-tuning strategy pivots to high-value, difficult examples. We designed an Iterative Mining via Inference Consistency (IMIC) strategy to automatically filter these hard cases from the large-scale data pool. To ensure annotation quality, these select samples are processed through an AI-assisted pipeline: they are first pre-annotated by a foundation model, such as Gemini-2.5-Pro for complex tables, and then meticulously reviewed and corrected by human experts\footnote{Human review is augmented by our open-source QA tool, Dingo, which applies both rule-based and model-based checks. See \url{https://github.com/MigoXLab/dingo.}}. The final Supervised Fine-Tuning (SFT) dataset combines these high-quality hard cases with a smaller, randomly sampled set of regular examples, equipping \mineru{} to excel in complex, real-world parsing scenarios.

\subsection{Task Reformulation and Enhancement}

To move beyond the limitations of existing document analysis methods, we systematically reformulated the core tasks of layout analysis, formula recognition, and table recognition. This involved defining more robust standards, designing novel task paradigms, and introducing specialized metrics and representations.

\subsubsection{Layout Analysis}

\textbf{A Unified Tagging System}. A fundamental challenge in layout analysis is the lack of a standardized tagging system. Existing datasets suffer from widespread inconsistencies in element definitions, granularity, and scope. To address this, we engineered a hierarchical and comprehensive tagging system by analyzing a vast corpus of documents. Our system is defined by three key principles:

\begin{itemize} 
    \item \textbf{Comprehensive Coverage:} It includes non-body content often ignored by others, such as headers, footers, and page numbers, which is critical for downstream applications like RAG. 
    \item \textbf{Fine Granularity:} It decomposes complex elements. For instance, figures are sub-categorized into image, chart, and chemical\_structure, with distinct tags for their associated captions. 
    \item \textbf{Semantic Distinction:} Visually distinct text blocks like code, algorithms, references, and lists are assigned their own categories to preserve crucial semantic information. 
\end{itemize} 

\Cref{tab:Categories} presents a comparison with mainstream tagging systems, highlighting the superior coverage and granularity of our proposed system.

\begin{table}[h]
\footnotesize
\centering
\setlength{\tabcolsep}{4pt} 
\renewcommand{\arraystretch}{1.2} 
\begin{tabular}{@{}llll@{}}
\toprule
\multicolumn{1}{l}{\textbf{Category}} & \textbf{MinerU2-pipeline} & \textbf{PaddleOCR} & \textbf{MinerU2.5} \\
\midrule
\multirow{12}{*}{\textbf{Textual}} 
    & text & text, toc, abstract & text \\
    & title & title, page\_title & title \\
    & $\times$ & $\times$ & phonetic \\
    & image\_caption & common\_caption & image\_caption \\
    & image\_footnote & common\_footnote & image\_footnote \\
    & table\_caption & common\_caption & table\_caption \\
    & table\_footnote & common\_footnote & table\_footnote \\
    & $\times$ & code & code \\
    & $\times$ & $\times$ & code\_caption\\
    & $\times$ & $\times$ & algorithm \\
    & $\times$ & ref\_text, ref\_block & reference \\
    & $\times$ & $\times$ & list \\
\addlinespace[2pt]
\midrule
\textbf{Image} 
    & image & image, seal, chart, molecular & image \\
\addlinespace[2pt]
\midrule
\textbf{Table} 
    & table & table & table \\
\addlinespace[2pt]
\midrule
\multirow{2}{*}{\textbf{Equation}}
    & equation & equation & equation \\
    & $\times$ & $\times$ & equation\_block \\
\addlinespace[2pt]
\midrule
\multirow{5}{*}{\textbf{Page Margins}}
    & $\times$ & header & header \\
    & $\times$ & footer & footer \\
    & $\times$ & aside\_text & aside\_text \\
    & $\times$ & page\_number & page\_number \\
    & $\times$ & page\_footnote & page\_footnote \\
\bottomrule
\end{tabular}
\caption{Comparison of category support across different OCR systems.}
\label{tab:Categories}
\end{table}

\textbf{An Enhanced Multi-Task Paradigm}. Traditional methods often treat layout analysis as a standard object detection task, which ignores element rotation and defers reading order prediction to downstream modules. This approach not only impairs the recognition of rotated elements but also increases system coupling. We propose an enhanced paradigm that redefines layout analysis as a multi-task problem. This paradigm simultaneously predicts four key attributes for each document element in a single inference pass: its \textbf{Position}, \textbf{Class}, \textbf{Rotation Angle}, and \textbf{Reading Order}. This integrated design effectively resolves the challenge of parsing rotated elements and streamlines the entire document analysis pipeline.

\begin{figure*}[t]
    \centering
    \includegraphics[width=1.0\linewidth]{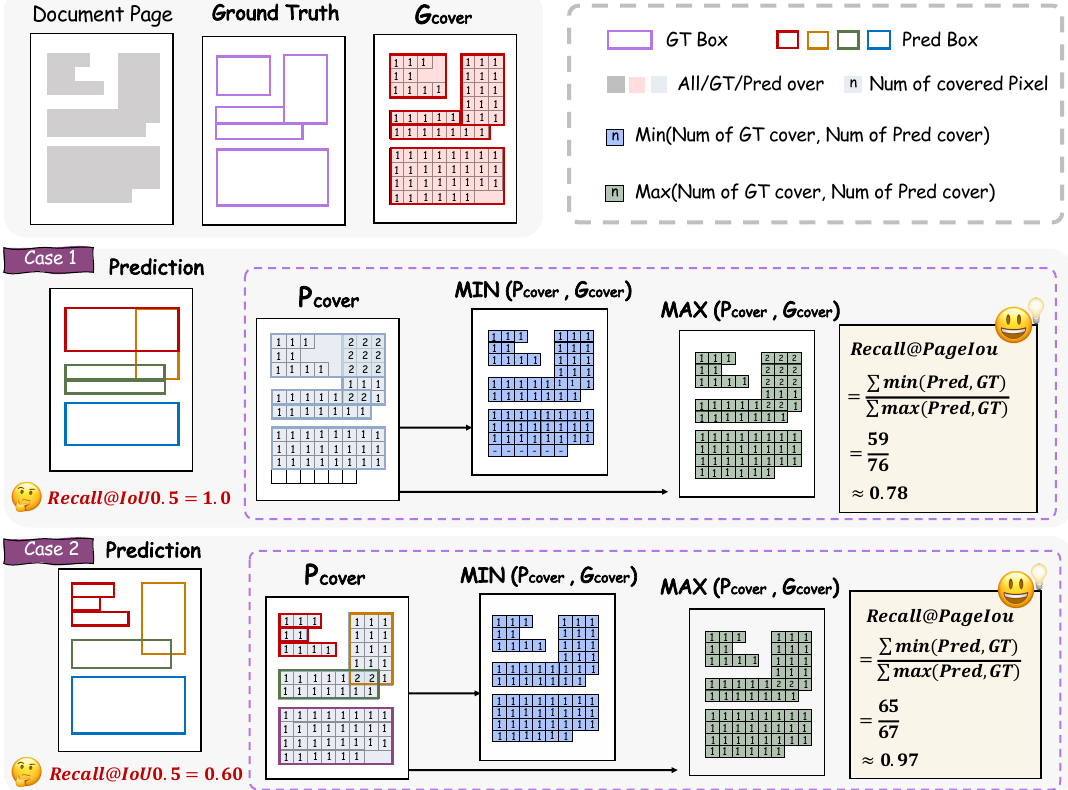}
    \caption{\textbf{Illustration of the proposed PageIoU metric.} Case 1 and Case 2 show that IoU-based recall may produce contradictory results compared with visual inspection, whereas PageIoU provides a page-level coverage score that aligns more closely with qualitative observations.}
    \label{fig:pageiou}
\end{figure*}

\textbf{PageIoU: A New Metric for Layout Quality}. Layout analysis is typically evaluated with object detection metrics like mAP, which rely on a fixed Intersection over Union (IoU) threshold. While effective for well-defined objects, this approach is ill-suited for document layouts where text block boundaries are often ambiguous. This can lead to a discrepancy where quantitative IoU-based scores do not align with qualitative visual assessment.

As illustrated in~\Cref{fig:pageiou}, a prediction that coarsely covers a paragraph (Case 1) can achieve a perfect recall score (Recall@IoU0.5 = 1.0), while a more accurate line-by-line prediction (Case 2) is penalized for not matching the paragraph-level ground truth, yielding a lower score (Recall@IoU0.5 = 0.6). Visually, however, Case 2 is clearly a better fit.




To better evaluate document layout analysis, we introduce \textbf{PageIoU}, 
a page-level coverage metric that measures the spatial consistency between 
predicted layouts and ground-truth annotations.  
Let the predicted layout be
\[
P = \{bbox_i \mid i = 1, 2, \dots, n\},
\]
and the ground truth be
\[
G = \{bbox_j \mid j = 1, 2, \dots, m\},
\]
where each $bbox$ denotes a bounding box on the page. 
We first compute coverage maps for both prediction and ground truth. 
For example, the ground-truth coverage map is defined as:
\[
G_{cover} = \left\{ \sum_{j=1}^{m} 1_{p \in bbox_j} \,\middle|\, p \in M \right\},
\]
where $p$ is a page pixel and $M$ denotes the non-background region of the page. 
Similarly, $P_{cover}$ can be obtained. Based on these, PageIoU is defined as:
\[
\text{PageIoU}(P, G) =
\frac{|P_{cover} \cap G_{cover}|}{|P_{cover} \cup G_{cover}|}
= \frac{\sum_{p \in M} \min \{ P_{cover}(p), G_{cover}(p) \}}
{\sum_{p \in M} \max \{ P_{cover}(p), G_{cover}(p) \}}.
\]

Here, $|\cdot|$ denotes the summation over all pixel values, 
while $\cap$ and $\cup$ correspond to the pixel-wise minimum and maximum 
of coverage counts, respectively. As shown in~\Cref{fig:pageiou}, PageIoU aligns with human perception, scoring the qualitatively poor prediction 0.78 and the superior one 0.97.


\begin{figure*}[t]
    \centering
    \includegraphics[width=1.0\linewidth]{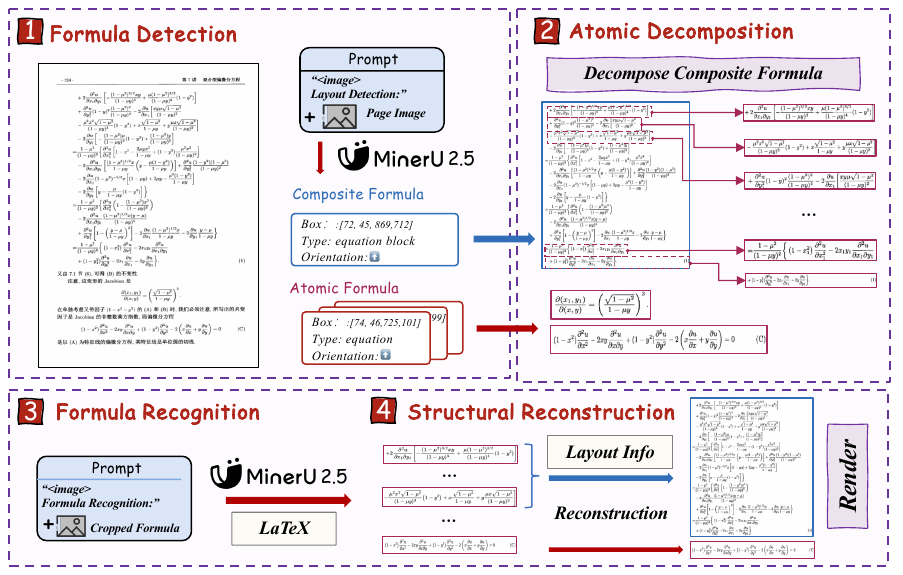}
    \caption{\textbf{The proposed ADR framework.} First, a compound formula is decomposed into atomic lines via layout analysis. Next, each line is individually recognized into LaTeX. Finally, the individual results are structurally recombined to produce the complete output.}
    \label{fig:formula_recognition}
\end{figure*}

\subsubsection{Formula Recognition}

\textbf{Decoupling Atomic and Compound Formulas}. Existing models struggle with long or multi-line formulas, and VLMs are prone to severe structural hallucinations. We identify the root cause as the tendency to treat all formulas as monolithic entities, failing to account for internal complexity. To this end, \mineru{} introduces a "whole-part" decoupling philosophy, classifying formulas into two types based on their structural and semantic integrity:

\begin{itemize} 
    \item \textbf{Atomic Formulas:} The smallest, indivisible semantic units with a tight 2D topology (e.g., a single fraction, a matrix). 
    \item \textbf{Compound Formulas:} An ordered set of atomic formulas composed vertically with specific alignment relationships (e.g., a multi-line derivation aligned at the equal signs). 
\end{itemize}

\textbf{The Atomic Decomposition \& Recombination (ADR) Framework}. To handle the complexity of compound formulas, we propose the ADR framework, which implements a multi-stage "divide and conquer" strategy. As illustrated in \Cref{fig:formula_recognition}, the ADR pipeline is powered by our versatile MinerU2.5 model, which acts as both a layout analyzer and a recognition engine, guided by task-specific prompts. The process begins with an initial layout analysis pass, where MinerU2.5, guided by a layout detection prompt, identifies and classifies all formula regions on the page as either atomic or compound. Next, in the decomposition stage, each identified compound formula is segmented into an ordered sequence of its constituent atomic formula lines, which are then cropped as individual images. In the third stage, these simple, semantically independent atomic formula images are fed back into the MinerU2.5 model. This time, using a formula recognition prompt, the model performs high-precision translation of each image into its corresponding LaTeX string. Finally, a lightweight recombination step uses the positional information from the initial layout pass to structurally reassemble the individual LaTeX strings into a single, coherent block, correctly formatting them within environments like align. This approach transforms a single, difficult recognition task into a series of simpler ones, ensuring both high-fidelity recognition of each component and the logical integrity of the overall structure.

\begin{figure*}[t]
    \centering
    \includegraphics[width=1.0\linewidth]{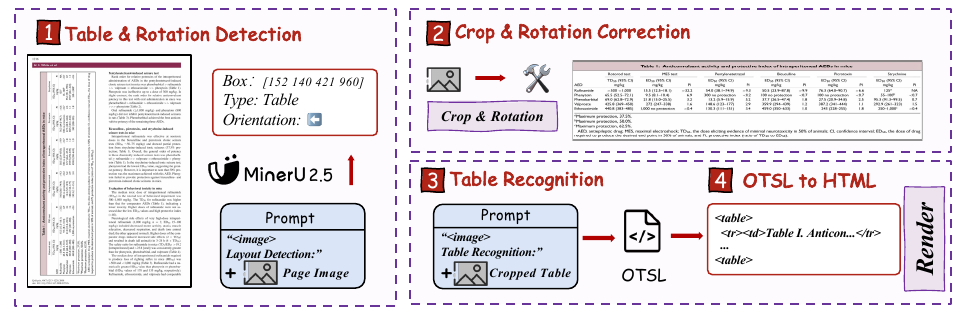}
    \caption{\textbf{The Table Recognition Pipeline.} The pipeline first detects a table and its rotation, then corrects its geometry. Next, the rectified image is recognized into the OTSL result, which is finally converted to standard HTML.}
    \label{fig:table_recognition}
\end{figure*}

\subsubsection{Table Recognition}

\textbf{Overcoming Long-Sequence Dependencies}. A primary challenge in table recognition is parsing complex, long tables, especially for VLM-based approaches that target HTML. We attribute this difficulty to two inherent weaknesses of the HTML representation: (1) its complex, non-visual syntax must be learned implicitly by the model; and (2) its high token redundancy results in excessively long sequences, degrading performance on large tables. (The issue of rotated tables is effectively handled by our enhanced layout paradigm.)

\textbf{OTSL: An Optimized Table Structure Language}. To robustly handle complex tables, we propose a four-stage recognition pipeline, as depicted in \Cref{fig:table_recognition}. The first two stages handle geometric normalization: the system detects the table's bounding box and rotation angle, then corrects the image by cropping and rotating it to a canonical orientation. For the crucial third stage, table recognition, we leverage the Optimized Table-Structure Language (OTSL)~\cite{lysak2023optimized}, an intermediate representation developed by IBM [citation, 2023]. We adopted OTSL for its significant advantages over HTML as a target for VLMs. Its minimalist design features a direct structural correspondence to a table's visual 2D matrix, reducing the number of structural tokens from over 28 to just 5 and shortening the average sequence length by approximately 50\%. This makes it a far more effective target for model generation. The final stage is a straightforward conversion from the OTSL output into standard HTML.

\subsection{Iterative Mining via Inference Consistency}

\begin{figure*}[!t]
    \centering
    \includegraphics[width=1.0\linewidth]{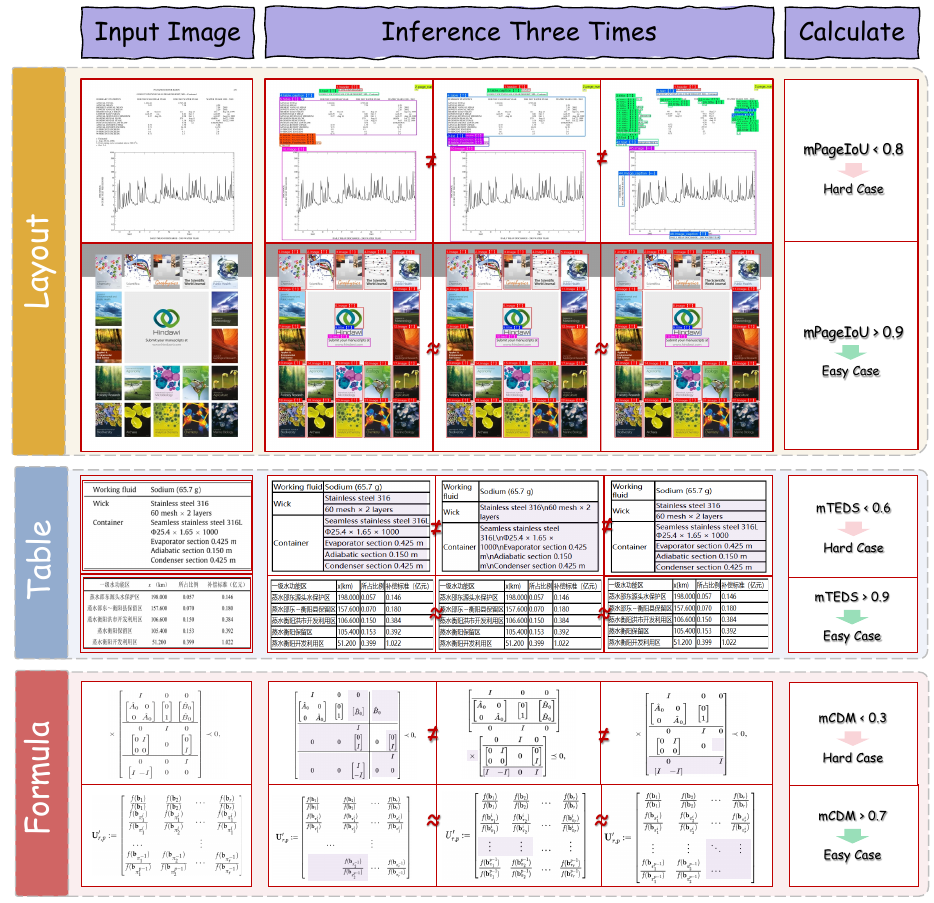}
    \caption{llustration of the proposed \textbf{IMIC (Iterative Mining via Inference Consistency) strategy}. From top to bottom: (a) Layout analysis, (b) Table recognition, and (c) Formula recognition. For each task, the model performs multiple stochastic inference runs, and the pairwise consistency between outputs is calculated with task-specific metrics (PageIoU, TEDS, CDM). Samples with low consistency are automatically identified as hard cases and prioritized for manual annotation.}
    \label{fig:IMIC}
\end{figure*}




To enable continuous model improvement and the efficient expansion of our high-quality training dataset, we introduce the IMIC (Iterative Mining via Inference Consistency) strategy. IMIC automatically identifies the most challenging samples—or "hard cases"—for the current model from a large corpus of unlabeled data. This allows us to direct limited human annotation efforts toward the data that offers the maximum value for model improvement.

The core principle of IMIC leverages the stochasticity inherent in model inference. For a given sample, if the model has learned its features robustly, multiple inference passes with stochastic sampling enabled should yield highly consistent outputs. Conversely, significant divergence across outputs suggests the sample lies near the model's decision boundary—a 'hard case' where its predictions are uncertain. Such samples are the most valuable candidates for manual annotation, as they directly target the model's specific weaknesses.

As illustrated in \Cref{fig:IMIC}, the implementation is tailored to each recognition task:

\begin{itemize} 
    \item \textbf{Layout analysis:} For full document pages, we perform multiple inference runs and measure consistency by calculating the pairwise PageIoU between the resulting layouts. Samples falling below a predefined similarity threshold are flagged as hard cases for precise manual annotation. 
    \item \textbf{Formula Recognition:} For cropped formula images, consistency is assessed using the pairwise CDM~\cite{wang2025image}  across multiple outputs. Samples with low consistency are prioritized for manual correction.
    \item \textbf{Table Recognition:} For cropped table images, we use the TEDS (Tree-Edit-Distance-based Similarity) score to evaluate consistency across multiple recognized structures. Low-consistency samples are routed to the manual annotation workflow. 
\end{itemize}
\begin{table*}[!t]
  \centering
  \resizebox{1\textwidth}{!}{
    \begin{tabular}{c|l l|c|c c c c c}
    \toprule
    \textbf{Model Type} & \textbf{Methods} & \textbf{Parameters} & \textbf{Overall}$\uparrow$ & \textbf{Text\textsuperscript{Edit}}$\downarrow$ & \textbf{Formula\textsuperscript{CDM}}$\uparrow$ & \textbf{Table\textsuperscript{TEDS}}$\uparrow$ & \textbf{Table\textsuperscript{TEDS-S}}$\uparrow$ & \textbf{Read Order\textsuperscript{Edit}}$\downarrow$ \\
    \midrule
    \multirow{4}{*}{\makecell{\textbf{Pipeline}\\\textbf{Tools}}} &
    Marker-1.8.2~\cite{vik2024marker} & - & 71.30  & 0.206  & 76.66  & 57.88  & 71.17  & 0.250  \\
    & MinerU2-pipeline~\cite{wang2024mineru} & - & 75.51  & 0.209  & 76.55  & 70.90  & 79.11  & 0.225  \\
    & PP-StructureV3~\cite{cui2025paddleocr} & - & 86.73  & 0.073  & 85.79  & 81.68  & 89.48  & 0.073  \\
    \midrule
    \multirow{5}{*}{\makecell{\textbf{General}\\\textbf{VLMs}}}
    & GPT-4o~\cite{achiam2023gpt}  & - & 75.02  & 0.217  & 79.70  & 67.07  & 76.09  & 0.148  \\
    & InternVL3-76B~\cite{zhu2025internvl3} & 76B & 80.33  & 0.131  & 83.42  & 70.64  & 77.74  & 0.113  \\
    & InternVL3.5-241B~\cite{wang2025internvl3} & 241B & 82.67  & 0.142  & 87.23  & 75.00  & 81.28  & 0.125  \\
    & Qwen2.5-VL-72B~\cite{bai2025qwen2} & 72B & 87.02  & 0.094  & \underline{88.27}  & 82.15  & 86.22  & 0.102  \\
    & Gemini-2.5 Pro~\cite{comanici2025gemini} & - & 88.03  & 0.075  & 85.82  & 85.71  & 90.29  & 0.097  \\
    \midrule
    \multirow{12}{*}{\makecell{\textbf{Specialized}\\\textbf{VLMs}}}
    & Dolphin~\cite{feng2025dolphin} & 322M & 74.67  & 0.125  & 67.85  & 68.70  & 77.77  & 0.124  \\
    & OCRFlux~\cite{OCRFlux2025} & 3B & 74.82  & 0.193  & 68.03  & 75.75  & 80.23  & 0.202  \\
    & Mistral-OCR~\cite{mistral2025} & - & 78.83  & 0.164  & 82.84  & 70.03  & 78.04  & 0.144  \\
    & POINTS-Reader~\cite{liu2025points} & 3B & 80.98 & 0.134 & 79.20 & 77.13 & 81.66 & 0.145 \\
    & olmOCR-7B~\cite{poznanski2025olmocr} & 7B  & 81.79  & 0.096  & 86.04  & 68.92  & 74.77  & 0.121 \\
    & MinerU2-VLM\cite{wang2024mineru} & 0.9B & 85.56  & 0.078  & 80.95  & 83.54  & 87.66  & 0.086  \\
    & Nanonets-OCR-s~\cite{nanonets2025} & 3.7B & 85.59  & 0.093  & 85.90  & 80.14  & 85.57  & 0.108  \\
    & MonkeyOCR-pro-1.2B~\cite{li2025monkeyocr} & 1.9B & 86.96  & 0.084  & 85.02  & 84.24  & 89.02  & 0.130  \\
    & MonkeyOCR-3B~\cite{li2025monkeyocr} & 3.7B & 87.13 & 0.075 & 87.45 & 81.39 & 85.92 & 0.129 \\ 
    & dots.ocr~\cite{dots.ocr} & 3B & 88.41  & \underline{0.048}  & 83.22  & \underline{86.78}  & 90.62  & \underline{0.053}  \\
    & MonkeyOCR-pro-3B~\cite{li2025monkeyocr} & 3.7B & \underline{88.85} & 0.075 & 87.25 & \underline{86.78} & \underline{90.63} & 0.128 \\
    \rowcolor{gray!20}
    & \textbf{MinerU2.5} & 1.2B & \textbf{90.67}  & \textbf{0.047}  & \textbf{88.46}  & \textbf{88.22}  & \textbf{92.38}  & \textbf{0.044}  \\
    \bottomrule
    \end{tabular}%
  }
  \vspace{-4pt}
  \caption{Performance comparison of document parsing methods on OmniDocBench across text, formula, table, and reading order extraction tasks.}
  \label{tab:OmniDocbench-results}%
  \vspace{-4pt}
\end{table*}

\begin{table*}[!t]
  \small
  \centering
  \resizebox{1\textwidth}{!}{
    \begin{tabular}{c|l|ccccccccc}
    \toprule
  \textbf{Model Type} & \textbf{Models} & \textbf{Slides} & \makecell{\textbf{Academic}\\\textbf{Papers}} & \textbf{Book} & \textbf{Textbook} & \makecell{\textbf{Exam}\\\textbf{Papers}} & \textbf{Magazine} & \textbf{Newspaper} & \textbf{Notes} & \makecell{\textbf{Financial}\\\textbf{Report}} \\
     \midrule
     \multirow{4}{*}{\makecell{\textbf{Pipeline}\\\textbf{Tools}}}  
    & Marker-1.8.2~\cite{vik2024marker} & 0.1796 & 0.0412 & 0.1010 & 0.2908 & 0.2958 & 0.1111 & 0.2717 & 0.4656 & 0.0341 \\
    & MinerU2-pipeline~\cite{wang2024mineru} & 0.4244 & 0.0230 & 0.2628 & 0.1224 & 0.0822 & 0.3950 & 0.0736 & 0.2603 & 0.0411 \\
    & PP-StructureV3~\cite{cui2025paddleocr}  & 0.0794 & 0.0236 & 0.0415 & 0.1107 & 0.0945 & 0.0722 & \underline{0.0617} & 0.1236 & 0.0181 \\
    \midrule
    \multirow{5}{*}{\makecell{\textbf{General}\\\textbf{VLMs}}} 
    & GPT-4o~\cite{achiam2023gpt}  & 0.1019 & 0.1203 & 0.1288 & 0.1599 & 0.1939 & 0.1420 & 0.6254 & 0.2611 & 0.3343 \\
    & InternVL3-76B~\cite{zhu2025internvl3}  & 0.0349 & 0.1052 & 0.0629 & 0.0827 & 0.1007 & 0.0406 & 0.5826 & \underline{0.0924} & 0.0665 \\
    & InternVL3.5-241B~\cite{wang2025internvl3}  & 0.0475 & 0.0857 & \textbf{0.0237} & 0.1061 & 0.0933 & 0.0577 & 0.6403 & 0.1357 & 0.1117 \\
    & Qwen2.5-VL-72B~\cite{bai2025qwen2}  & 0.0422 & 0.0801 & 0.0586 & 0.1146 & \underline{0.0681} & 0.0964 & 0.2380 & 0.1232 & 0.0264 \\
    & Gemini-2.5 Pro~\cite{comanici2025gemini}  & 0.0326 & \underline{0.0182} & 0.0694 & 0.1618 & 0.0937 & \textbf{0.0161} & 0.1347 & 0.1169 & 0.0169 \\  
    \midrule
    \multirow{13}{*}{\makecell{\textbf{Specialized}\\\textbf{VLMs}}}
    & Dolphin~\cite{feng2025dolphin}  & 0.0957 & 0.0453 & 0.0616 & 0.1333 & 0.1684 & 0.0702 & 0.2388 & 0.2561 & 0.0186 \\
    & OCRFlux~\cite{OCRFlux2025}  & 0.0870 & 0.0867 & 0.0818 & 0.1843 & 0.2072 & 0.1048 & 0.7304 & 0.1567 & 0.0193 \\
    & Mistral-OCR~\cite{mistral2025}  & 0.0917 & 0.0531 & 0.0610 & 0.1349 & 0.1341 & 0.0581 & 0.5643 & 0.3097 & 0.0523 \\
    & POINTS-Reader~\cite{liu2025points}  & 0.0334 & 0.0779 & 0.0671 & 0.1372 & 0.1901 & 0.1343 & 0.3789 & 0.0937 & 0.0951 \\  
    & olmOCR-7B~\cite{poznanski2025olmocr}  & 0.0497 & 0.0365 & 0.0539 & 0.1204 & 0.0728 & 0.0697 & 0.2916 & 0.1220 & 0.0459 \\
    & MinerU2-VLM\cite{wang2024mineru}  & 0.0745 & \textbf{0.0104} & 0.0357 & 0.1276 & 0.0698 & 0.0652 & 0.1831 & \textbf{0.0803} & 0.0236 \\
    & Nanonets-OCR-s~\cite{nanonets2025}  & 0.0551 & 0.0578 & 0.0606 & 0.0931 & 0.0834 & 0.0917 & 0.1965 & 0.1606 & 0.0395 \\
    & MonkeyOCR-pro-1.2B~\cite{li2025monkeyocr}  & 0.0961 & 0.0354 & 0.0530 & 0.1110 & 0.0887 & 0.0494 & 0.0995 & 0.1686 & 0.0198 \\ 
    & MonkeyOCR-3B~\cite{li2025monkeyocr}  & 0.0904 & 0.0362 & 0.0489 & 0.1072 & 0.0745 & 0.0475 & 0.0962 & 0.1165 & 0.0196 \\
    & dots.ocr~\cite{dots.ocr}  & \textbf{0.0290} & 0.0231 & 0.0433 & \underline{0.0788} & \textbf{0.0467} & \underline{0.0221} & 0.0667 & 0.1116 & \textbf{0.0076} \\
    & MonkeyOCR-pro-3B~\cite{li2025monkeyocr}  & 0.0879 & 0.0459 & 0.0517 & 0.1067 & 0.0726 & 0.0482 & 0.0937 & 0.1141 & 0.0211 \\
    \rowcolor{gray!20}
     & \textbf{MinerU2.5}  & \underline{0.0294} & 0.0235 & \underline{0.0332} & \textbf{0.0499} & \underline{0.0681} & 0.0316 & \textbf{0.0540} & 0.1161 & \underline{0.0104} \\
    \bottomrule
    \end{tabular}%
  }
    \vspace{-2pt}
    \caption{Document Parsing Performance in Text Edit Distance on OmniDocBench: evaluation using edit distance across 9 PDF page types.}    
    \vspace{-10pt}
  \label{tab:OmniDocbench-results-2}%
\end{table*}%

\vspace{-0.2cm}
\section{Evaluation}
\label{section:eval}
In this section, we present a comprehensive quantitative evaluation of \mineru{} to demonstrate its effectiveness in document parsing tasks. Specifically, we compare \mineru{} against leading general-purpose VLMs including GPT-4o \citep{achiam2023gpt}, Gemini-2.5 Pro \citep{comanici2025gemini}, and Qwen2.5-VL \citep{bai2025qwen2}, as well as state-of-the-art domain-specific VLMs such as dots.ocr \citep{dots.ocr}, MonkeyOCR \citep{li2025monkeyocr}, and olmOCR \citep{poznanski2025olmocr}. The evaluation is organized into two parts: \Cref{sec:eval_end2end} presents full-document parsing results across multiple benchmarks, while \Cref{sec:eval_single} focuses on element-specific capabilities including layout analysis, formula recognition, and table recognition.

\begin{table*}[!t]
    \centering
    \setlength{\tabcolsep}{1.2mm}{
    \footnotesize
    \begin{tabular}{lcccccccccccc}
        \toprule
        \multirow{2}{*}{\textbf{Model}}   & \multicolumn{2}{c}{Edit Distance $\downarrow$}  & \multicolumn{2}{c}{F1-score $\uparrow$} & \multicolumn{2}{c}{Precision$\uparrow$} & \multicolumn{2}{c}{Recall$\uparrow$} & \multicolumn{2}{c}{BLEU$\uparrow$} & \multicolumn{2}{c}{METEOR$\uparrow$} \\
        \cmidrule(r){2-3} \cmidrule(r){4-5} \cmidrule(r){6-7} \cmidrule(r){8-9} \cmidrule(r){10-11} \cmidrule(r){12-13}  
        & en & zh & en & zh & en & zh & en & zh & en & zh & en & zh\\
        \midrule
        Mathpix~\cite{Mathpix2025} & 0.064  & 0.223 & 0.930  & 0.919  & \textbf{0.950}  & 0.952  & 0.911  & 0.889  & 0.901  & 0.593  & 0.924  & 0.768 \\ 
        PP-StructureV3~\cite{cui2025paddleocr} & 0.068  & 0.210 & 0.871  & 0.929  & 0.856  & 0.924  & 0.892  & 0.935  & 0.796  & 0.570  & 0.902  & 0.802 \\ 
        MinerU2-pipeline~\cite{wang2024mineru} & 0.099  & 0.225 & 0.663  & 0.919  & 0.635  & 0.908  & 0.703  & 0.934  & 0.504  & 0.571  & 0.670  & 0.810  \\
        PaddleOCR~\cite{cui2025paddleocr} & 0.323 & 0.649 & 0.707 & 0.864 & 0.690 & 0.912 & 0.730 & 0.821 & 0.517 & 0.537 & 0.674 & 0.699 \\
        \midrule
        Gemini-2.5 Pro~\cite{comanici2025gemini} & 0.080  & 0.204 & 0.922  & 0.927  & 0.940  & 0.959  & 0.906  & 0.898  & 0.877  & 0.690  & 0.921  & 0.862  \\
        GPT-4o~\cite{achiam2023gpt} & 0.085  & 0.450 & 0.919  & 0.686  & 0.929  & 0.694  & 0.910  & 0.703  & 0.870  & 0.354  & 0.922  & 0.495  \\ 
        Qwen2.5-VL-72B~\cite{bai2025qwen2} & 0.093  & 0.140 & 0.923  & 0.940  & 0.936  & 0.956  & 0.912  & 0.926  & 0.879  & 0.798  & 0.924  & 0.876   \\ 
        InternVL3-76B~\cite{zhu2025internvl3} & 0.125  & 0.282 & 0.828  & 0.871  & 0.842  & 0.889  & 0.817  & 0.856  & 0.728  & 0.527  & 0.829  & 0.759 \\
        Qwen2-VL-7B~\cite{wang2024qwen2} & 0.165 & 0.270 & 0.849 & 0.883 & 0.834 & 0.847 & 0.873 & 0.942 & 0.795 & 0.578 & 0.859 & 0.763 \\
        MiniCPM-V2.6-8B~\cite{yao2024minicpm} & 0.244 & 0.437 & 0.804 & 0.778 & 0.793 & 0.721 & 0.837 & 0.875 & 0.695 & 0.431 & 0.640 & 0.642 \\
        \midrule
        MinerU2-VLM~\cite{wang2024mineru} & \underline{0.048}  & 0.182 & 0.936  & 0.941  & 0.926  & 0.927  & \underline{0.947}  & 0.958  & 0.893  & 0.611  & \textbf{0.950}  & 0.837   \\
        Ocean-OCR~\cite{chen2025ocean} & 0.057 & \textbf{0.062} & \underline{0.937} & \underline{0.962} & 0.932 & \underline{0.956} & \textbf{0.956} & \textbf{0.974} & \underline{0.906} & \textbf{0.912} & \underline{0.945} & \textbf{0.916} \\
        MonkeyOCR-pro-1.2B~\cite{li2025monkeyocr} & 0.064  & 0.190 & 0.929  & 0.934  & 0.918  & 0.925  & 0.944  & 0.948  & 0.884  & 0.699  & 0.941  & 0.850  \\
        SmolDocling~\cite{nassar2025smoldocling} & 0.080  & 0.878 & 0.899  & 0.157  & 0.895  & 0.140  & 0.912  & 0.268  & 0.839  & 0.048  & 0.907  & 0.151 \\
        dots.ocr~\cite{dots.ocr} & 0.083  & 0.179 & 0.904  & 0.931  & 0.920  & 0.951  & 0.890  & 0.913  & 0.849  & 0.639  & 0.911  & 0.842 \\ 
        GOT\cite{wei2024general} & 0.084 & 0.117 & 0.895 & 0.928 & 0.891 & 0.934 & 0.906 & 0.929 &0.835 & 0.805 & 0.874 & 0.848\\
        \rowcolor{gray!20}
        \textbf{MinerU2.5} & \textbf{0.033}  & \underline{0.082} & \textbf{0.945} & \textbf{0.965} & \underline{0.948} & \textbf{0.966} & 0.942 & \underline{0.964} & \textbf{0.909} & \underline{0.817} & \textbf{0.950} & \underline{0.887}   \\
        \bottomrule
    \end{tabular}
}
    \caption{Evaluation results on Ocean-OCR bench on dense English (en) and Chinese (zh) OCR for document-level pages. Some model results are sourced from the OceanOCR official reports.}
    \label{tab:OceanOCR}
\end{table*}

\begin{table}[!t]
\centering
\small
\begin{tabular}{lccccccccc}
\toprule
\textbf{Model} & {Overall} & {AR} & {OSM} & {TA} & {OS} & {HF} & {MC} & {LTT} & {Base}  \\
\midrule
    MinerU2-pipeline\cite{wang2024mineru} & 55.6 & 61.8 & 13.5 & 60.9 & 17.3 & \textbf{96.6} & 59.0 & 39.1 & 96.6  \\
    Nanonets-OCR-s\cite{nanonets2025} & 60.7 & 63.9 & 41.0 & 77.7 & 39.5 & 40.7 & 69.9 & 53.4 & \underline{99.3}  \\
    GPT-4o\cite{achiam2023gpt} & 63.2 & 44.1 & 37.6 & 69.1 & 40.9 & 94.2 & 68.9 & 54.1 & 96.7  \\
    MonkeyOCR-pro-1.2B\cite{li2025monkeyocr} & 64.3 & 65.4 & 26.9 & 60.3 & 31.2 & 93.3 & 66.2 & \underline{81.7} & 89.5  \\
    Qwen2.5-VL-72B\cite{bai2025qwen2} & 64.8 & \underline{72.2} & \underline{51.1} & 67.3 & 38.6 & 73.6 & 68.3 & 49.1 & 98.3  \\ 
    MonkeyOCR-pro-3B\cite{li2025monkeyocr} & 68.8 & 67.7 & 28.4 & 74.6 & 36.1 & 91.2 & 76.6 & 80.1 & 95.3  \\
    olmOCR\cite{poznanski2025olmocr} & 71.8 & 63.9 & 41.0 & 72.9 & \textbf{43.9} & \underline{95.1} & 77.3 & 81.2 & 98.9  \\
    dots.ocr\cite{dots.ocr} & \underline{73.6} & 66.3 & 35.8 & \textbf{88.3} & \underline{40.9} & 94.1 & \textbf{82.4} & 81.2 & \textbf{99.5} \\
    \rowcolor{gray!20}
    \textbf{MinerU2.5} & \textbf{75.2} & \textbf{76.6} & \textbf{54.6} & \underline{84.9} & 33.7 & \textbf{96.6} & \underline{78.2} & \textbf{83.5} & 93.7  \\
    \bottomrule
    \end{tabular}
\caption{Evaluation results on olmOCR-bench grouped by document types, including arXiv Math(AR), Old Scans Math (OSM), Tables (TA), Old Scans (OS), Headers Footers (HF), Multi Column (MC) and Long Tiny Text (LTT). Results on AR and OSM are replaced with ExpRate, and other results are sourced from the official reports of olmOCR-bench and dots.ocr. The Overall Score (Overall) represents the average across all document types.}
\label{tab:olmocr-bench-results}
\end{table}

\subsection{Full-Document Parsing Task}
\label{sec:eval_end2end}
We evaluate \mineru{}'s full document parsing performance on three prominent benchmarks: OmniDocBench~\cite{ouyang2025omnidocbench}, Ocean-OCR~\cite{chen2025ocean} benchmarks, and olmOCR-bench~\cite{poznanski2025olmocr}. These benchmarks provide comprehensive evaluation from different dimensions, covering diverse document types, various quality conditions, and different parsing challenges to thoroughly assess the model's robustness and generalization capabilities.
\begin{itemize}
\item \textbf{OmniDocBench}~\cite{ouyang2025omnidocbench}: This evaluation dataset is designed for diverse document parsing in real-world scenarios, encompassing nine document types, four layout types, and three language types. It offers a comprehensive assessment of parsing scores for text, formulas, tables, and reading order in full-document parsing, as well as for element-specific parsing tasks.
\item \textbf{olmOCR-bench}~\cite{poznanski2025olmocr}: This evaluation dataset comprises 1,402 PDF documents sourced from various repositories, organized into seven subsets. Certain test patterns are applicable across all document types (e.g., presence, absence, reading order), while others are specifically targeted at challenging yet crucial content extraction objectives (e.g., tables, mathematical formulas).
\item \textbf{Ocean-OCR benchmark} ~\cite{chen2025ocean}: This evaluation dataset consists of 100 images from English papers and 100 images from Chinese papers. It primarily evaluates the ability of text parsing and employs several text OCR-related evaluation metrics, such as Normalized Edit Distance, F1 Score, Precision, Recall, BLEU, and METEOR.
\end{itemize}

\subsubsection{Evaluation Details and Metrics}


For OmniDocBench \citep{ouyang2025omnidocbench}, we evaluate on the latest version with three key improvements:
\begin{itemize}
\item Enhanced resolution for Notes and Newspapers from 72 to 200 DPI, enabling more accurate evaluation of fine-grained text and handwritten content.
\item An addition of 374 pages to balance Chinese-English content distribution and enrich mathematical formula coverage. Currently, it contains a total of 1,355 pages.
\item Evaluation methodology updated to hybrid matching algorithm.
\end{itemize}

The Overall score combines three core metrics:

$$\textbf{Overall} = \frac{(1-{\mathbf{Text^{Edit}}}) \times 100 + {\mathbf{Table^{TEDS}}} +{\mathbf{Formula^{CDM}}}}{3}$$

For olmOCR-bench~\cite{poznanski2025olmocr}, we replace the formula scores of Arxiv Math (AR) and Old Scans Math (OSM) with the more reliable ExpRate of CDM~\cite{wang2025image}. The original evaluation compares LaTeX formulas by parsing them into abstract syntax trees and matching Unicode tokens, which is overly sensitive to syntax variations (e.g., $\backslash$\textit{cdots} vs. $\backslash$\textit{dotsb}) that render identically but are scored as different. To avoid this bias, we adopt ExpRate, which directly compares rendered outputs, assigning 1 for exact matches and 0 otherwise. 

\subsubsection{Evaluation Results}
\mineru{} demonstrates exceptional performance across all benchmarks, achieving state-of-the-art results in most metrics (\Cref{tab:OmniDocbench-results,tab:OmniDocbench-results-2,tab:olmocr-bench-results,tab:OceanOCR}).

As shown in \Cref{tab:OmniDocbench-results}, \mineru{} achieves an overall score of 90.67 on OmniDocBench, outperforming the second-best model MonkeyOCR-pro-3B~\cite{li2025monkeyocr} by 1.82 and dots.ocr \citep{dots.ocr} by 2.26 points. In text recognition tasks, \mineru{} achieves the lowest edit distance of 0.047, marginally better than dots.ocr at 0.048 and significantly outperforming PP-StructureV3~\cite{cui2025paddleocr}, which scores 0.073. For formula recognition, \mineru{} leads with a CDM score of 88.46, exceeding both Qwen2.5-VL-72B at 88.27 and MonkeyOCR-3B at 87.45. In table recognition tasks, \mineru{} achieves the highest TEDS score of 88.22 and TEDS-S score of 92.38. For reading order evaluation, it maintains the best edit distance of 0.044. The document-type specific results presented in \Cref{tab:OmniDocbench-results-2} demonstrate that \mineru{} achieves best or second-best performance in 6 out of 9 categories. For textbooks, it delivers the best performance with an edit distance of 0.0499, substantially outperforming dots.ocr's 0.0788. For newspapers, \mineru{} leads with a score of 0.0540, surpassing all competing models. In both financial reports and slides categories, \mineru{} achieves second-best performance with scores of 0.0104 and 0.0294 respectively.

For the results of the Ocean-OCR benchmark presented in \Cref{tab:OceanOCR}, \mineru{} demonstrates exceptional performance in dense OCR tasks. On English documents, it achieves the lowest edit distance of 0.033 and the highest F1-score of 0.945, accompanied by best-in-class BLEU and METEOR scores of 0.909 and 0.950 respectively. For Chinese documents, \mineru{} achieves the highest F1-score of 0.965 and Precision of 0.966, while maintaining strong BLEU and METEOR scores of 0.817 and 0.887 respectively.

The results of olmOCR-bench are shown in \Cref{tab:olmocr-bench-results}, where \mineru{} achieves an overall score of 75.2, surpassing dots.ocr's 73.6 by 1.6 points. In the arXiv Math category, it leads with a score of 76.6, outperforming Qwen2.5-VL-72B~\cite{bai2025qwen2}'s 72.2 by 4.4 points. For Old Scans Math, \mineru{} dominates with a score of 54.6, exceeding all other evaluated models. In the Long Tiny Text category, it achieves 83.5, surpassing MonkeyOCR-pro-1.2B~\cite{li2025monkeyocr} which scores 81.7.

\begin{table}[!t]
\setlength{\tabcolsep}{4pt}
\resizebox{\linewidth}{!}{
\begin{tabular}{lcccccccccccccccccc}
\toprule[.9pt]
\multirow{2}{*}{Method} &
\multicolumn{3}{c}{Textual} &
\multicolumn{3}{c}{Image} &
\multicolumn{3}{c}{Table} &
\multicolumn{3}{c}{Equation} &
\multicolumn{3}{c}{Page Margins} &
\multicolumn{3}{c}{Full Page} \\
\cmidrule(rl){2-4} \cmidrule(rl){5-7} \cmidrule(rl){8-10} \cmidrule(rl){11-13} \cmidrule(rl){14-16} \cmidrule(rl){17-19}
 & P$\uparrow$  & R$\uparrow$ & F1$\uparrow$
 & P$\uparrow$  & R$\uparrow$ & F1$\uparrow$
 & P$\uparrow$  & R$\uparrow$ & F1$\uparrow$
 & P$\uparrow$  & R$\uparrow$ & F1$\uparrow$
 & P$\uparrow$  & R$\uparrow$ & F1$\uparrow$
 & P$\uparrow$  & R$\uparrow$ & F1$\uparrow$
 \\
\midrule
\multicolumn{19}{c}{OmniDocBench~\cite{ouyang2025omnidocbench}} \\
\midrule
LayoutLMv3~\cite{huang2022layoutlmv3} & 90.4  & 48.2  & 58.1  & 72.1  & 51.2  & 57.2  & 72.6  & 55.1  & 61.0  & - & 36.9  & - & - & - & - & - & - & -\\
MinerU2-VLM~\cite{wang2024mineru} & 90.3  & 95.6  & 91.9  & 87.2  & 91.0  & 90.9  & 96.0  & 97.1  & 97.8  & 87.4  & 95.8  & 90.5  & - & -  & - & - & - & -\\
DocLayout-YOLO~\cite{zhao2024doclayout} & 95.4  & \textbf{98.3}  & 96.5  & \underline{87.6}  & \textbf{96.7}  & \underline{94.7}  & 94.9  & \underline{98.1}  & \underline{98.4}  & \underline{95.3}  & 90.6  & 93.8  & - & \textbf{98.7}  & - & 92.3  & \textbf{97.7}  & 94.1 \\
PP-StructureV3~\cite{cui2025paddleocr} & \underline{96.8}  & 96.7  & \underline{96.6}  & 86.4  & 92.1  & 92.9  & \textbf{96.6}  & 97.4  & 98.2  & \textbf{96.5}  & \underline{97.6}  & \textbf{96.7}  & \textbf{92.9}  & 86.2  & \underline{88.1}  & \underline{94.8}  & 96.2  & \underline{94.6} \\
\rowcolor{gray!20}
MinerU2.5 & \textbf{97.2}  & \underline{98.0}  & \textbf{97.5}  & \textbf{89.6}  & \underline{94.3}  & \textbf{95.0}  & \underline{96.0}  & \textbf{98.1}  & \textbf{98.4}  & 92.4  & \textbf{99.6}  & \underline{94.7}  & \underline{89.9}  & \underline{95.4}  & \textbf{91.4}  & \textbf{95.8}  & \underline{97.0}  & \textbf{95.9} \\

\midrule
\multicolumn{19}{c}{D\textsuperscript{4}LA~\cite{da2023vision}} \\
\midrule
LayoutLMv3~\cite{huang2022layoutlmv3} &86.9  & 41.2  & 52.4  & \textbf{59.3}  & 32.0  & 31.4  & 59.3  & 41.8  & 43.3  & - & 50.5  & - & - & - & - & - & - & -\\
MinerU2-VLM~\cite{wang2024mineru} & 88.3  & 88.9  & 87.9  & \underline{56.7}  & 35.0  & 38.1  & \underline{89.1}  & \underline{84.1}  & \underline{90.6}  & \underline{38.3}  & \underline{99.4}  & 79.1  & - & - & - & - & - & -\\
DocLayout-YOLO~\cite{zhao2024doclayout} & 86.3  & \underline{97.8}  & \underline{90.8}  & 41.5  & \underline{92.9}  & 62.6  & 87.6  & \textbf{89.0}  & 89.8  & 31.9  & 80.2  & \textbf{91.1}  & - & \underline{95.0}  & - & 82.6  & \textbf{95.4}  & \underline{87.3} \\
PP-StructureV3~\cite{cui2025paddleocr} & \underline{88.5}  & 93.5  & 90.0  & 50.1  & 82.3  & \underline{67.9}  & 87.1  & 81.1  & 89.7  & 24.6  & 85.9  & \underline{92.1}  & \textbf{76.8}  & 84.2  & \underline{79.1}  & \underline{85.7}  & 91.0  & 86.0 \\
\rowcolor{gray!20}
MinerU2.5 & \textbf{91.8}  & \textbf{98.3}  & \textbf{94.6}  & 53.8  & \textbf{94.3}  & \textbf{72.8}  & \textbf{91.9}  & 78.9  & \textbf{91.4}  & \textbf{46.0}  & \textbf{100.0}  & 91.0  & \underline{75.9}  & \textbf{97.6}  & \textbf{84.2}  & \textbf{90.4}  & \underline{92.5}  & \textbf{90.2} \\

\midrule
\multicolumn{19}{c}{DocLaynet~\cite{pfitzmann2022doclaynet}} \\
\midrule
LayoutLMv3~\cite{huang2022layoutlmv3} &
88.8  & 59.3  & 67.9  & 79.0  & 50.3  & 61.9  & 75.2  & 54.9  & 61.8  & - & 31.9  & - & - & - & - & - & - & -\\
MinerU2-VLM~\cite{wang2024mineru} &
88.1  & 96.1  & 91.7  & 85.5  & 78.1  & 91.3  & 94.9  & \underline{94.4}  & 95.6  & 83.9  & \underline{97.0}  & 90.0  & - & - & - & - & - & -\\
DocLayout-YOLO~\cite{zhao2024doclayout} &
86.9  & 96.8  & 91.2  & 85.8  & \underline{96.2}  & 91.3  & 92.0  & \textbf{95.7}  & 94.8  & 80.5  & 86.9  & 82.8  & - & \underline{97.7}  & - & 88.0  & \underline{96.3}  & 90.9 \\
PP-StructureV3~\cite{cui2025paddleocr} &
\textbf{90.9}  & \underline{97.3}  & \underline{93.8}  & \underline{91.7}  & 90.4  & \underline{94.2}  & \textbf{96.4}  & 93.7  & \underline{96.7}  & \underline{88.8}  & 96.0  & \underline{92.1}  & \textbf{76.8}  & 79.3  & \textbf{77.4}  & \underline{92.4}  & 95.7  & \underline{93.0} \\
\rowcolor{gray!20}
MinerU2.5 &
\underline{90.2}  & \textbf{99.6}  & \textbf{94.8}  & \textbf{92.5}  & \textbf{96.3}  & \textbf{95.9}  & \underline{96.3}  & 93.5  & \textbf{97.1}  & \textbf{88.9}  & \textbf{98.6}  & \textbf{93.5}  & \underline{76.3}  & \textbf{98.9}  & \textbf{86.3}  & \textbf{92.8}  & \textbf{97.7}  & \textbf{94.6} \\

\bottomrule[.9pt]
\end{tabular}
}
\vspace{3mm}
\caption{Comparison of layout analysis performance (Precision@PageIoU, Recall@PageIoU, F1-score@PageIoU) across different methods and content types on multiple layout analysis benchmarks.}
\label{tab:layout_detection_task}
\end{table}
\subsection{Element-Specific Parsing Task}
\label{sec:eval_single}

\subsubsection{Layout Analysis}
We validate the effectiveness of our layout analysis by performing a fair, zero-shot comparison with leading methods on three publicly available datasets:
\begin{itemize}
\item \textbf{OmniDocBench} \cite{ouyang2025omnidocbench}: A recent benchmark for document parsing that includes detailed layout annotations.
\item \textbf{D\textsuperscript{4}LA} \cite{da2023vision}: Contains 11,092 noisy document images annotated with 27 categories, split into 8,868 training and 2,224 test images. We use its test set with annotations for evaluation.
\item \textbf{DocLayNet} \cite{pfitzmann2022doclaynet}: A large-scale dataset of 80,863 pages from 7 document types, manually annotated with 11 categories. We use its validation set with annotations for evaluation.
\end{itemize}

We compare our \mineru{} with several recent methods, including LayoutLMv3~\cite{huang2022layoutlmv3}, MinerU2-VLM~\cite{wang2024mineru}, DocLayout-YOLO~\cite{zhao2024doclayout} and PP-StructureV3~\cite{cui2025paddleocr}. For a equitable assessment, we evaluate all models without dataset-specific training. To account for differences in detection granularity and category definitions, we unified the evaluation by mapping all labels to five broad categories and using the PageIoU metric, which assesses the spatial overlap without considering category labels for the ``Full Page'' score.

The results in \Cref{tab:layout_detection_task} show that \mineru{} significantly outperforms other models, achieving the top Full Page F1-score@PageIoU across all benchmarks. It also secures leading F1-scores@PageIoU for the majority of individual element types. This consistent superiority confirms that the PageIoU metric provides a robust basis for comparison, effectively capturing model performances independent of annotation inconsistencies.

\begin{table}[!t]
\setlength{\tabcolsep}{4pt}
\resizebox{\linewidth}{!}{
\begin{tabular}{lcccccccccc}
\toprule[.9pt]
\multirow{2}{*}{Method} &
\multicolumn{2}{c}{PubTabNet} &
\multicolumn{2}{c}{FinTabNet} &
\multicolumn{2}{c}{CC-OCR} &
\multicolumn{2}{c}{OCRBench v2} &
\multicolumn{2}{c}{In-house TR Benchmark} \\
\cmidrule(rl){2-3} \cmidrule(rl){4-5} \cmidrule(rl){6-7} \cmidrule(rl){8-9} \cmidrule(rl){10-11}
 & TEDS$\uparrow$  & TEDS-S$\uparrow$ & TEDS$\uparrow$  & TEDS-S$\uparrow$ & TEDS$\uparrow$  & TEDS-S$\uparrow$  & TEDS$\uparrow$   & TEDS-S$\uparrow$  & TEDS$\uparrow$ & TEDS-S$\uparrow$   \\
\midrule
RapidTable~\cite{rapidtable}        & 86.57     & \textbf{96.43}      & 73.77     & 84.84      & 50.93          & 65.84          & 65.55          & 77.73          & 51.96          & 71.94  \\
MiniCPM-V 4.5~\cite{yu2025minicpmv45cookingefficient}      & 80.30     & 87.67      & \underline{85.41}     & \underline{89.18}      & 68.49          & 77.55          & 80.28          & 85.65          & 55.47          & 69.61    \\
InternVL3.5-241B~\cite{wang2025internvl3}  & 83.75     & 88.76      & 84.74     & 87.92      & 62.87          & 69.52          & 79.5           & 85.81          & 56.32          & 69.3      \\
Qwen2.5-VL-7B~\cite{bai2025qwen2}     & 81.60 & 86.78  & 82.58 & 87.46  & 78.29          & 84.26          & 77.44          & 84.71          & 57.34          & 73.17          \\
Qwen2.5-VL-72B~\cite{bai2025qwen2}    & 84.39 & 87.91  & 82.90 & 87.13  & \underline{81.22}          & \underline{86.48}    & 81.33          & 86.58          & 62.79          & 76.91     \\
GPT-4o~\cite{achiam2023gpt}            & 76.53 & 86.16  & 83.94 & 87.00  & 66.98          & 79.04          & 70.51          & 79.55          & 46.99          & 70.29          \\
Gemini-2.5 Pro~\cite{comanici2025gemini}    & -     & -      & -     & -      & \textbf{85.56} & \textbf{90.07} & \textbf{88.94} & \underline{89.47}    & \underline{69.72}    & \underline{81.29}   \\
dots.ocr~\cite{dots.ocr}          & \textbf{90.65}  & \underline{93.76}     & 84.12     & 87.86      & 75.42         & 81.65        & 82.04        & 86.27        & 66.91       & 79.27      \\
Nanonets-OCR-s~\cite{nanonets2025}      & 63.58     & 75.68      & 68.06     & 73.6       & 66.15          & 71.33          & 69.66          & 76.28          & 54.35          & 66.12       \\
MinerU2-VLM~\cite{wang2024mineru}       & 88.11 & 90.85  & 78.49 & 83.03  & 64.61 & 71.8     & 73.22  & 78.24 & 63.54 & 76.66 \\
\rowcolor{gray!20}
MinerU2.5     & \underline{89.07} & 93.11  & \textbf{95.97} & \textbf{97.61}  & 79.76 & 85.16     & \underline{87.13}  & \textbf{90.62} & \textbf{71.48} & \textbf{82.83} \\
\bottomrule[.9pt]
\end{tabular}
}
\vspace{3mm}
\caption{Table Recognition Performance. \mineru{} achieves SOTA performance on most benchmarks among TEDS and TEDS-S metrics, and the remaining ones are also generally competitive with the SOTA. (CCOCR and OCRBench v2 are OCR evaluation benchmarks, we only select the subsets that contain tables. PubTabNet and FinTabNet have a large number of images, so we have not evaluate Gemini-2.5 Pro on them.).}
\label{tab:table_recognition_task}
\end{table}

\subsubsection{Table Recognition}
We evaluate representative methods, covering traditional table recognition methods, general multimodal large models and document parsing models, on five table recognition benchmarks as shown in \Cref{tab:table_recognition_task}. Below is an introduction to each benchmark:
\begin{itemize}
    \item \textbf{PubTabNet} ~\cite{zhong2020image} is the first large-scale table recognition dataset that provides annotations (in HTML format) of table images, captured from scientific articles. PubTabNet contains 9k tables in its test set.
    \item \textbf{FinTabNet} ~\cite{zheng2021global} is a dataset containing tables from the annual reports of 500 companies. The major challenge of this benchmark is that financial tables largely differ from scientific and government document tables in that the former has fewer graphical lines, larger gaps within each table, and more color variations. FinTabNet contains 10k tables in its test set.
    \item \textbf{CC-OCR} ~\cite{yang2024cc} and \textbf{OCRBench v2} ~\cite{fu2024ocrbench} are both designed to evaluate the OCR capabilities of multimodal large models and contain several OCR tasks. We only retain the data related to document recognition and those images that include tables. After filtering, CC-OCR remains 300 images and OCRBench v2 remains 700 images.
    \item \textbf{In-house TR Benchmark}. To better evaluate the table recognition accuracy of different methods, we considering various table attributes such as the number of table rows and columns, the number of merged cells, the length of the table, the length of the cell content, the type of cell content, the line style of the table, and construct a very diverse evaluation set, which contains approximately 500 tables.
\end{itemize}

\mineru{} achieves SOTA performance on most benchmarks, and shows competitive results with the SOTA on the remaining ones. 
Specifically, for PubTabNet, Rapidtable~\cite{rapidtable} achieves the best performance in the TEDS-S metric, while dots.ocr~\cite{dots.ocr} excel in the TEDS metric. Meanwhile, despite using only 20\% of the PubTabNet training set, \mineru{} still demonstrate comparable results, coming second and third in TEDS and TEDS-S, respectively.
For FinTabNet, \mineru{} achieves the best result and outperform other methods by a significant margin, this could be mainly credited to the large-scale high-quality table data we extracted from financial reports for training. 
On CC-OCR benchmark, \mineru{} came third after Gemini-2.5 Pro and Qwen2.5-VL-72B. 
On OCRBench v2 benchmark, \mineru{}'s performance is competitive to that of Gemini-2.5 Pro, and it significantly outperform other methods. 
On the diverse In-house TR Benchmark, \mineru{} and Gemini-2.5 Pro both significantly outperform other methods, with \mineru{} achieving a slight advantage over Gemini-2.5 Pro.

\begin{table}[!t]
\setlength{\tabcolsep}{4pt}
\resizebox{\linewidth}{!}{
\begin{tabular}{lcccccccc}
\toprule[.9pt]
\multirow{2}{*}{Method} &
\multicolumn{5}{c}{Public Dataset} &
\multicolumn{3}{c}{In-house Dataset} \\
\cmidrule(rl){2-6} \cmidrule(rl){7-9}
& CPE & HWE & SCE & SPE & LaTeX-80M$^M$ & Chinese & Fuzzy Math & Complex \\
\midrule
UniMERNet$^*$~\cite{wang2024unimernet} & \textbf{98.2} & \textbf{96.5} & 95.4 & \textbf{99.2} & 83.9 & 84.0 & 84.3 & 67.9 \\
PP-Formula\_plus-L~\cite{liu2025pp} & \underline{98.2} & \underline{94.7} & \underline{95.7} & \underline{99.2} & 85.9 & 84.0 & 86.5 & 76.5 \\
Gemini-2.5-flash~\cite{comanici2025gemini}  & 89.2 & 90.0 & 85.1 & 97.5 & 78.7 & 88.1 & 89.4 & 80.1 \\
Qwen2.5-VL-72B~\cite{bai2025qwen2}   & 88.9 & 91.8 & 95.5 & 96.2 & 83.4 & \textbf{90.8} & 86.7 & 81.4 \\
GPT-4o~\cite{achiam2023gpt} & 82.7 & 85.9 & 87.8 & 96.7 & 73.4 & 88.3 & 85.0 & 78.6 \\
InternVL3.5-241B~\cite{wang2025internvl3} & 91.7 & 93.2 & 95.1 & 97.8 & \underline{86.9} & 82.7 & \underline{90.3} & \underline{82.0} \\
dots.ocr~\cite{dots.ocr} & 86.8 & 90.5 & 94.7 & 97.5 & 81.8 & 74.4 & 86.2 & 77.4 \\
\rowcolor{gray!20}
MinerU2.5 & 96.6 & 94.4 & \textbf{96.4} & 98.4 & \textbf{90.6} & \underline{90.7} & \textbf{92.6} & \textbf{82.2} \\
\bottomrule[.9pt]
\end{tabular}
}
\vspace{3mm}
\caption{Formula Recognition Performance~(CDM metric used for evaluation). \mineru{} achieves 4 SOTA results and one second-best result across 7 benchmarks. Latex-80M$^M$ denotes the matrix benchmark of Latex-80M dataset. $^*$ indicates that the UniMERNet results are based on an improved version compared to the publicly available open-source implementation.}
\label{tab:formula_recognition}
\end{table}

\subsubsection{Formula Recognition}

For formula recognition, comparison models include various approaches, covering specialized formula recognition models, document parsing models, and general vision-language models. The evaluation datasets consist of the following:

\begin{itemize}
    \item \textbf{UniMER-Test}~\cite{wang2024unimernet}  is a comprehensive evaluation dataset for general formula recognition. Targeted at real-world formula recognition across various scenarios, UniMER-Test includes four subsets: CPE~(complex printed equations), HWE~(handwritten equations), SPE~(screen printed equations), and SCE~(simple printed equations).
    \item \textbf{LaTeX-80M\textsuperscript{\textit{M}}} is a matrix subset of LaTeX-80M\footnote{\url{https://github.com/OleehyO/TexTeller}}, featuring intricate mathematical structures encompassing matrices, conditional expressions, and nested combinations.
    \item \textbf{In-house dataset} consists of the following subsets: (1) Chinese, targeted at evaluation on real-world document equations which contain Chinese characters. (2) Fuzzy math, which focuses on authentic mathematics textbooks and exam documents characterized by compromised visual quality due to factors like blur, degeneration, watermarks, and so on. (3) Complex, an extremely difficult dataset aimed at assessing the ability of converting the most complex mathematical formulas to LaTeX codes.
    
\end{itemize}

Results are shown in \Cref{tab:formula_recognition} and the CDM~\cite{wang2025image} metric is used for evaluation. Across all seven evaluation datasets, \mineru{} achieves the best results in four datasets and one second-best result, demonstrating SOTA formula recognition capabilities. Specifically, on public datasets, \mineru{} achieves best CDM results of 96.4 on SCE and 90.6 on LaTeX-80M$^M$, showcasing leading performance in scenarios involving blurred screenshots and complex matrices. Besides, on CPE, HWE, and SPE, while being slightly outperformed by specialized formula recognition models, \mineru{} still deliver comparable performance. On in-house evaluation datasets, \mineru{}'s performance in Chinese text recognition is on par with Qwen2.5-VL-72B, leading to a second-place result of 90.6. Meanwhile, \mineru{} achieves the best results on both the real-world mathematic documents~(Fuzzy Math) and extremely hard formula recognition~(Complex).

\section{Conclusion}
\label{section:conclusion}

In this paper, we present \mineru{}, a 1.2B-parameter vision-language model that achieves a new state-of-the-art in efficient document parsing through its innovative decoupled, coarse-to-fine strategy. By separating global layout analysis from local recognition, it delivers unprecedented accuracy in a lightweight model, effectively resolving the trade-off between performance and cost. Beyond its standalone capabilities, the primary significance of \mineru{} lies in its role as a foundational tool for the LLM era. Its ability to rapidly convert vast, unstructured document collections into clean, structured data is invaluable for curating high-quality pre-training corpora. Furthermore, by preserving the semantic integrity of tables, formulas, and layouts, it is poised to significantly enhance the quality and reliability of Retrieval-Augmented Generation (RAG) systems, unlocking the vast knowledge contained within complex documents for next-generation AI applications.

\clearpage

\clearpage
\newpage
\bibliographystyle{plainnat}
\setcitestyle{numbers}
\bibliography{paper}

\begin{thebibliography}{63}
\providecommand{\natexlab}[1]{#1}
\providecommand{\url}[1]{\texttt{#1}}
\expandafter\ifx\csname urlstyle\endcsname\relax
  \providecommand{\doi}[1]{doi: #1}\else
  \providecommand{\doi}{doi: \begingroup \urlstyle{rm}\Url}\fi

\bibitem[Achiam et~al.(2023)Achiam, Adler, Agarwal, Ahmad, Akkaya, Aleman, Almeida, Altenschmidt, Altman, Anadkat, et~al.]{achiam2023gpt}
Josh Achiam, Steven Adler, Sandhini Agarwal, Lama Ahmad, Ilge Akkaya, Florencia~Leoni Aleman, Diogo Almeida, Janko Altenschmidt, Sam Altman, Shyamal Anadkat, et~al.
\newblock Gpt-4 technical report.
\newblock \emph{arXiv preprint arXiv:2303.08774}, 2023.

\bibitem[Bai et~al.(2022)Bai, Liu, Meng, Li, Liu, Xie, Zheng, Wang, Hou, Wei, et~al.]{bai2022wukong}
Haoli Bai, Zhiguang Liu, Xiaojun Meng, Wentao Li, Shuang Liu, Nian Xie, Rongfu Zheng, Liangwei Wang, Lu~Hou, Jiansheng Wei, et~al.
\newblock Wukong-reader: Multi-modal pre-training for fine-grained visual document understanding.
\newblock \emph{arXiv preprint arXiv:2212.09621}, 2022.

\bibitem[Bai et~al.(2025)Bai, Chen, Liu, Wang, Ge, Song, Dang, Wang, Wang, Tang, et~al.]{bai2025qwen2}
Shuai Bai, Keqin Chen, Xuejing Liu, Jialin Wang, Wenbin Ge, Sibo Song, Kai Dang, Peng Wang, Shijie Wang, Jun Tang, et~al.
\newblock Qwen2. 5-vl technical report.
\newblock \emph{arXiv preprint arXiv:2502.13923}, 2025.

\bibitem[Blecher et~al.(2023)Blecher, Cucurull, Scialom, and Stojnic]{blecher2023nougat}
Lukas Blecher, Guillem Cucurull, Thomas Scialom, and Robert Stojnic.
\newblock Nougat: Neural optical understanding for academic documents.
\newblock \emph{arXiv preprint arXiv:2308.13418}, 2023.

\bibitem[chatdoc com(2025)]{OCRFlux2025}
chatdoc com.
\newblock Ocrflux.
\newblock \url{https://github.com/chatdoc-com/OCRFlux}, 2025.
\newblock Accessed:2025-09-25.

\bibitem[Chen et~al.(2025)Chen, Guo, Li, Zhang, Lin, Kuang, Zhang, Ming, Zhang, Wang, et~al.]{chen2025ocean}
Song Chen, Xinyu Guo, Yadong Li, Tao Zhang, Mingan Lin, Dongdong Kuang, Youwei Zhang, Lingfeng Ming, Fengyu Zhang, Yuran Wang, et~al.
\newblock Ocean-ocr: Towards general ocr application via a vision-language model.
\newblock \emph{arXiv preprint arXiv:2501.15558}, 2025.

\bibitem[Comanici et~al.(2025)Comanici, Bieber, Schaekermann, Pasupat, Sachdeva, Dhillon, Blistein, Ram, Zhang, Rosen, et~al.]{comanici2025gemini}
Gheorghe Comanici, Eric Bieber, Mike Schaekermann, Ice Pasupat, Noveen Sachdeva, Inderjit Dhillon, Marcel Blistein, Ori Ram, Dan Zhang, Evan Rosen, et~al.
\newblock Gemini 2.5: Pushing the frontier with advanced reasoning, multimodality, long context, and next generation agentic capabilities.
\newblock \emph{arXiv preprint arXiv:2507.06261}, 2025.

\bibitem[Cui et~al.(2025)Cui, Sun, Lin, Gao, Zhang, Liu, Wang, Zhang, Zhou, Liu, et~al.]{cui2025paddleocr}
Cheng Cui, Ting Sun, Manhui Lin, Tingquan Gao, Yubo Zhang, Jiaxuan Liu, Xueqing Wang, Zelun Zhang, Changda Zhou, Hongen Liu, et~al.
\newblock Paddleocr 3.0 technical report.
\newblock \emph{arXiv preprint arXiv:2507.05595}, 2025.

\bibitem[Da et~al.(2023)Da, Luo, Zheng, and Yao]{da2023vision}
Cheng Da, Chuwei Luo, Qi~Zheng, and Cong Yao.
\newblock Vision grid transformer for document layout analysis.
\newblock In \emph{Proceedings of the IEEE/CVF international conference on computer vision}, pages 19462--19472, 2023.

\bibitem[Dehghani et~al.(2023)Dehghani, Mustafa, Djolonga, Heek, Minderer, Caron, Steiner, Puigcerver, Geirhos, Alabdulmohsin, et~al.]{dehghani2023patch}
Mostafa Dehghani, Basil Mustafa, Josip Djolonga, Jonathan Heek, Matthias Minderer, Mathilde Caron, Andreas Steiner, Joan Puigcerver, Robert Geirhos, Ibrahim~M Alabdulmohsin, et~al.
\newblock Patch n’pack: Navit, a vision transformer for any aspect ratio and resolution.
\newblock \emph{Advances in Neural Information Processing Systems}, 36:\penalty0 2252--2274, 2023.

\bibitem[Feng et~al.(2025)Feng, Wei, Fei, Shi, Han, Liao, Lu, Wu, Liu, Lin, et~al.]{feng2025dolphin}
Hao Feng, Shu Wei, Xiang Fei, Wei Shi, Yingdong Han, Lei Liao, Jinghui Lu, Binghong Wu, Qi~Liu, Chunhui Lin, et~al.
\newblock Dolphin: Document image parsing via heterogeneous anchor prompting.
\newblock \emph{arXiv preprint arXiv:2505.14059}, 2025.

\bibitem[Fu et~al.(2024)Fu, Kuang, Song, Huang, Yang, Li, Zhu, Luo, Wang, Lu, et~al.]{fu2024ocrbench}
Ling Fu, Zhebin Kuang, Jiajun Song, Mingxin Huang, Biao Yang, Yuzhe Li, Linghao Zhu, Qidi Luo, Xinyu Wang, Hao Lu, et~al.
\newblock Ocrbench v2: An improved benchmark for evaluating large multimodal models on visual text localization and reasoning.
\newblock \emph{arXiv preprint arXiv:2501.00321}, 2024.

\bibitem[Guo et~al.(2025)Guo, Wu, Zhu, Leng, Shi, Chen, Fan, Wang, Jiang, Wang, et~al.]{guo2025seed1}
Dong Guo, Faming Wu, Feida Zhu, Fuxing Leng, Guang Shi, Haobin Chen, Haoqi Fan, Jian Wang, Jianyu Jiang, Jiawei Wang, et~al.
\newblock Seed1. 5-vl technical report.
\newblock \emph{arXiv preprint arXiv:2505.07062}, 2025.

\bibitem[Huang et~al.(2022)Huang, Lv, Cui, Lu, and Wei]{huang2022layoutlmv3}
Yupan Huang, Tengchao Lv, Lei Cui, Yutong Lu, and Furu Wei.
\newblock Layoutlmv3: Pre-training for document ai with unified text and image masking.
\newblock In \emph{Proceedings of the 30th ACM international conference on multimedia}, pages 4083--4091, 2022.

\bibitem[Kim et~al.(2022)Kim, Hong, Yim, Nam, Park, Yim, Hwang, Yun, Han, and Park]{kim2022ocr}
Geewook Kim, Teakgyu Hong, Moonbin Yim, JeongYeon Nam, Jinyoung Park, Jinyeong Yim, Wonseok Hwang, Sangdoo Yun, Dongyoon Han, and Seunghyun Park.
\newblock Ocr-free document understanding transformer.
\newblock In \emph{European Conference on Computer Vision}, pages 498--517. Springer, 2022.

\bibitem[Kwon et~al.(2023)Kwon, Li, Zhuang, Sheng, Zheng, Yu, Gonzalez, Zhang, and Stoica]{kwon2023efficient}
Woosuk Kwon, Zhuohan Li, Siyuan Zhuang, Ying Sheng, Lianmin Zheng, Cody~Hao Yu, Joseph~E. Gonzalez, Hao Zhang, and Ion Stoica.
\newblock Efficient memory management for large language model serving with pagedattention.
\newblock In \emph{Proceedings of the ACM SIGOPS 29th Symposium on Operating Systems Principles}, 2023.

\bibitem[Li et~al.(2025)Li, Liu, Liu, Ma, Zhang, Zhang, Guo, Zhang, Wang, and Bai]{li2025monkeyocr}
Zhang Li, Yuliang Liu, Qiang Liu, Zhiyin Ma, Ziyang Zhang, Shuo Zhang, Zidun Guo, Jiarui Zhang, Xinyu Wang, and Xiang Bai.
\newblock Monkeyocr: Document parsing with a structure-recognition-relation triplet paradigm.
\newblock \emph{arXiv preprint arXiv:2506.05218}, 2025.

\bibitem[Liao et~al.(2023)Liao, RoyChowdhury, Li, Bansal, Zhang, Tu, Satzoda, Manmatha, and Mahadevan]{liao2023doctr}
Haofu Liao, Aruni RoyChowdhury, Weijian Li, Ankan Bansal, Yuting Zhang, Zhuowen Tu, Ravi~Kumar Satzoda, R~Manmatha, and Vijay Mahadevan.
\newblock Doctr: Document transformer for structured information extraction in documents.
\newblock In \emph{Proceedings of the IEEE/CVF International Conference on Computer Vision}, pages 19584--19594, 2023.

\bibitem[Lin(2024)]{lin2024revolutionizing}
Demiao Lin.
\newblock Revolutionizing retrieval-augmented generation with enhanced pdf structure recognition.
\newblock \emph{arXiv preprint arXiv:2401.12599}, 2024.

\bibitem[Liu et~al.(2024{\natexlab{a}})Liu, Yin, Cao, Jiang, Li, Liu, Jiang, Sun, and Xu]{liu2024hrvda}
Chaohu Liu, Kun Yin, Haoyu Cao, Xinghua Jiang, Xin Li, Yinsong Liu, Deqiang Jiang, Xing Sun, and Linli Xu.
\newblock Hrvda: High-resolution visual document assistant.
\newblock In \emph{Proceedings of the IEEE/CVF conference on computer vision and pattern recognition}, pages 15534--15545, 2024{\natexlab{a}}.

\bibitem[Liu et~al.(2025{\natexlab{a}})Liu, Cui, Du, Liu, and Pan]{liu2025pp}
Hongen Liu, Cheng Cui, Yuning Du, Yi~Liu, and Gang Pan.
\newblock Pp-formulanet: Bridging accuracy and efficiency in advanced formula recognition.
\newblock \emph{arXiv preprint arXiv:2503.18382}, 2025{\natexlab{a}}.

\bibitem[Liu et~al.(2025{\natexlab{b}})Liu, Zhao, Tian, Wang, Ye, You, Yu, Wu, Zhou, Yu, et~al.]{liu2025points}
Yuan Liu, Zhongyin Zhao, Le~Tian, Haicheng Wang, Xubing Ye, Yangxiu You, Zilin Yu, Chuhan Wu, Xiao Zhou, Yang Yu, et~al.
\newblock Points-reader: Distillation-free adaptation of vision-language models for document conversion.
\newblock \emph{arXiv preprint arXiv:2509.01215}, 2025{\natexlab{b}}.

\bibitem[Liu et~al.(2024{\natexlab{b}})Liu, Yang, Liu, Li, Ma, Zhang, and Bai]{liu2024textmonkey}
Yuliang Liu, Biao Yang, Qiang Liu, Zhang Li, Zhiyin Ma, Shuo Zhang, and Xiang Bai.
\newblock Textmonkey: An ocr-free large multimodal model for understanding document.
\newblock \emph{arXiv preprint arXiv:2403.04473}, 2024{\natexlab{b}}.

\bibitem[Livathinos et~al.(2025)Livathinos, Auer, Lysak, Nassar, Dolfi, Vagenas, Ramis, Omenetti, Dinkla, Kim, et~al.]{livathinos2025docling}
Nikolaos Livathinos, Christoph Auer, Maksym Lysak, Ahmed Nassar, Michele Dolfi, Panos Vagenas, Cesar~Berrospi Ramis, Matteo Omenetti, Kasper Dinkla, Yusik Kim, et~al.
\newblock Docling: An efficient open-source toolkit for ai-driven document conversion.
\newblock \emph{arXiv preprint arXiv:2501.17887}, 2025.

\bibitem[Lysak et~al.(2023)Lysak, Nassar, Livathinos, Auer, and Staar]{lysak2023optimized}
Maksym Lysak, Ahmed Nassar, Nikolaos Livathinos, Christoph Auer, and Peter Staar.
\newblock Optimized table tokenization for table structure recognition.
\newblock In \emph{International Conference on Document Analysis and Recognition}, pages 37--50. Springer, 2023.

\bibitem[Mandalm(2025)]{nanonets2025}
Souvik Mandalm.
\newblock Nanonets-ocr-s.
\newblock \url{https://nanonets.com/research/nanonets-ocr-s/}, 2025.
\newblock Accessed:2025-09-25.

\bibitem[Mathpix(2025)]{Mathpix2025}
Mathpix.
\newblock Mathpix.
\newblock \url{https://mathpix.com/}, 2025.
\newblock Accessed:2025-09-25.

\bibitem[Nassar et~al.(2025)Nassar, Marafioti, Omenetti, Lysak, Livathinos, Auer, Morin, de~Lima, Kim, Gurbuz, et~al.]{nassar2025smoldocling}
Ahmed Nassar, Andres Marafioti, Matteo Omenetti, Maksym Lysak, Nikolaos Livathinos, Christoph Auer, Lucas Morin, Rafael~Teixeira de~Lima, Yusik Kim, A~Said Gurbuz, et~al.
\newblock Smoldocling: An ultra-compact vision-language model for end-to-end multi-modal document conversion.
\newblock \emph{arXiv preprint arXiv:2503.11576}, 2025.

\bibitem[Niu et~al.(2025)Niu, Zheng, Miao, Dong, Ge, Liang, Lu, Zeng, Zheng, He, et~al.]{niu2025native}
Junbo Niu, Yuanhong Zheng, Ziyang Miao, Hejun Dong, Chunjiang Ge, Hao Liang, Ma~Lu, Bohan Zeng, Qiahao Zheng, Conghui He, et~al.
\newblock Native visual understanding: Resolving resolution dilemmas in vision-language models.
\newblock \emph{arXiv preprint arXiv:2506.12776}, 2025.

\bibitem[OpenDataLab(2025)]{pdfExtractKit2025}
OpenDataLab.
\newblock Pdf-extract-kit.
\newblock \url{https://github.com/opendatalab/PDF-Extract-Kit}, 2025.
\newblock Accessed:2025-09-25.

\bibitem[Ouyang et~al.(2025)Ouyang, Qu, Zhou, Zhu, Zhang, Lin, Wang, Zhao, Jiang, Zhao, et~al.]{ouyang2025omnidocbench}
Linke Ouyang, Yuan Qu, Hongbin Zhou, Jiawei Zhu, Rui Zhang, Qunshu Lin, Bin Wang, Zhiyuan Zhao, Man Jiang, Xiaomeng Zhao, et~al.
\newblock Omnidocbench: Benchmarking diverse pdf document parsing with comprehensive annotations.
\newblock In \emph{Proceedings of the Computer Vision and Pattern Recognition Conference}, pages 24838--24848, 2025.

\bibitem[Paruchuri(2025)]{vik2024marker}
Vik Paruchuri.
\newblock Marker.
\newblock \url{https://github.com/datalab-to/marker}, 2025.
\newblock Accessed:2025-09-25.

\bibitem[Paruchuri and Team(2025)]{paruchuri2025surya}
Vikas Paruchuri and Datalab Team.
\newblock Surya: A lightweight document ocr and analysis toolkit.
\newblock \url{https://github.com/VikParuchuri/surya}, 2025.
\newblock Accessed:2025-09-25.

\bibitem[Pfitzmann et~al.(2022)Pfitzmann, Auer, Dolfi, Nassar, and Staar]{pfitzmann2022doclaynet}
Birgit Pfitzmann, Christoph Auer, Michele Dolfi, Ahmed~S Nassar, and Peter Staar.
\newblock Doclaynet: A large human-annotated dataset for document-layout segmentation.
\newblock In \emph{Proceedings of the 28th ACM SIGKDD conference on knowledge discovery and data mining}, pages 3743--3751, 2022.

\bibitem[Poznanski et~al.(2025)Poznanski, Rangapur, Borchardt, Dunkelberger, Huff, Lin, Wilhelm, Lo, and Soldaini]{poznanski2025olmocr}
Jake Poznanski, Aman Rangapur, Jon Borchardt, Jason Dunkelberger, Regan Huff, Daniel Lin, Christopher Wilhelm, Kyle Lo, and Luca Soldaini.
\newblock olmocr: Unlocking trillions of tokens in pdfs with vision language models.
\newblock \emph{arXiv preprint arXiv:2502.18443}, 2025.

\bibitem[RapidAI(2024)]{rapidtable}
RapidAI.
\newblock Rapid table.
\newblock \url{https://github.com/RapidAI/RapidTable}, 2024.
\newblock Accessed: 2025-9-25.

\bibitem[rednote(2025)]{dots.ocr}
rednote.
\newblock dots.ocr: Multilingual document layout parsing in a single vision-language model.
\newblock \url{https://github.com/rednote-hilab/dots.ocr}, 2025.
\newblock Accessed:2025-09-25.

\bibitem[Shi et~al.(2016)Shi, Caballero, Husz{\'a}r, Totz, Aitken, Bishop, Rueckert, and Wang]{shi2016real}
Wenzhe Shi, Jose Caballero, Ferenc Husz{\'a}r, Johannes Totz, Andrew~P Aitken, Rob Bishop, Daniel Rueckert, and Zehan Wang.
\newblock Real-time single image and video super-resolution using an efficient sub-pixel convolutional neural network.
\newblock In \emph{Proceedings of the IEEE conference on computer vision and pattern recognition}, pages 1874--1883, 2016.

\bibitem[Su et~al.(2024)Su, Ahmed, Lu, Pan, Bo, and Liu]{su2024roformer}
Jianlin Su, Murtadha Ahmed, Yu~Lu, Shengfeng Pan, Wen Bo, and Yunfeng Liu.
\newblock Roformer: Enhanced transformer with rotary position embedding.
\newblock \emph{Neurocomputing}, 568:\penalty0 127063, 2024.

\bibitem[Tang et~al.(2023)Tang, Yang, Wang, Fang, Liu, Zhu, Zeng, Zhang, and Bansal]{tang2023unifying}
Zineng Tang, Ziyi Yang, Guoxin Wang, Yuwei Fang, Yang Liu, Chenguang Zhu, Michael Zeng, Cha Zhang, and Mohit Bansal.
\newblock Unifying vision, text, and layout for universal document processing.
\newblock In \emph{Proceedings of the IEEE/CVF conference on computer vision and pattern recognition}, pages 19254--19264, 2023.

\bibitem[Team(2025)]{mistral2025}
Mistral~AI Team.
\newblock Mistral-ocr.
\newblock \url{https://mistral.ai/news/mistral-ocr?utm_source=ai-bot.cn}, 2025.
\newblock Accessed:2025-09-25.

\bibitem[Team(2024)]{team2024qwen2}
Qwen Team.
\newblock Qwen2 technical report.
\newblock \emph{arXiv preprint arXiv:2407.10671}, 2024.

\bibitem[Wan et~al.(2024)Wan, Song, Yu, Liu, Cheng, Huang, Bai, Yao, and Yang]{wan2024omniparser}
Jianqiang Wan, Sibo Song, Wenwen Yu, Yuliang Liu, Wenqing Cheng, Fei Huang, Xiang Bai, Cong Yao, and Zhibo Yang.
\newblock Omniparser: A unified framework for text spotting key information extraction and table recognition.
\newblock In \emph{Proceedings of the IEEE/CVF conference on computer vision and pattern recognition}, pages 15641--15653, 2024.

\bibitem[Wang et~al.(2024{\natexlab{a}})Wang, Chen, Liu, Chen, Lin, Han, et~al.]{wang2024yolov10}
Ao~Wang, Hui Chen, Lihao Liu, Kai Chen, Zijia Lin, Jungong Han, et~al.
\newblock Yolov10: Real-time end-to-end object detection.
\newblock \emph{Advances in Neural Information Processing Systems}, 37:\penalty0 107984--108011, 2024{\natexlab{a}}.

\bibitem[Wang et~al.(2024{\natexlab{b}})Wang, Gu, Liang, Xu, Zhang, Shi, and He]{wang2024unimernet}
Bin Wang, Zhuangcheng Gu, Guang Liang, Chao Xu, Bo~Zhang, Botian Shi, and Conghui He.
\newblock Unimernet: A universal network for real-world mathematical expression recognition.
\newblock \emph{arXiv preprint arXiv:2404.15254}, 2024{\natexlab{b}}.

\bibitem[Wang et~al.(2024{\natexlab{c}})Wang, Xu, Zhao, Ouyang, Wu, Zhao, Xu, Liu, Qu, Shang, et~al.]{wang2024mineru}
Bin Wang, Chao Xu, Xiaomeng Zhao, Linke Ouyang, Fan Wu, Zhiyuan Zhao, Rui Xu, Kaiwen Liu, Yuan Qu, Fukai Shang, et~al.
\newblock Mineru: An open-source solution for precise document content extraction.
\newblock \emph{arXiv preprint arXiv:2409.18839}, 2024{\natexlab{c}}.

\bibitem[Wang et~al.(2025{\natexlab{a}})Wang, Wu, Ouyang, Gu, Zhang, Xia, Shi, Zhang, and He]{wang2025image}
Bin Wang, Fan Wu, Linke Ouyang, Zhuangcheng Gu, Rui Zhang, Renqiu Xia, Botian Shi, Bo~Zhang, and Conghui He.
\newblock Image over text: Transforming formula recognition evaluation with character detection matching.
\newblock In \emph{Proceedings of the Computer Vision and Pattern Recognition Conference}, pages 19681--19690, 2025{\natexlab{a}}.

\bibitem[Wang et~al.(2024{\natexlab{d}})Wang, Bai, Tan, Wang, Fan, Bai, Chen, Liu, Wang, Ge, et~al.]{wang2024qwen2}
Peng Wang, Shuai Bai, Sinan Tan, Shijie Wang, Zhihao Fan, Jinze Bai, Keqin Chen, Xuejing Liu, Jialin Wang, Wenbin Ge, et~al.
\newblock Qwen2-vl: Enhancing vision-language model's perception of the world at any resolution.
\newblock \emph{arXiv preprint arXiv:2409.12191}, 2024{\natexlab{d}}.

\bibitem[Wang et~al.(2025{\natexlab{b}})Wang, Gao, Gu, Pu, Cui, Wei, Liu, Jing, Ye, Shao, et~al.]{wang2025internvl3}
Weiyun Wang, Zhangwei Gao, Lixin Gu, Hengjun Pu, Long Cui, Xingguang Wei, Zhaoyang Liu, Linglin Jing, Shenglong Ye, Jie Shao, et~al.
\newblock Internvl3. 5: Advancing open-source multimodal models in versatility, reasoning, and efficiency.
\newblock \emph{arXiv preprint arXiv:2508.18265}, 2025{\natexlab{b}}.

\bibitem[Wang et~al.(2021)Wang, Xu, Cui, Shang, and Wei]{wang2021layoutreader}
Zilong Wang, Yiheng Xu, Lei Cui, Jingbo Shang, and Furu Wei.
\newblock Layoutreader: Pre-training of text and layout for reading order detection.
\newblock \emph{arXiv preprint arXiv:2108.11591}, 2021.

\bibitem[Wang et~al.(2023)Wang, Zhou, Wei, Lee, and Tata]{wang2023vrdu}
Zilong Wang, Yichao Zhou, Wei Wei, Chen-Yu Lee, and Sandeep Tata.
\newblock Vrdu: A benchmark for visually-rich document understanding.
\newblock In \emph{Proceedings of the 29th ACM SIGKDD Conference on Knowledge Discovery and Data Mining}, pages 5184--5193, 2023.

\bibitem[Wei et~al.(2024)Wei, Liu, Chen, Wang, Kong, Xu, Ge, Zhao, Sun, Peng, et~al.]{wei2024general}
Haoran Wei, Chenglong Liu, Jinyue Chen, Jia Wang, Lingyu Kong, Yanming Xu, Zheng Ge, Liang Zhao, Jianjian Sun, Yuang Peng, et~al.
\newblock General ocr theory: Towards ocr-2.0 via a unified end-to-end model.
\newblock \emph{arXiv preprint arXiv:2409.01704}, 2024.

\bibitem[Yang et~al.(2024)Yang, Tang, Li, Wang, Wan, Zhong, Liu, Yang, Wang, Bai, et~al.]{yang2024cc}
Zhibo Yang, Jun Tang, Zhaohai Li, Pengfei Wang, Jianqiang Wan, Humen Zhong, Xuejing Liu, Mingkun Yang, Peng Wang, Shuai Bai, et~al.
\newblock Cc-ocr: A comprehensive and challenging ocr benchmark for evaluating large multimodal models in literacy.
\newblock \emph{arXiv preprint arXiv:2412.02210}, 2024.

\bibitem[Yao et~al.(2024)Yao, Yu, Zhang, Wang, Cui, Zhu, Cai, Li, Zhao, He, et~al.]{yao2024minicpm}
Yuan Yao, Tianyu Yu, Ao~Zhang, Chongyi Wang, Junbo Cui, Hongji Zhu, Tianchi Cai, Haoyu Li, Weilin Zhao, Zhihui He, et~al.
\newblock Minicpm-v: A gpt-4v level mllm on your phone.
\newblock \emph{arXiv preprint arXiv:2408.01800}, 2024.

\bibitem[Yu et~al.(2025)Yu, Wang, Wang, Huang, Ma, He, Cai, Chen, Huang, Zhao, Xu, Cui, Xu, Ruan, Zhang, Liu, Tang, Liu, Guo, Hu, He, Zhou, Cai, Qi, Guo, Chen, Zeng, Li, Cui, Ding, Han, Yao, Liu, and Sun]{yu2025minicpmv45cookingefficient}
Tianyu Yu, Zefan Wang, Chongyi Wang, Fuwei Huang, Wenshuo Ma, Zhihui He, Tianchi Cai, Weize Chen, Yuxiang Huang, Yuanqian Zhao, Bokai Xu, Junbo Cui, Yingjing Xu, Liqing Ruan, Luoyuan Zhang, Hanyu Liu, Jingkun Tang, Hongyuan Liu, Qining Guo, Wenhao Hu, Bingxiang He, Jie Zhou, Jie Cai, Ji~Qi, Zonghao Guo, Chi Chen, Guoyang Zeng, Yuxuan Li, Ganqu Cui, Ning Ding, Xu~Han, Yuan Yao, Zhiyuan Liu, and Maosong Sun.
\newblock Minicpm-v 4.5: Cooking efficient mllms via architecture, data, and training recipe.
\newblock \emph{arXiv preprint arXiv:2509.18154}, 2025.

\bibitem[Zhang et~al.(2024{\natexlab{a}})Zhang, Zhang, Wang, Ouyang, Wen, Li, Chow, He, and Zhang]{zhang2024ocr}
Junyuan Zhang, Qintong Zhang, Bin Wang, Linke Ouyang, Zichen Wen, Ying Li, Ka-Ho Chow, Conghui He, and Wentao Zhang.
\newblock Ocr hinders rag: Evaluating the cascading impact of ocr on retrieval-augmented generation.
\newblock \emph{arXiv preprint arXiv:2412.02592}, 2024{\natexlab{a}}.

\bibitem[Zhang et~al.(2024{\natexlab{b}})Zhang, Wang, Huang, Zhang, Wang, Liang, He, and Zhang]{zhang2024document}
Qintong Zhang, Bin Wang, Victor Shea-Jay Huang, Junyuan Zhang, Zhengren Wang, Hao Liang, Conghui He, and Wentao Zhang.
\newblock Document parsing unveiled: Techniques, challenges, and prospects for structured information extraction.
\newblock \emph{arXiv preprint arXiv:2410.21169}, 2024{\natexlab{b}}.

\bibitem[Zhao et~al.(2024{\natexlab{a}})Zhao, Zhang, Yu, Wang, Geng, Fu, Yang, Zhang, Jiang, and Cui]{zhao2024retrieval}
Penghao Zhao, Hailin Zhang, Qinhan Yu, Zhengren Wang, Yunteng Geng, Fangcheng Fu, Ling Yang, Wentao Zhang, Jie Jiang, and Bin Cui.
\newblock Retrieval-augmented generation for ai-generated content: A survey.
\newblock \emph{arXiv preprint arXiv:2402.19473}, 2024{\natexlab{a}}.

\bibitem[Zhao et~al.(2024{\natexlab{b}})Zhao, Kang, Wang, and He]{zhao2024doclayout}
Zhiyuan Zhao, Hengrui Kang, Bin Wang, and Conghui He.
\newblock Doclayout-yolo: Enhancing document layout analysis through diverse synthetic data and global-to-local adaptive perception.
\newblock \emph{arXiv preprint arXiv:2410.12628}, 2024{\natexlab{b}}.

\bibitem[Zheng et~al.(2024)Zheng, Yin, Xie, Sun, Huang, Yu, Cao, Kozyrakis, Stoica, Gonzalez, et~al.]{zheng2024sglang}
Lianmin Zheng, Liangsheng Yin, Zhiqiang Xie, Chuyue~Livia Sun, Jeff Huang, Cody~Hao Yu, Shiyi Cao, Christos Kozyrakis, Ion Stoica, Joseph~E Gonzalez, et~al.
\newblock Sglang: Efficient execution of structured language model programs.
\newblock \emph{Advances in neural information processing systems}, 37:\penalty0 62557--62583, 2024.

\bibitem[Zheng et~al.(2021)Zheng, Burdick, Popa, Zhong, and Wang]{zheng2021global}
Xinyi Zheng, Douglas Burdick, Lucian Popa, Xu~Zhong, and Nancy Xin~Ru Wang.
\newblock Global table extractor (gte): A framework for joint table identification and cell structure recognition using visual context.
\newblock In \emph{Proceedings of the IEEE/CVF winter conference on applications of computer vision}, pages 697--706, 2021.

\bibitem[Zhong et~al.(2020)Zhong, ShafieiBavani, and Jimeno~Yepes]{zhong2020image}
Xu~Zhong, Elaheh ShafieiBavani, and Antonio Jimeno~Yepes.
\newblock Image-based table recognition: data, model, and evaluation.
\newblock In \emph{European conference on computer vision}, pages 564--580. Springer, 2020.

\bibitem[Zhu et~al.(2025)Zhu, Wang, Chen, Liu, Ye, Gu, Tian, Duan, Su, Shao, et~al.]{zhu2025internvl3}
Jinguo Zhu, Weiyun Wang, Zhe Chen, Zhaoyang Liu, Shenglong Ye, Lixin Gu, Hao Tian, Yuchen Duan, Weijie Su, Jie Shao, et~al.
\newblock Internvl3: Exploring advanced training and test-time recipes for open-source multimodal models.
\newblock \emph{arXiv preprint arXiv:2504.10479}, 2025.

\end{thebibliography}

\clearpage

\beginappendix

\section{Qualitative examples}
\label{sec:app:qualitative}

This section presents qualitative examples illustrating the capabilities of the \mineru{} through document parsing outputs generated for various pages. This section is structured as follows: ~\Cref{subsec:overview} illustrates the \mineru{}'s performance on Document Parsing, Table Recognition and Formula Recognition among all types of documents. ~\Cref{subsec:compare_before} showcases specific attribute pages with improved performance. ~\Cref{subsec:compare_others} demonstrates \mineru{}'s performance on some complex pages compared to other models.

\begin{figure}[H]
\centering
\includegraphics[height=1.25\linewidth]{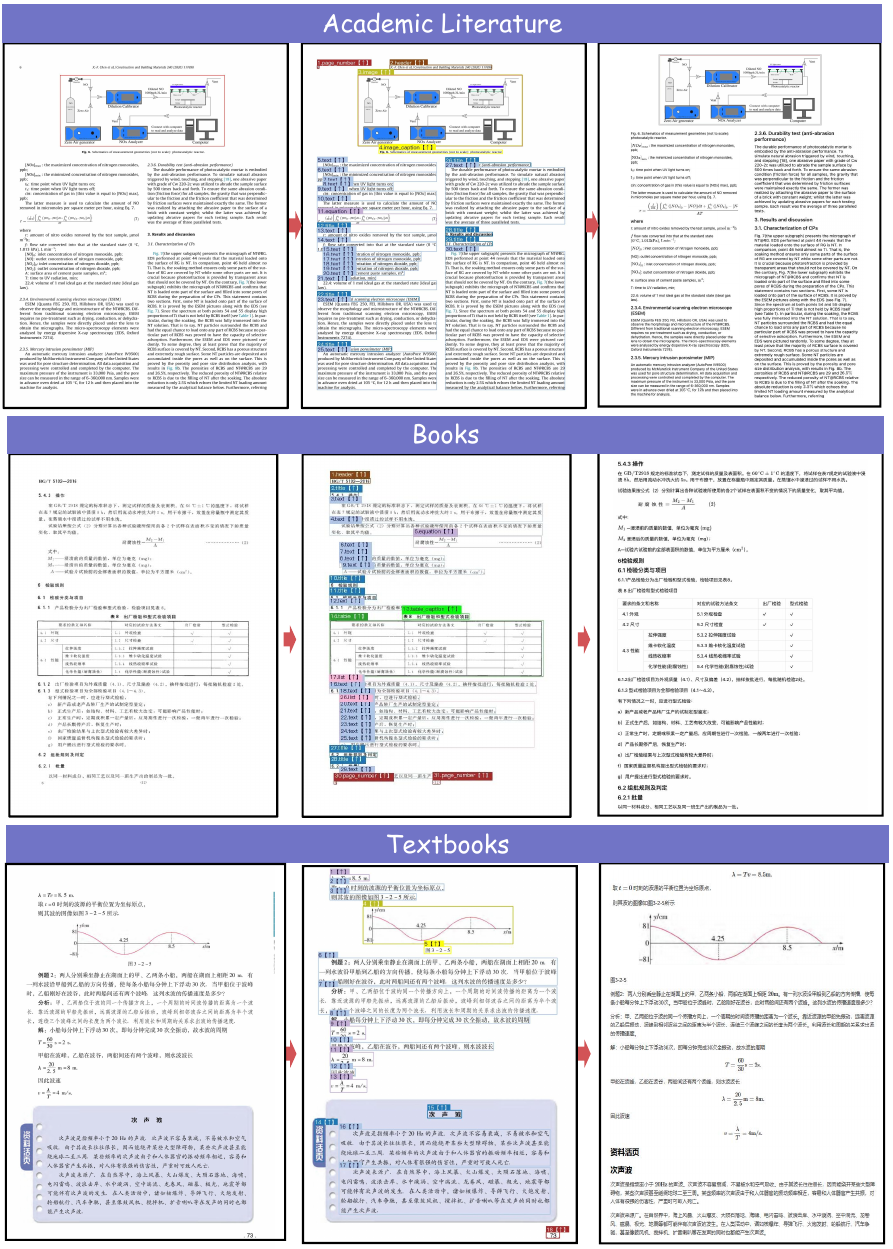}
\caption{The Layout and rendered markdown output for Academic literature, Books, Textbooks.}
\label{fig:pdf_type_1}
\end{figure}

\begin{figure}[H]
\centering
\includegraphics[height=1.25\linewidth]{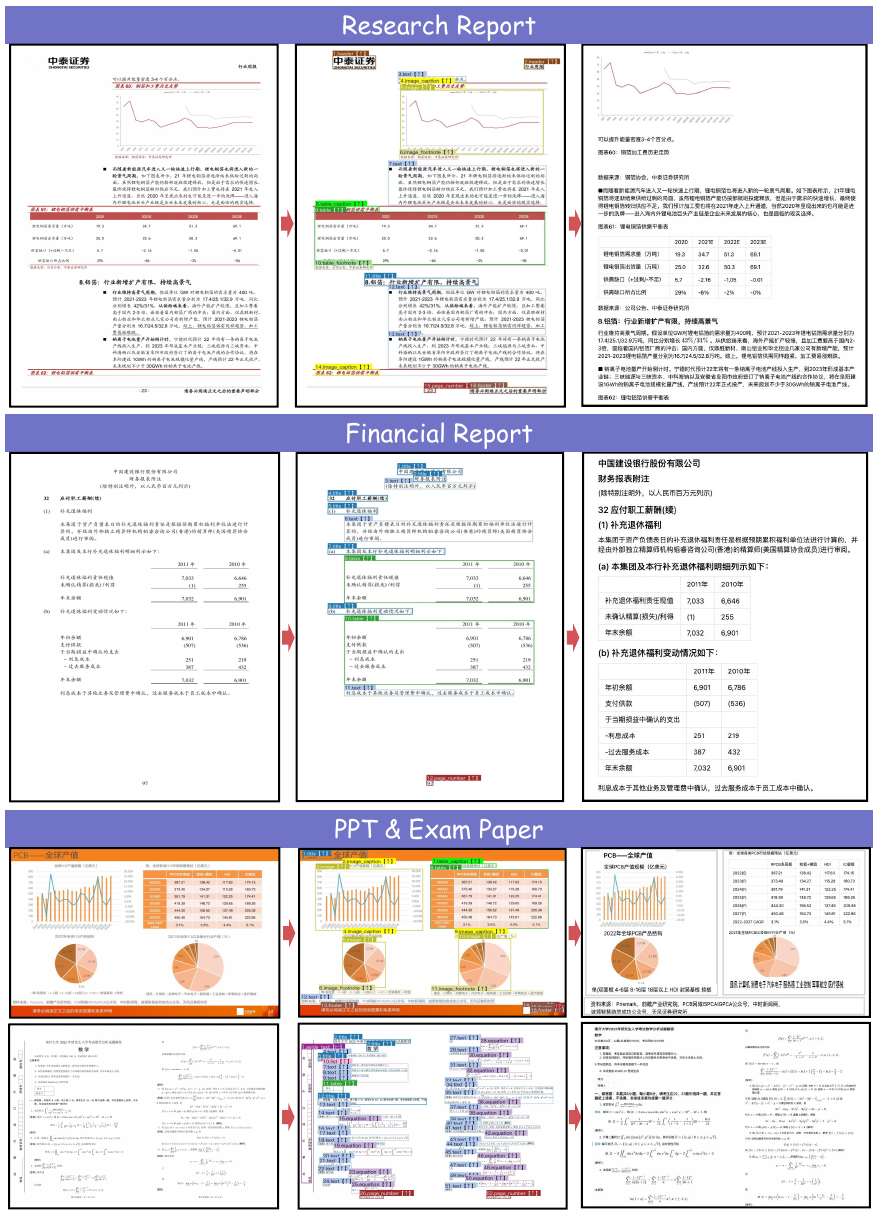}
\caption{The Layout and rendered markdown output for Research Report, Financial Report, Slides and Exam Paper.}
\label{fig:pdf_type_2}
\end{figure}

\begin{figure}[H]
\centering
\includegraphics[height=1.25\linewidth]{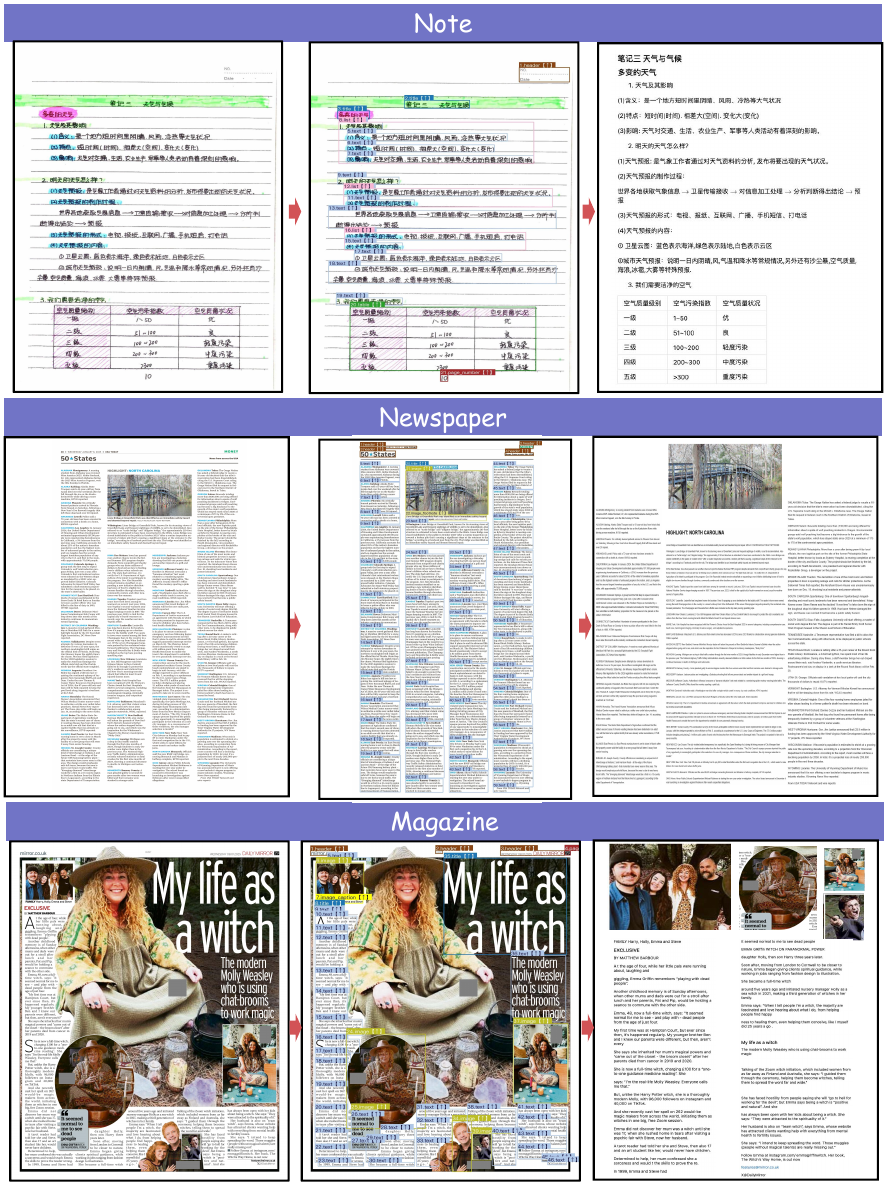}
\caption{The Layout and rendered markdown output for Note, Newspaper and Magazine.}
\label{fig:pdf_type_3}
\end{figure}

Examples demonstrating the Document Parsing performance among PDF types are provided in~\Cref{fig:pdf_type_1,fig:pdf_type_2,fig:pdf_type_3}, including Academic literature, Books, Textbooks, Research Report, Financial Report, Slides, Exam Paper, Note, Newspaper and Magazine. 

\begin{figure}[H]
\centering
\includegraphics[height=1.25\linewidth]{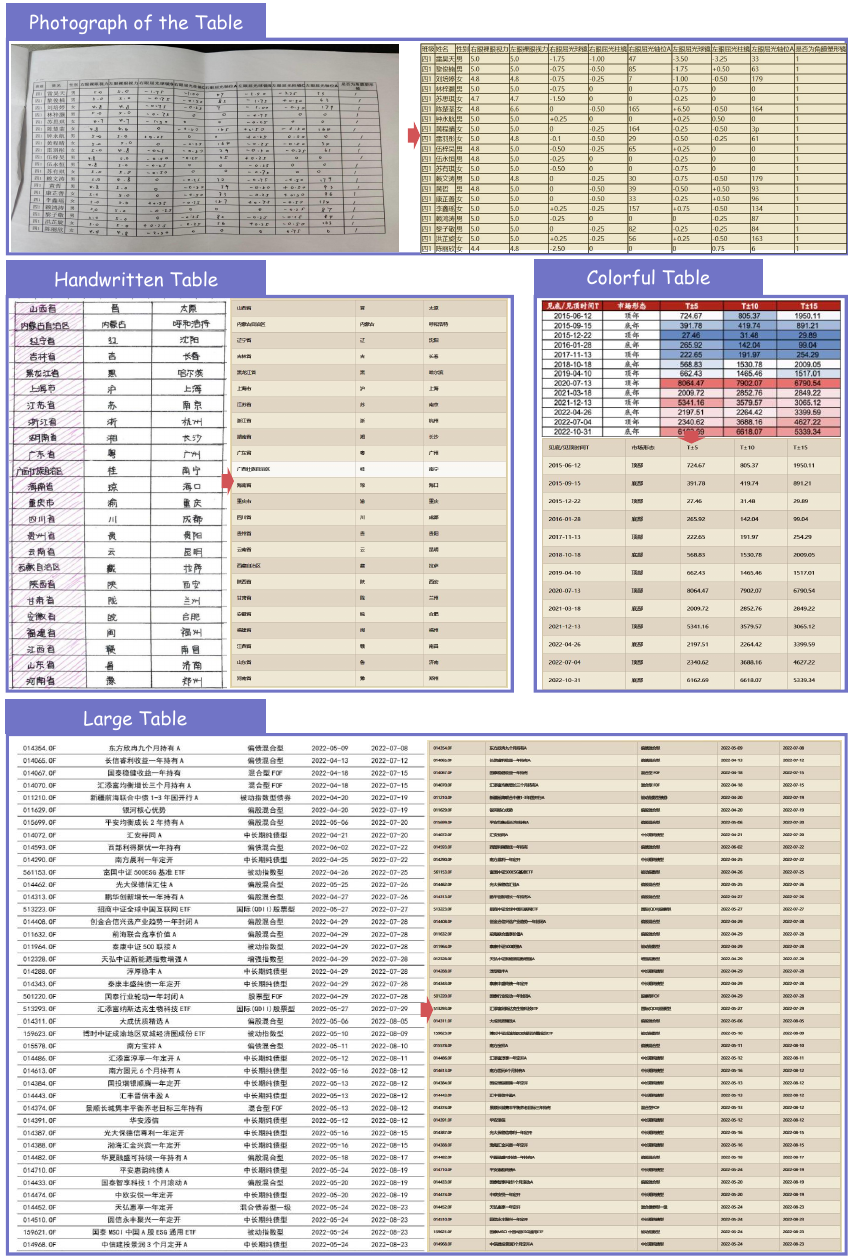}
\caption{The rendered outputs for various types of Tables.}
\label{fig:Table-Module-1}
\end{figure}

\begin{figure}[H]
\centering
\includegraphics[height=1.25\linewidth]{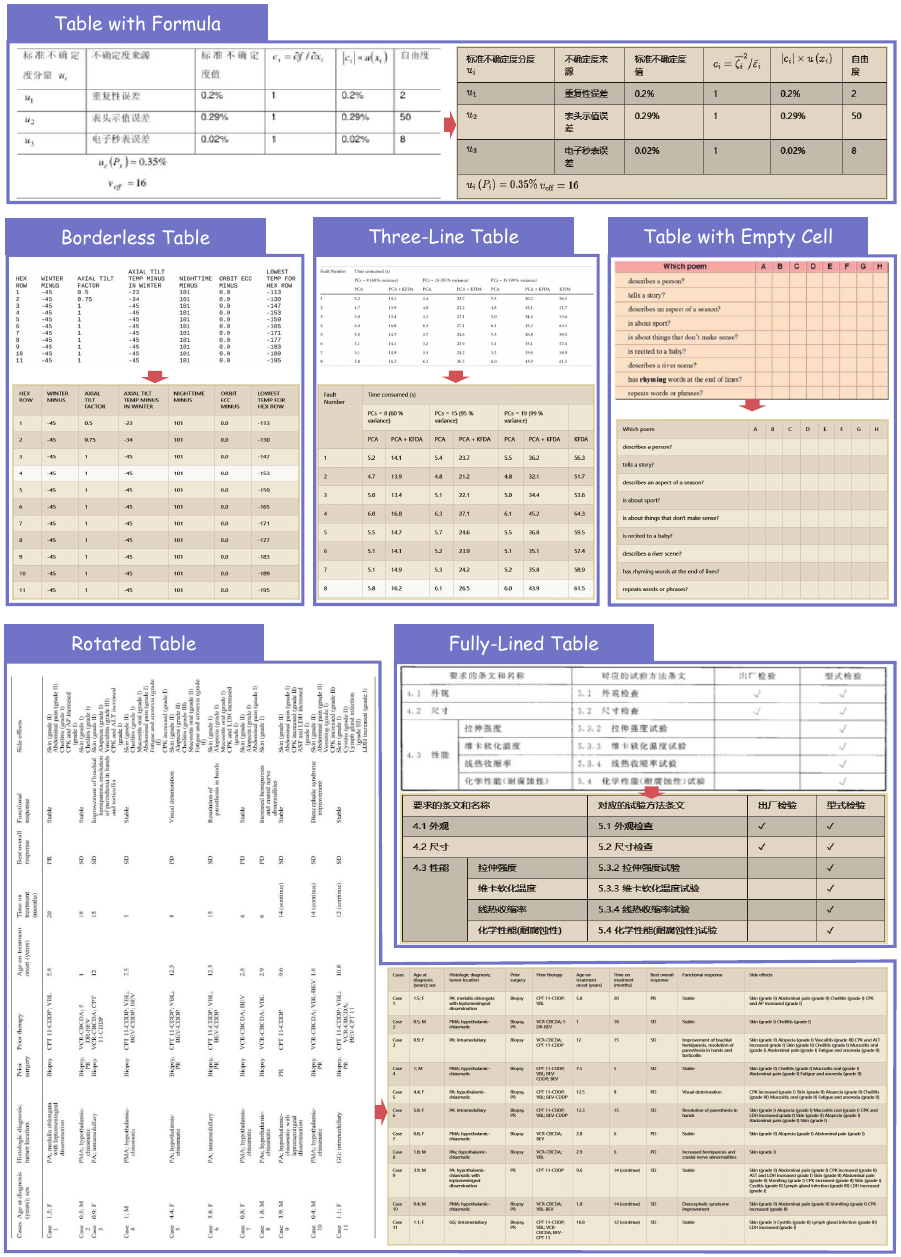}
\caption{The rendered outputs for various types of Tables.}
\label{fig:Table-Module-2}
\end{figure}

Table Recognition performance among various types of tables is demonstrated in ~\Cref{fig:Table-Module-1,fig:Table-Module-2}, including the photograph of the table, table with colorful background, table with formula, table with empty cells, handwritten table, large table, rotated table, no-line table, three-line table, and full-line table.

\begin{figure}[H]
\centering
\includegraphics[height=1.25\linewidth]{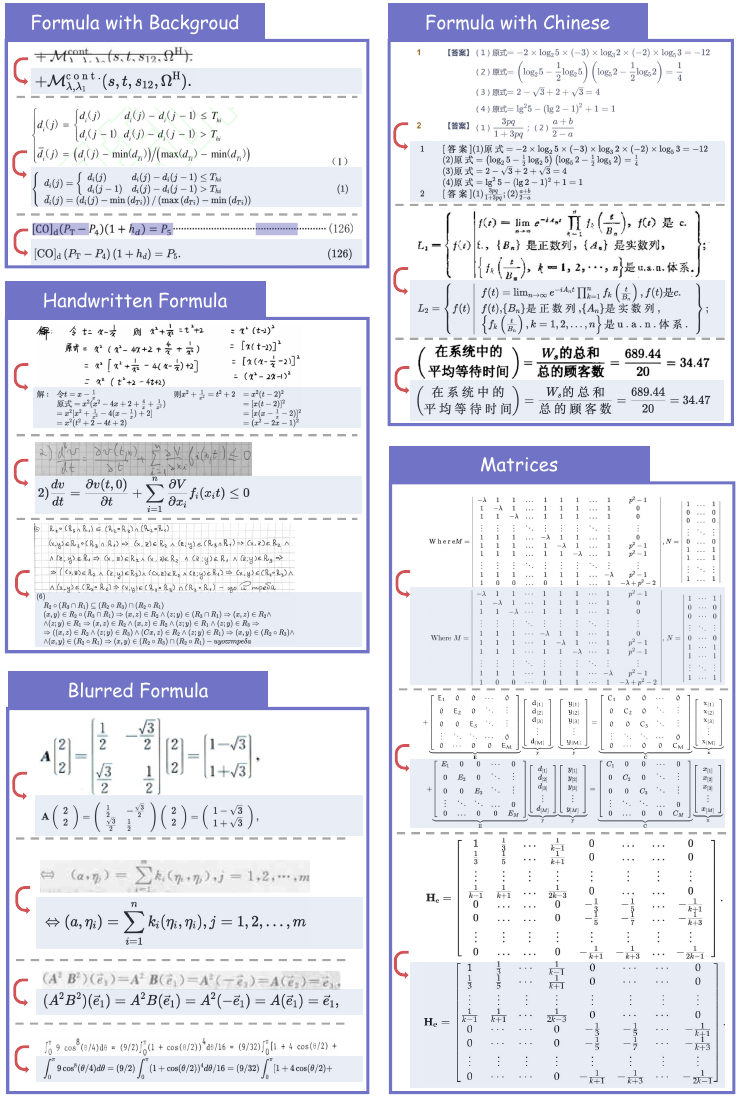}
\caption{The rendered outputs for various types of Formulas.}
\label{fig:Formula-Module-1}
\end{figure}

\begin{figure}[H]
\centering
\includegraphics[height=1.25\linewidth]{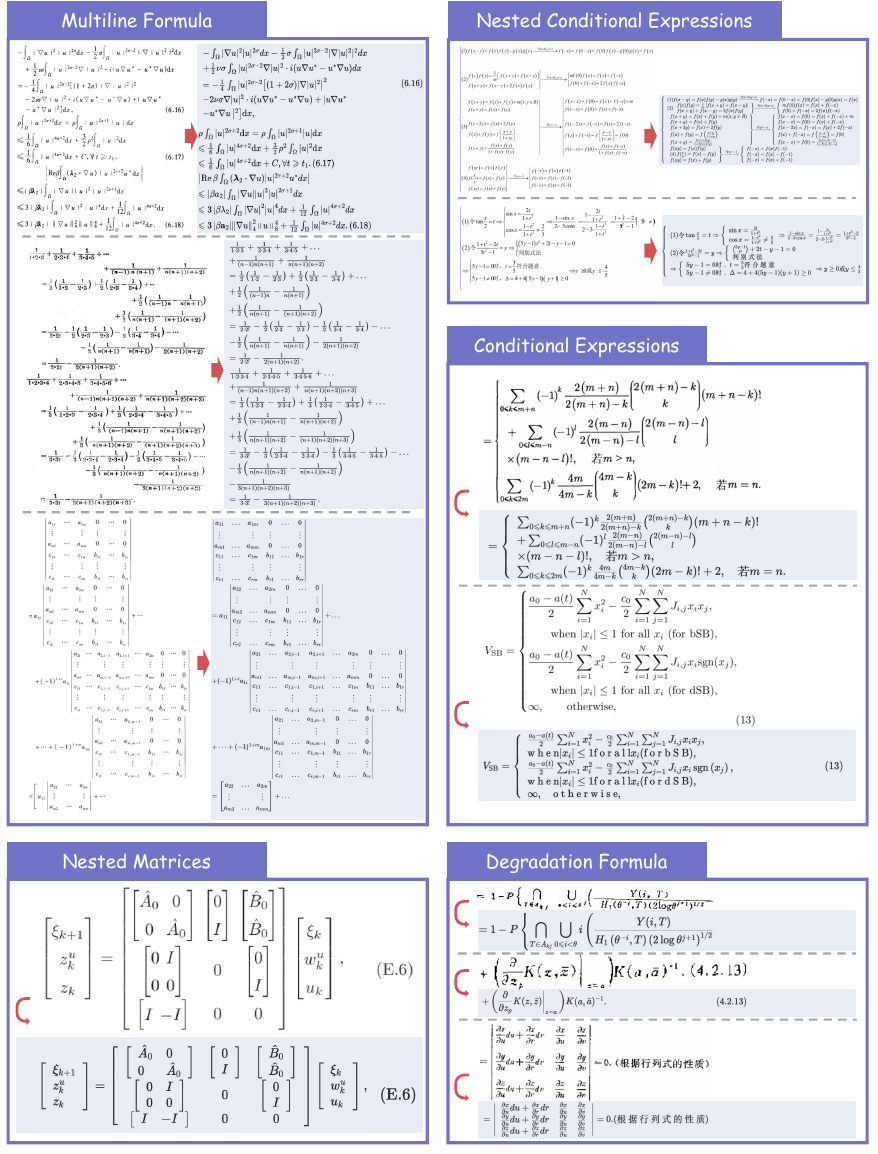}
\caption{The rendered outputs for various types of Formulas.}
\label{fig:Formula-Module-2}
\end{figure}

The performance of Formula Recognition among types of formulas is demonstrated in ~\Cref{fig:Formula-Module-1,fig:Formula-Module-2}, including formula with background, formula with Chinese, formula with matrix, formula with condition and nested condition, handwritten formula, blurred formula, multi-column formula, degradation formula. 

\begin{figure}[H]
\centering
\includegraphics[height=1.25\linewidth]{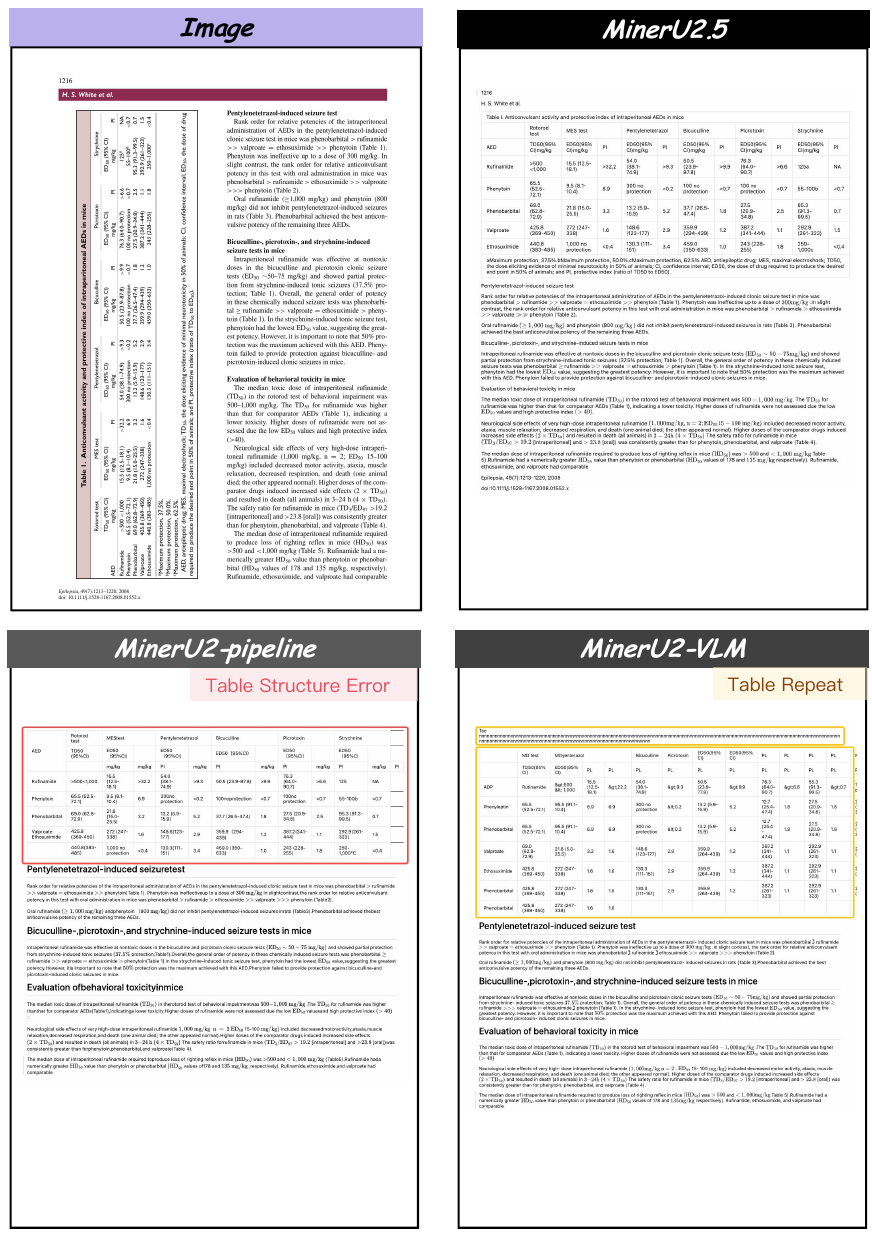}
\caption{Compare with Previous Version, MinerU2.5 performs better in rotated tables.}
\label{fig:compare_before_table_1}
\end{figure}

\begin{figure}[H]
\centering
\includegraphics[height=1.25\linewidth]{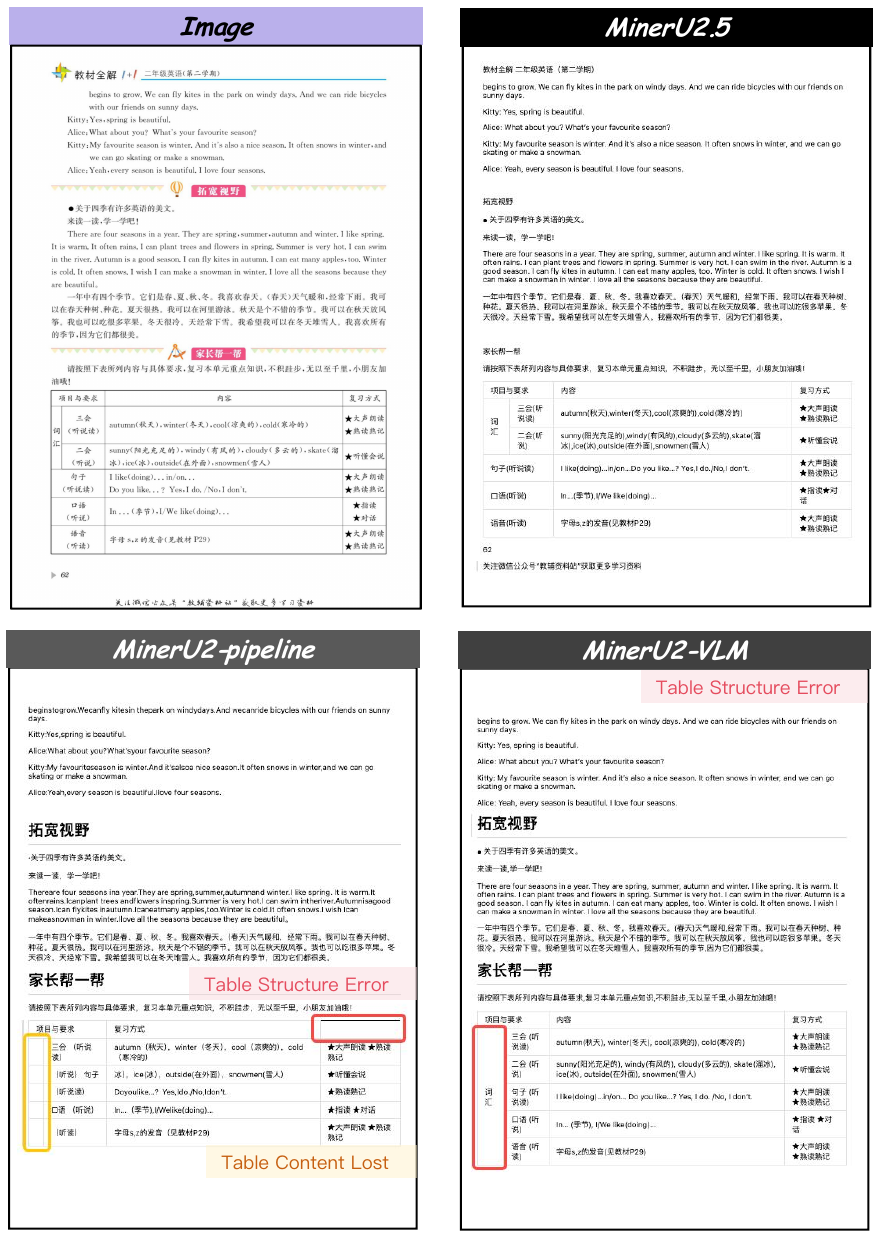}
\caption{Compare with Previous Version, MinerU2.5 performs better in tables with merged cells.}
\label{fig:compare_before_table_2}
\end{figure}

\begin{figure}[H]
\centering
\includegraphics[height=1.25\linewidth]{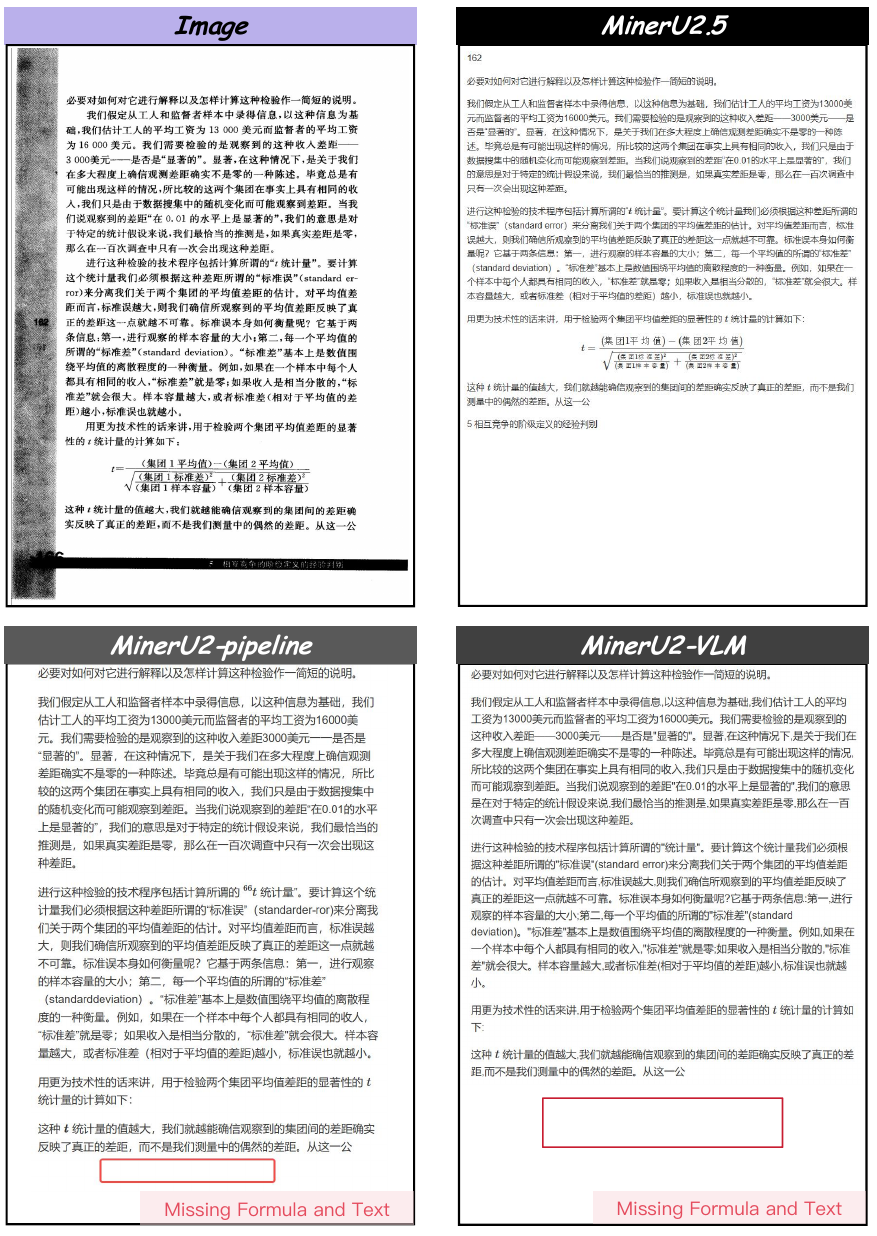}
\caption{Compare with Previous Version, MinerU2.5 performs better in Formula with Chinese.}
\label{fig:compare_before_formula_1}
\end{figure}

\begin{figure}[H]
\centering
\includegraphics[height=1.25\linewidth]{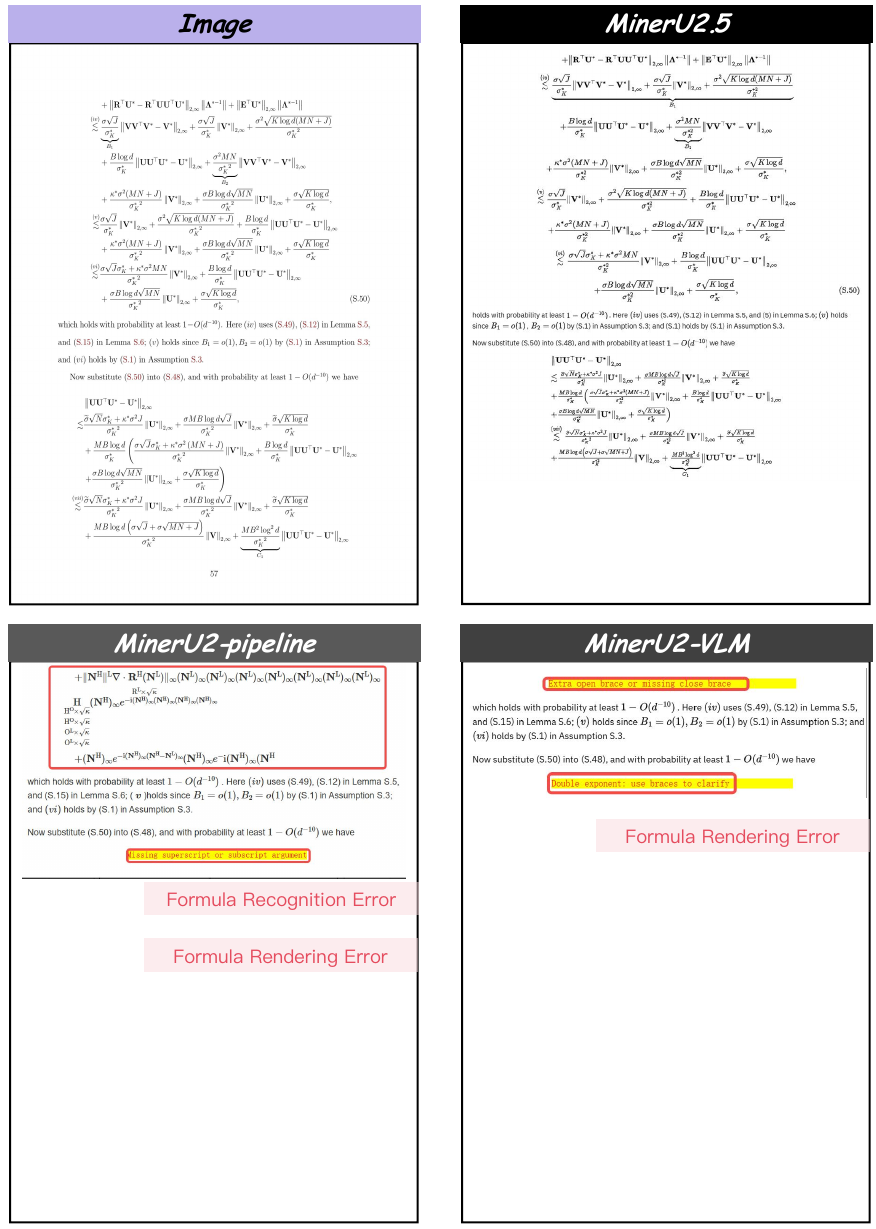}
\caption{Compare with Previous Version, MinerU2.5 performs better in multi-lines and complex Formula.}
\label{fig:compare_before_formula_2}
\end{figure}

\begin{figure}[H]
\centering
\includegraphics[height=1.25\linewidth]{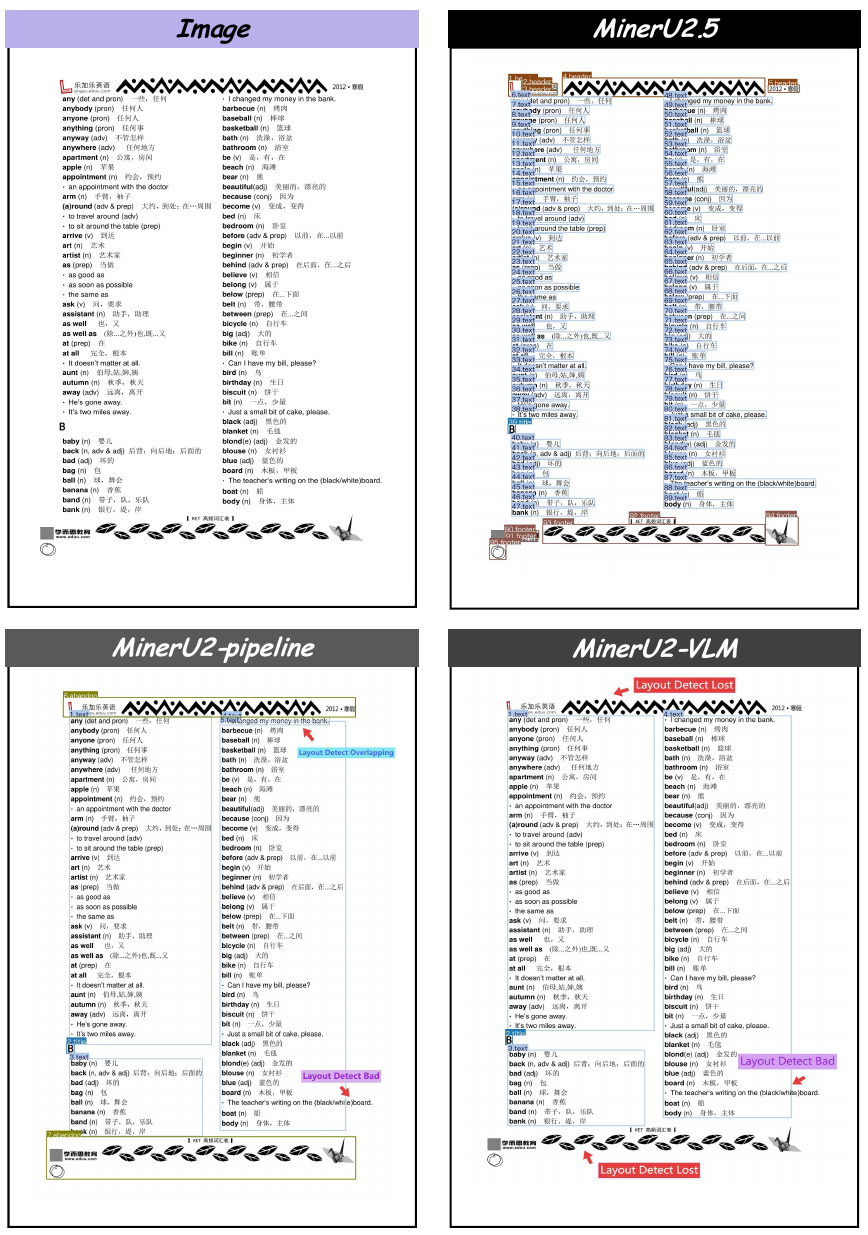}
\caption{Compare with Previous Version, MinerU2.5 achieve finer layout detection.}
\label{fig:compare_before_layout_1}
\end{figure}

\begin{figure}[H]
\centering
\includegraphics[height=1.25\linewidth]{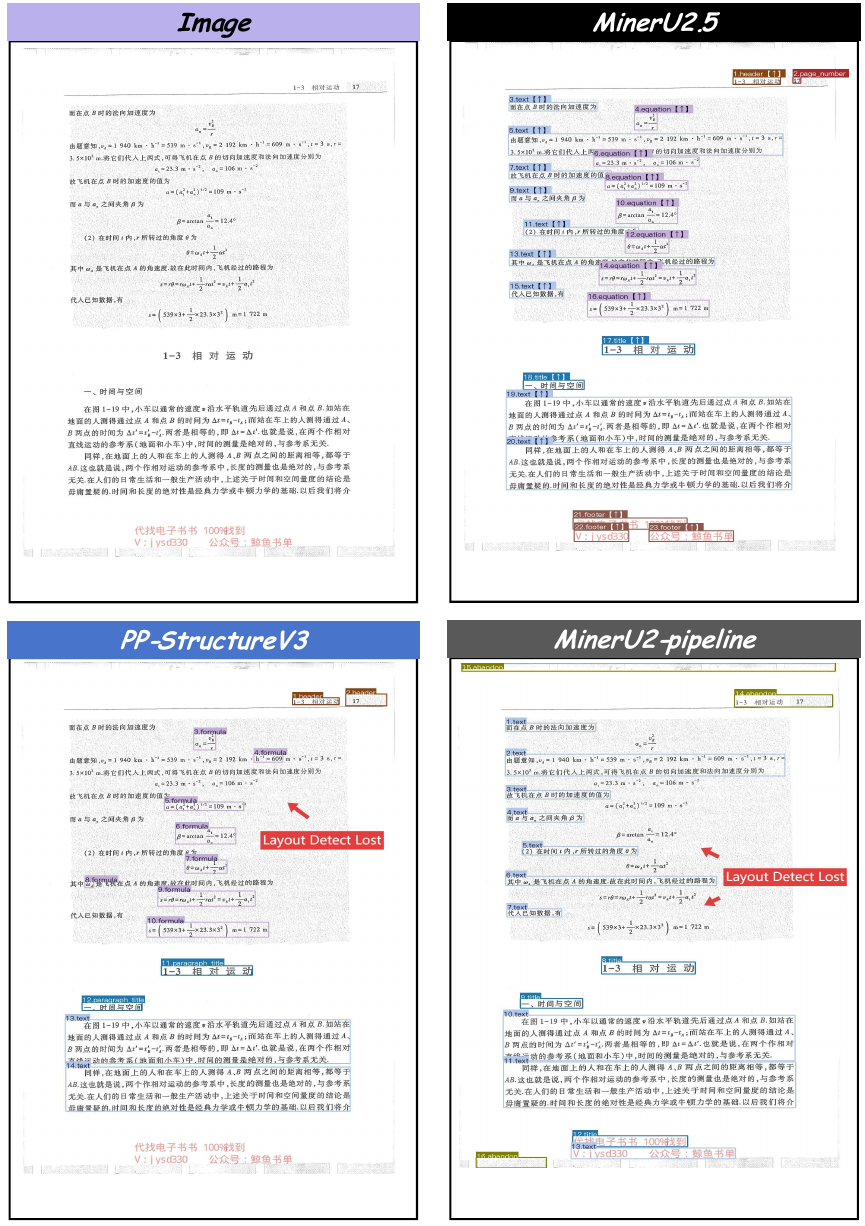}
\caption{Compare with Previous Version, MinerU2.5 achieve fewer detection omissions in watermark page.}
\label{fig:compare_before_layout_2}
\end{figure}

\Cref{fig:compare_before_table_1,fig:compare_before_table_2,fig:compare_before_formula_1,fig:compare_before_formula_2} demonstrate that \mineru{}'s document parsing ability improved when encounter rotated tables, table with merged cells, formula with Chinese and multi-line and complex formula, comparing with previous version (MinerU2-VLM, MinerU2-pipeline). Moreover, \mineru{} achieves finer bounding bbox in layout detection  and performs better on watermark pages than previous version, as illustrated in~\Cref{fig:compare_before_layout_1,fig:compare_before_layout_2}.

\mineru{} achieves outstanding performance in scenarios involving PDF pages with complex elements, and its performance is relatively better compared to existing state-of-the-art models. 

\begin{figure}[H]
\centering
\includegraphics[height=1.25\linewidth]{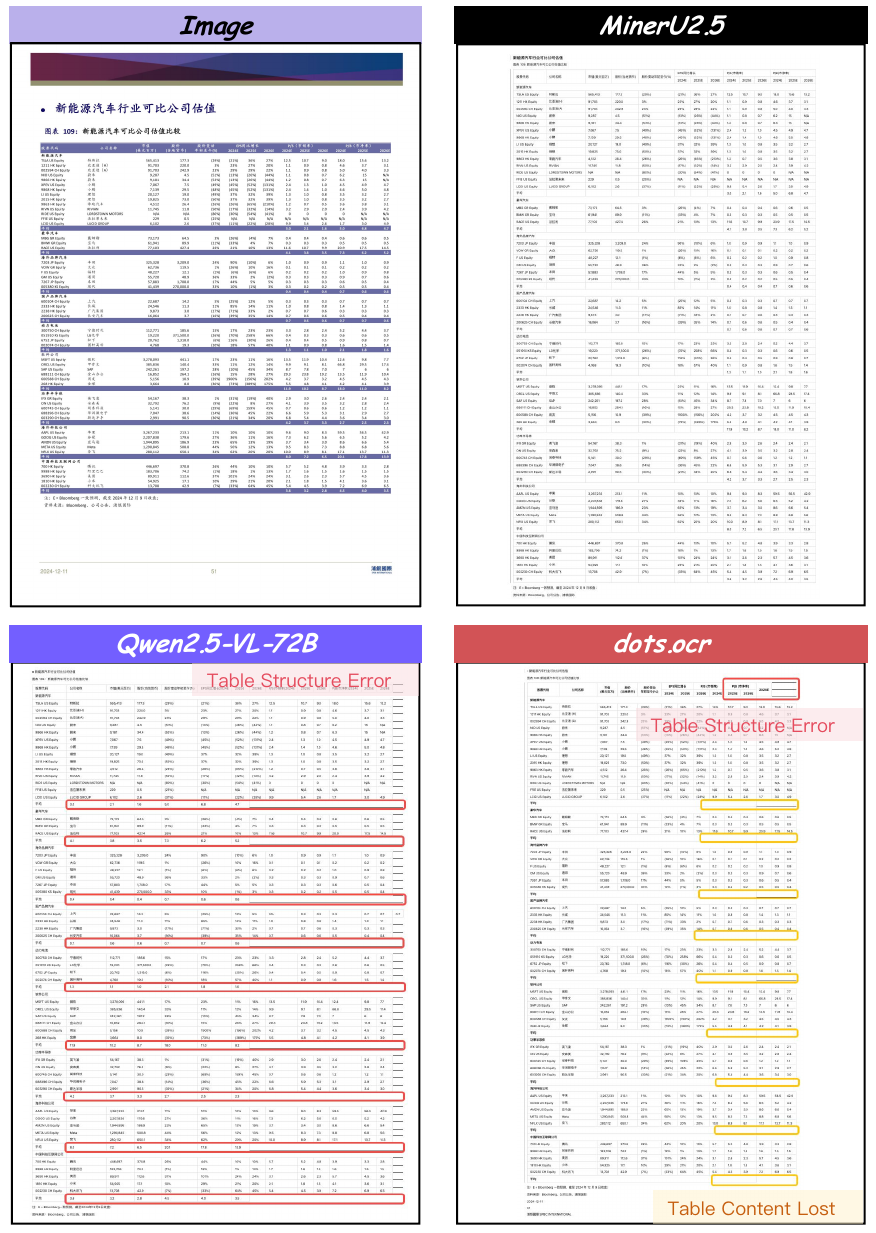}
\caption{Compare with others in Full-page table.}
\label{fig:compare_other_table-1}
\end{figure}

\begin{figure}[H]
\centering
\includegraphics[height=1.25\linewidth]{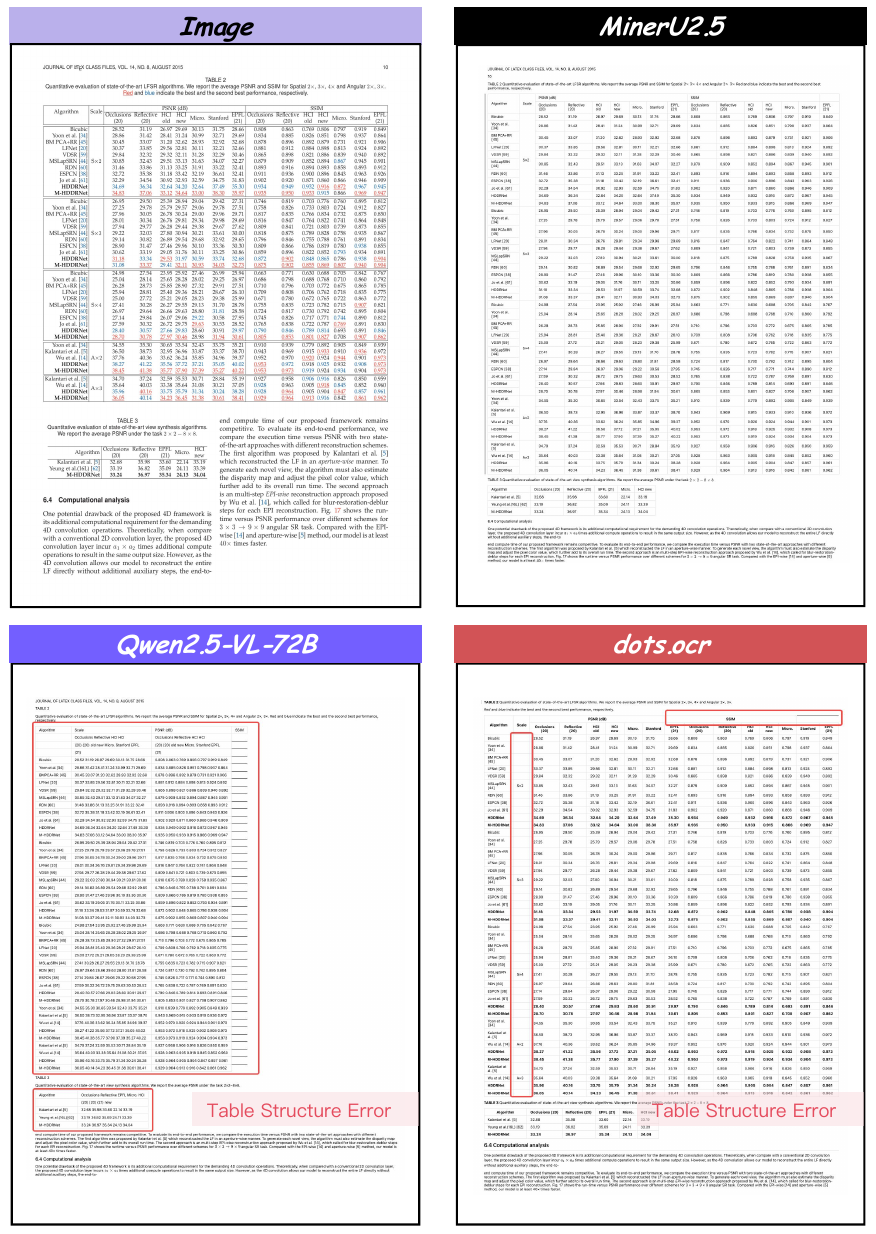}
\caption{Compare with others in content dense table.}
\label{fig:compare_other_table-2}
\end{figure}

\begin{figure}[H]
\centering
\includegraphics[height=1.25\linewidth]{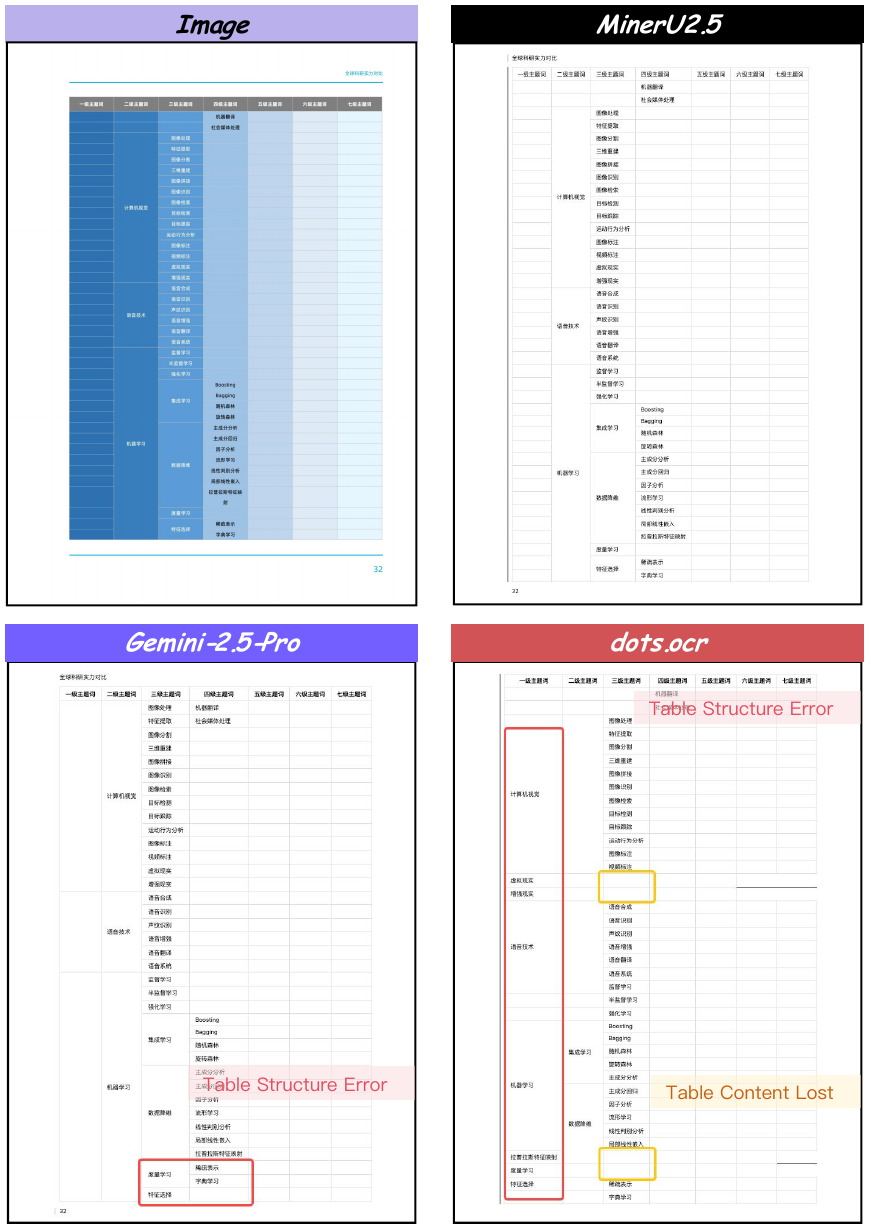}
\caption{Compare with others in Colored table with many empty cells.}
\label{fig:compare_other_table-3}
\end{figure}

\begin{figure}[H]
\centering
\includegraphics[height=1.25\linewidth]{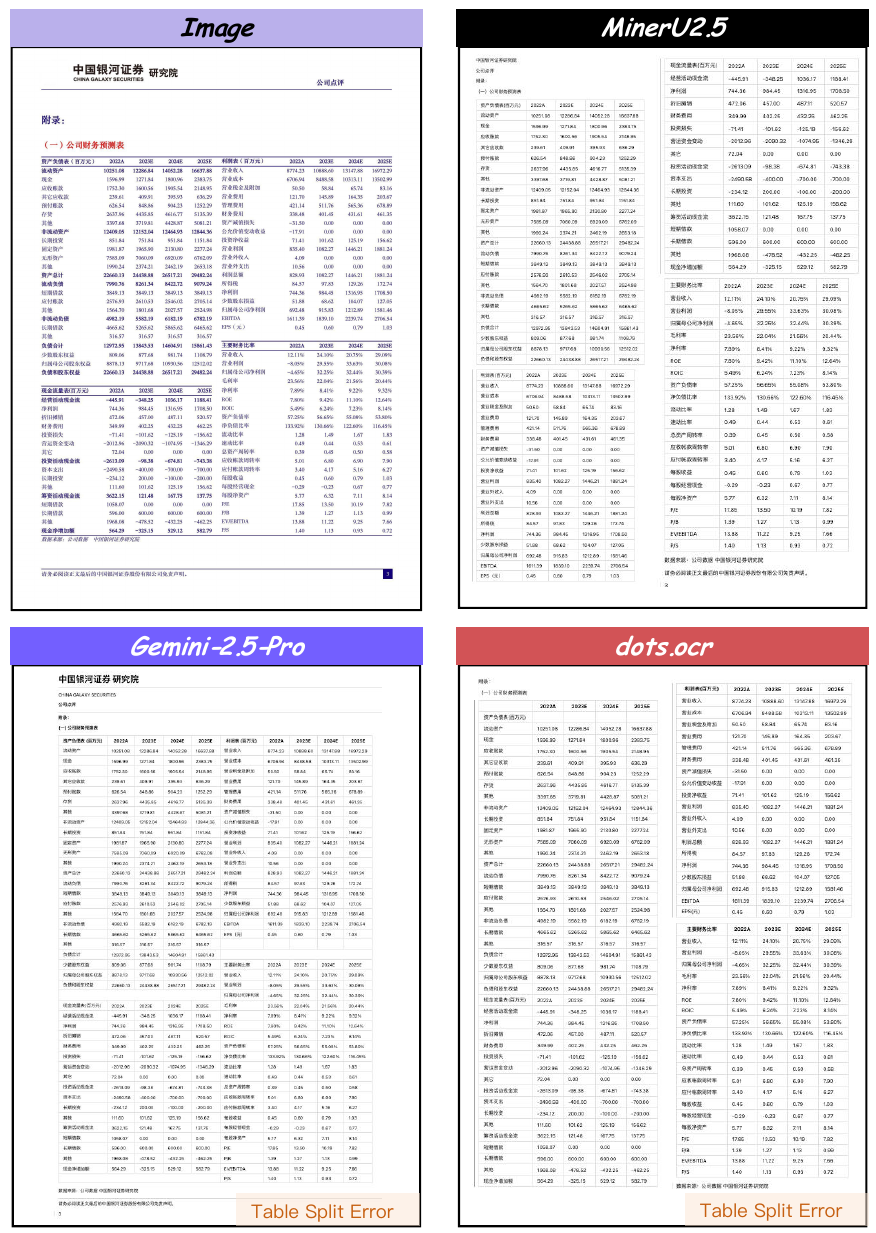}
\caption{Compare with others in Tightly arranged multiple tables.}
\label{fig:compare_other_table-4}
\end{figure}

\begin{figure}[H]
\centering
\includegraphics[height=1.25\linewidth]{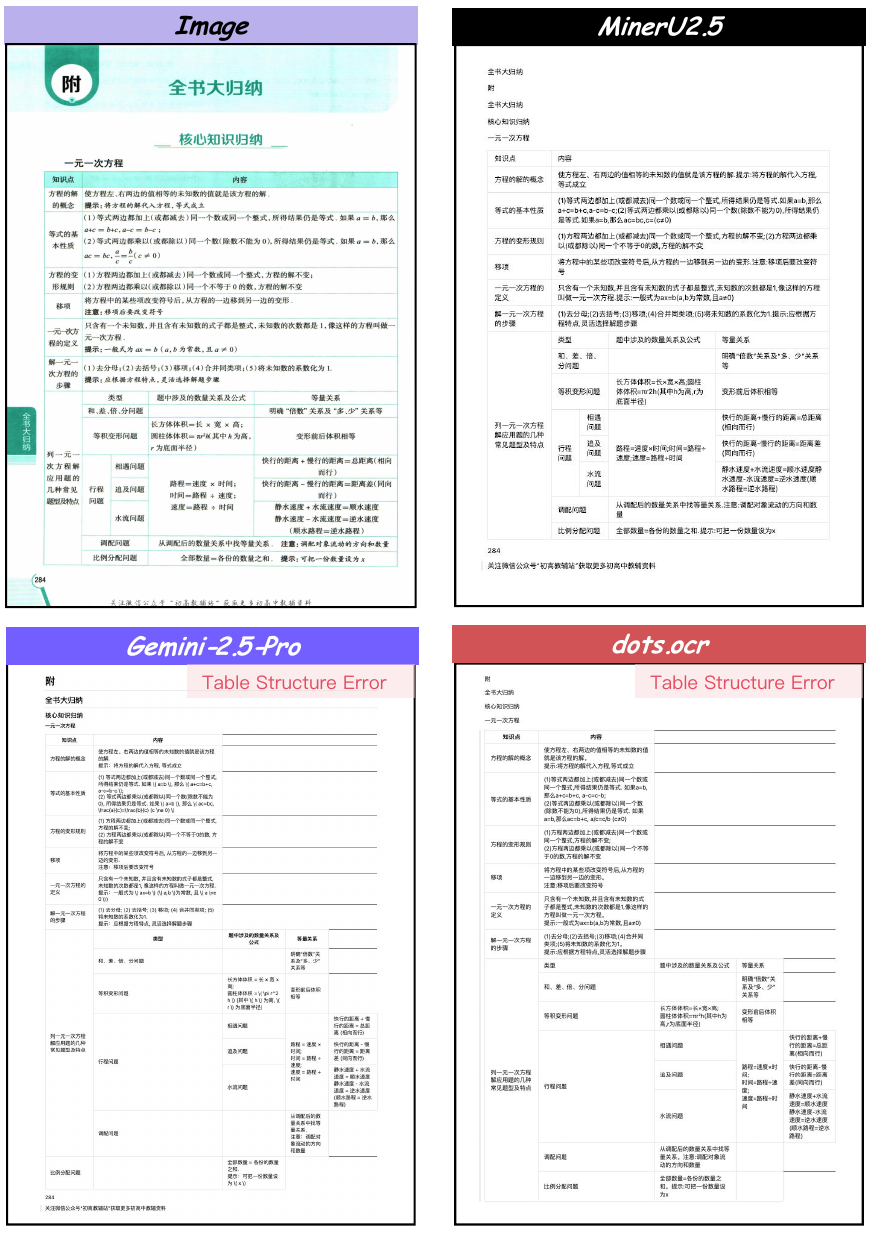}
\caption{Compare with others in Table with irregular merged cells.}
\label{fig:compare_other_table-5}
\end{figure}

\begin{figure}[H]
\centering
\includegraphics[height=1.25\linewidth]{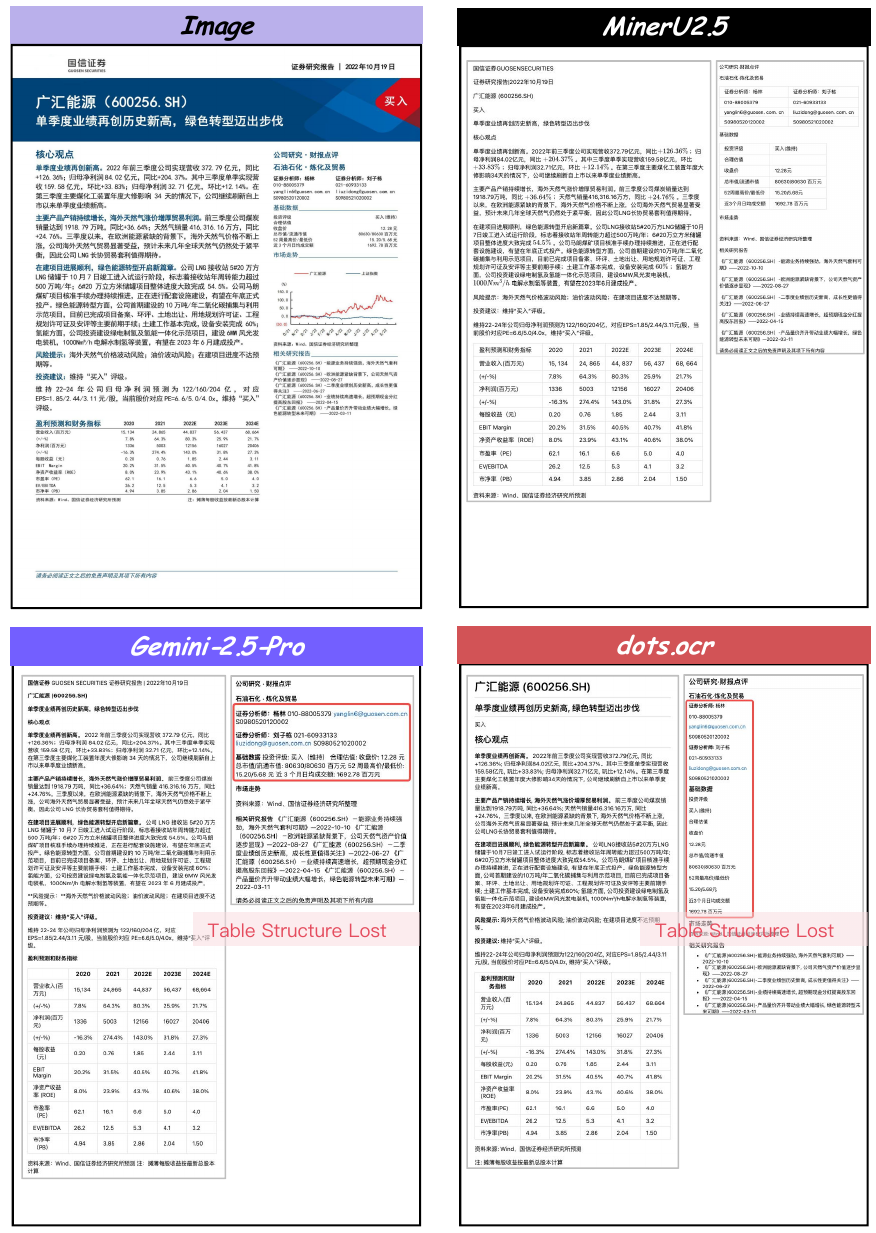}
\caption{Compare with others in Tables with No Frame.}
\label{fig:compare_other_table-6}
\end{figure}

\Cref{fig:compare_other_table-1,fig:compare_other_table-2,fig:compare_other_table-3,fig:compare_other_table-4,fig:compare_other_table-5,fig:compare_other_table-6} showcase the scenarios with complex tables in the page, including full-page table, content dense table, colorful table with amounts of empty cells, a tightly-arranged multiple table, table with irregular merged cells, a table without lines. \mineru{} can achieve better parsing outputs on these pages, while other models encounter errors such as table structure error, table structure lost, table content lost and table split error. 

\begin{figure}[H]
\centering
\includegraphics[height=1.25\linewidth]{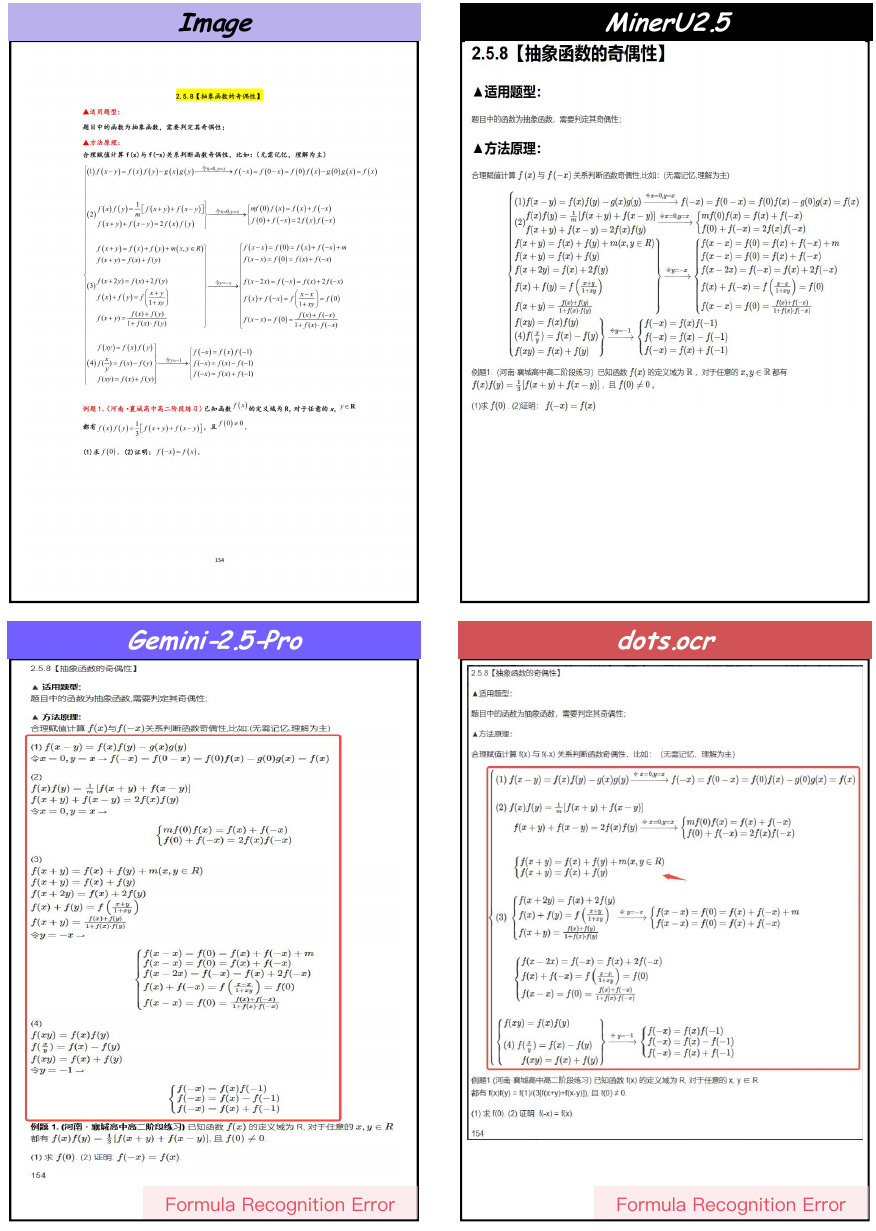}
\caption{Compare with others in Nested conditional expressions.}
\label{fig:compare_other_formula-1}
\end{figure}

\begin{figure}[H]
\centering
\includegraphics[height=1.25\linewidth]{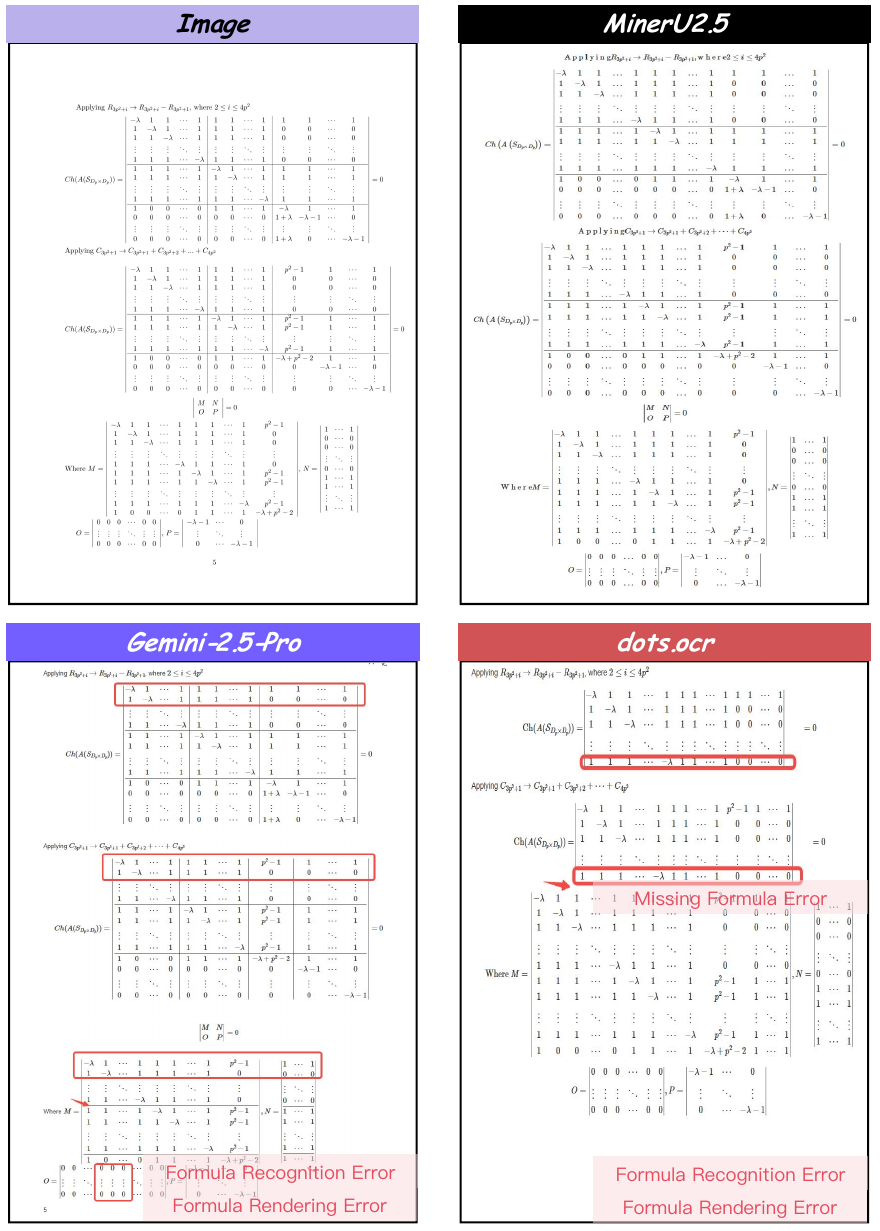}
\caption{Compare with others in Complex matrix.}
\label{fig:compare_other_formula-2}
\end{figure}

\begin{figure}[H]
\centering
\includegraphics[height=1.25\linewidth]{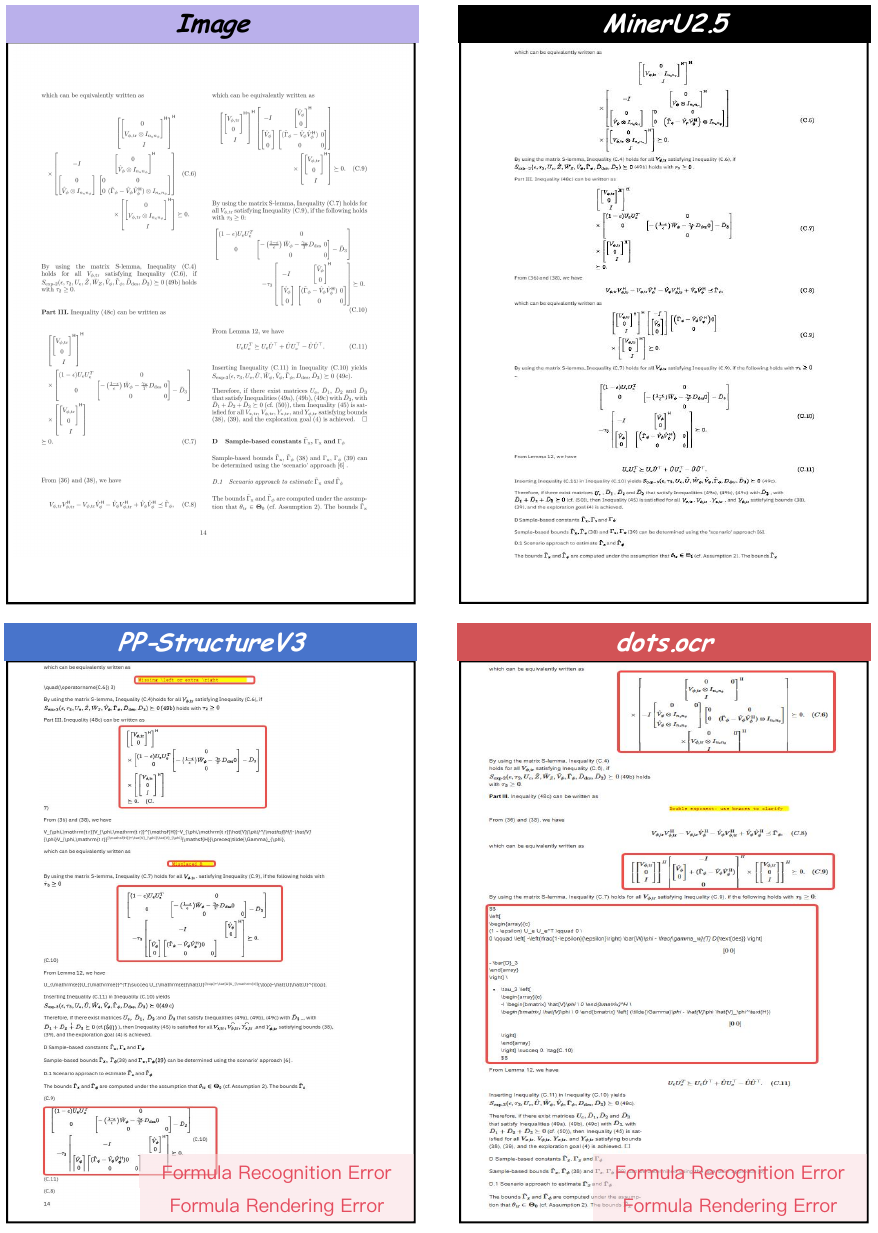}
\caption{Compare with others in Nested matrix.}
\label{fig:compare_other_formula-3}
\end{figure}

\Cref{fig:compare_other_formula-1,fig:compare_other_formula-2,fig:compare_other_formula-3} illustrates the performance of \mineru{} in the page with nested conditional expressions, complex matrix and nested matrix compared to other SOTA models, \mineru{} can correctly parse the complex formula while others might generate wrong outputs. 

\begin{figure}[H]
\centering
\includegraphics[height=1.25\linewidth]{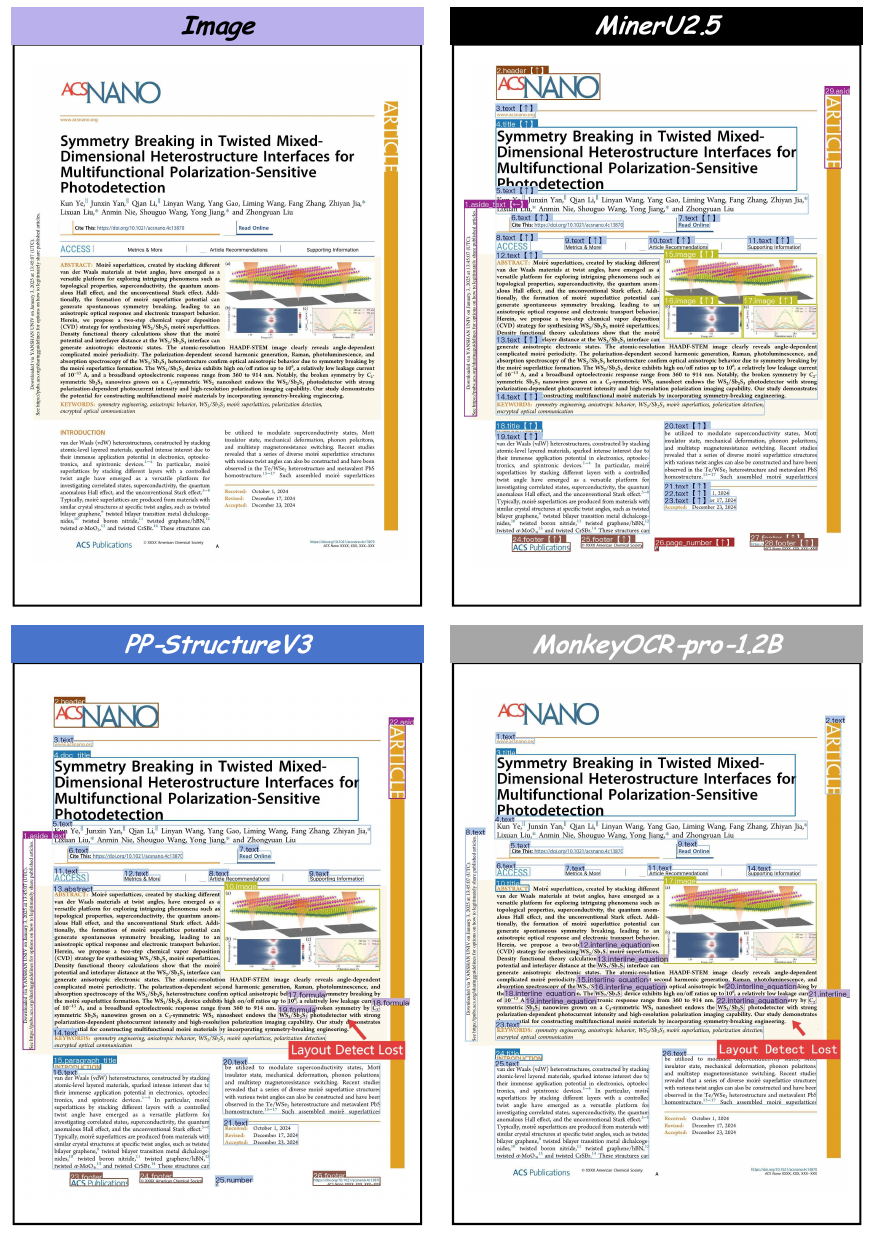}
\caption{Compare with others in Academic literature with alternating text and images.}
\label{fig:compare_other_layout-1}
\end{figure}

\begin{figure}[H]
\centering
\includegraphics[height=1.25\linewidth]{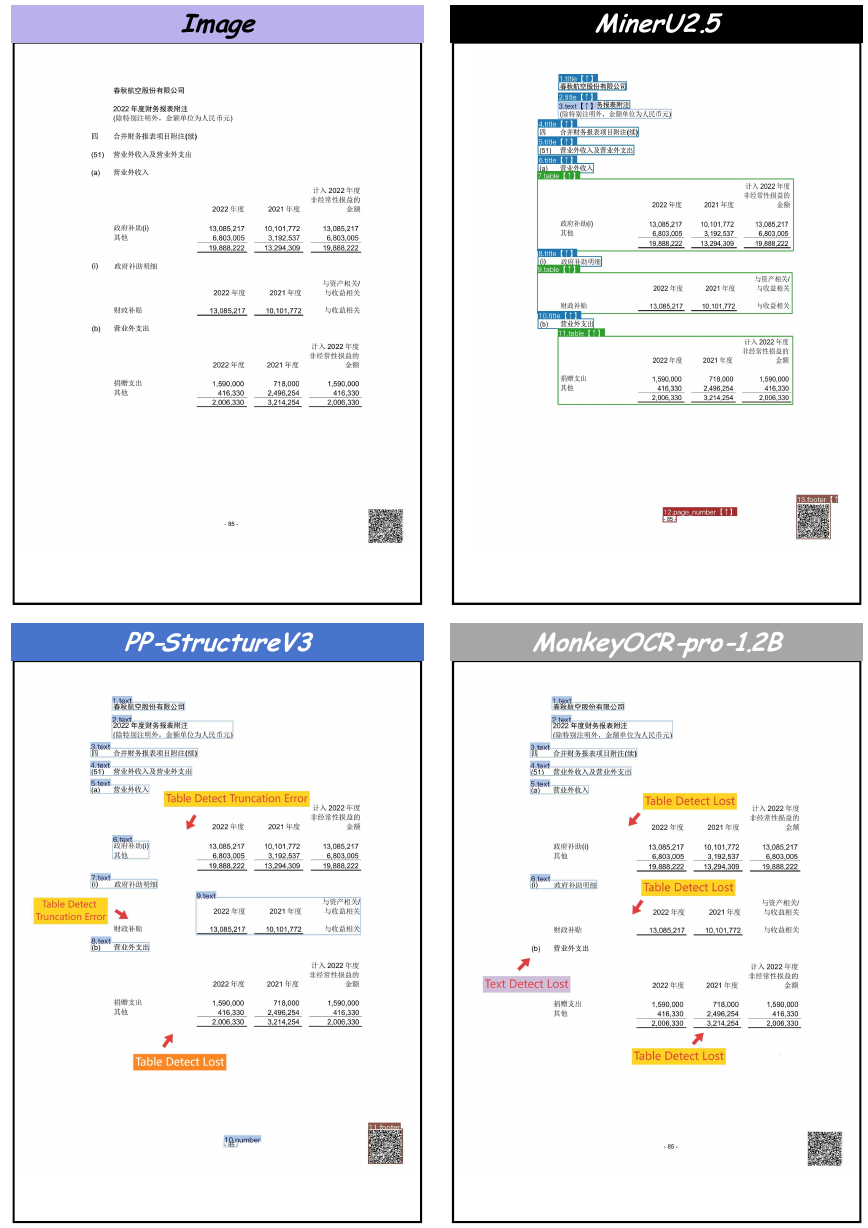}
\caption{Compare with others in Financial Report with Few Frame Tables.}
\label{fig:compare_other_layout-2}
\end{figure}

\begin{figure}[H]
\centering
\includegraphics[height=1.25\linewidth]{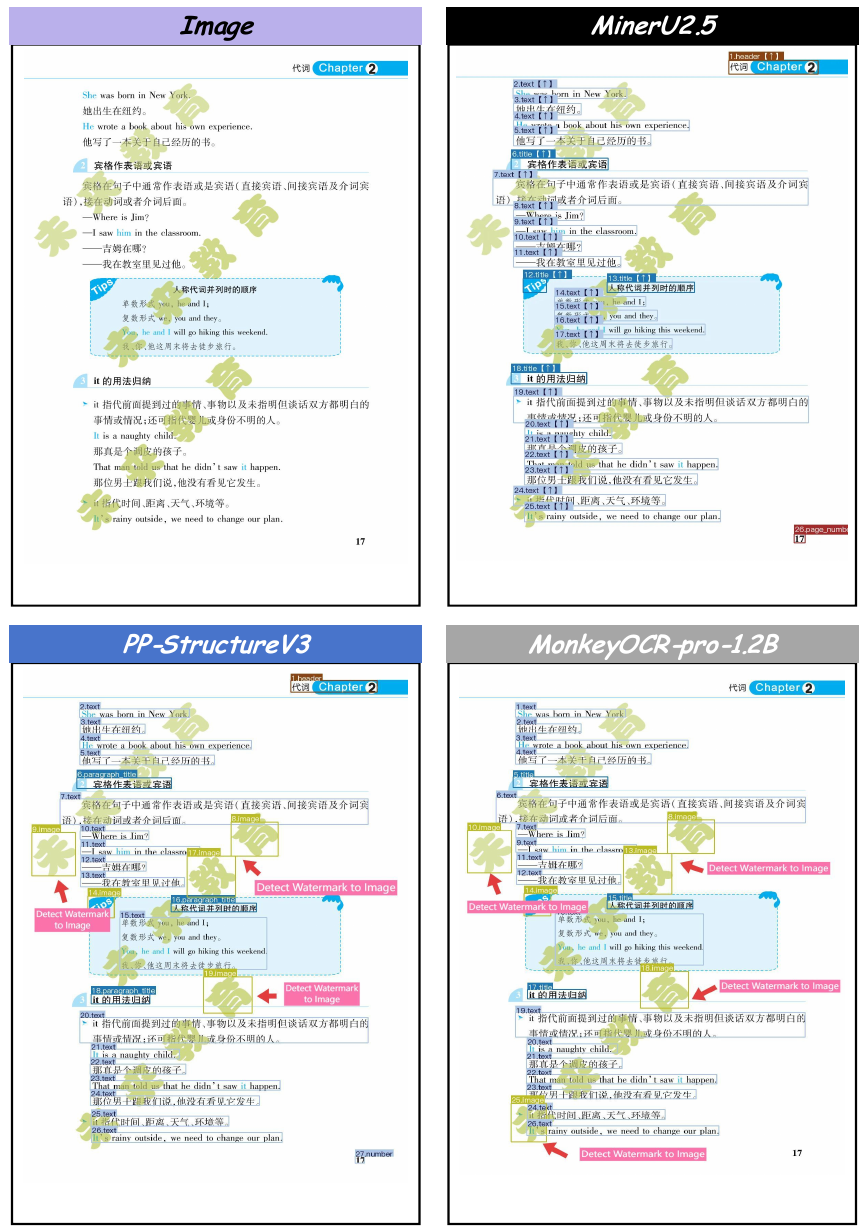}
\caption{Compare with others in Textbooks with watermarks.}
\label{fig:compare_other_layout-3}
\end{figure}

\Cref{fig:compare_other_layout-1,fig:compare_other_layout-2,fig:compare_other_layout-3} shows \mineru{}'s outstanding performance in pages with complex layout, e.g., alternating texts and images, with very-few frame tables, and pages with watermark compared with others.

\newpage
\subsection{Overview}
\label{subsec:overview}

\subsubsection{Among PDF types}











\newpage
\subsubsection{Among Table types}

\newpage
\subsubsection{Among Formula types}

\newpage
\subsection{Compare to Previous Versions}
\label{subsec:compare_before}

\subsubsection{Table}

\newpage
\subsubsection{Formula}

\newpage
\subsubsection{Layout\&OCR}

\newpage
\subsection{Compare with Others}
\label{subsec:compare_others}

\subsubsection{Table}

\newpage
\subsubsection{Formula}

\newpage
\subsubsection{Layout\&OCR}

\section{Prompt Details}
\label{app:prompt_detail}
Here, we provide a detailed description of the different prompts used during the two-stage inference of \mineru{}, along with their corresponding output formats.

\subsection{Layout Detection}
The layout detection output will include the relative coordinates, category, and rotation direction of each element in the document. Each element will be output in sequence, ensuring traceability for all layout data. The input image will be resized to a resolution of 1036 $\times$ 1036.

\textbf{Output format:}
\begin{itemize}
    \item \textbf{Box Coordinates:} x1, y1, x2, y2
    \item \textbf{Document Element Category:} title, text, image, etc.
    \item \textbf{Rotation Direction:} up, down, left, right
\end{itemize}

\textbf{Example:}
\begin{lstlisting}[numbers=none]
<|box_start|>100 200 300 400<|box_end|><|ref_start|>title<|ref_end|><|rotate_up|>
<|box_start|>400 500 600 700<|box_end|><|ref_start|>text<|ref_end|><|rotate_up|>
\end{lstlisting}

\subsection{Text Recognition}
The output will contain the recognized text results. The input image will retain its native resolution; however, the number of image tokens will be limited to the range of 4 to 2048. If this limit is exceeded, the image will be scaled accordingly.

\textbf{Output format:}
\begin{itemize}
    \item \textbf{OCR Results:} The raw OCR output
\end{itemize}

\textbf{Example:}
\begin{lstlisting}[numbers=none]
The results of the analyses of the uncertainty of the field data and related assumptions are shown in Figs 13 and 14.
\end{lstlisting}

\subsection{Formula Recognition}
Any formulas found in the image will be extracted and converted into LaTeX format. The input image will retain its native resolution; however, the number of image tokens will be limited to the range of 4 to 2048. If this limit is exceeded, the image will be scaled accordingly.

\textbf{Output format:}
\begin{itemize}
    \item \textbf{LaTeX Format:} The LaTeX representation of the formula
\end{itemize}

\textbf{Example:}
\begin{lstlisting}[numbers=none]
\[
\hat{F} = \operatorname{Concat}\left(\left[ F_{1}, F_{2}, \dots, F_{n} \right]\right) \tag{2}
\]

\[
M = \sigma \bigl( \mathrm{GELU}(\mathrm{BN}(\mathrm{Conv}_{gate}(\hat{F}))) \bigr) \tag{3}
\]
\end{lstlisting}

\subsection{Table Recognition}
The output will include the recognized tables, structured in an OTSL (Open Table Structure Language) format for easy data processing. The input image will retain its native resolution; however, the number of image tokens will be limited to the range of 4 to 2048. If this limit is exceeded, the image will be scaled accordingly.

\textbf{Output format:}
\begin{itemize}
    \item \textbf{OTSL Format:} The table represented in OTSL format
\end{itemize}

\textbf{Example:}
\begin{lstlisting}[numbers=none]
<fcel>Site<fcel>Cl<fcel>NO3<fcel>SO4<fcel>Na<fcel>Ca<fcel>K<fcel>Mg<fcel>NH4<fcel>References<nl>
<fcel>Cl dominance sites<lcel><lcel><lcel><lcel><lcel><lcel><lcel><lcel><lcel><nl>
<fcel>Comba<fcel>109.8<fcel>12.1<fcel>23.3<fcel>86.8<fcel>43.4<fcel>4.8<fcel>15.1<fcel>13.2<fcel>Present study<nl>
<fcel>Alibagh<fcel>236<fcel>9<fcel>36<fcel>220<fcel>46<fcel>5<fcel>64<fcel>8<fcel>Naik et al. (2002)<nl>
<fcel>Goa<fcel>113.4<fcel>5.5<fcel>27.4<fcel>97.2<fcel>41.5<fcel>2.5<fcel>24.5<fcel>5.5<fcel>Parashar et al. (2001)<nl>
<fcel>Bombay<fcel>138<fcel>-<fcel>10<fcel>115<fcel>36<fcel>3.6<fcel>24<fcel>-<fcel>Sequeira (1976)<nl>
<fcel>Na dominance sites<lcel><lcel><lcel><lcel><lcel><lcel><lcel><lcel><lcel><nl>
<fcel>Colaba<fcel>171<fcel>34<fcel>52<fcel>179<fcel>133<fcel>6<fcel>59<fcel>12<fcel>Naik et al. (2002)<nl>
<fcel>Silent Valley<fcel>43.0<fcel>21.0<fcel>20.0<fcel>46.0<fcel>43.0<fcel>4.0<fcel>14.0<fcel>3.0<fcel>Rao et al. (1995)<nl>
<fcel>Chembur<fcel>164.5<fcel>29.5<fcel>70.4<fcel>168.2<fcel>89.5<fcel>6.9<fcel>36.5<fcel>41.1<fcel>Khemani et al. (1994)<nl>
<fcel>Bhubaneswar<fcel>18<fcel>10<fcel>19.1<fcel>15<fcel>20.2<fcel>1.8<fcel>5.2<fcel>18.7<fcel>Das et al. (2005)<nl>
\end{lstlisting}

\end{document}